\documentclass{ieeeaccess}
\usepackage{cite}
\usepackage{amsmath,amssymb,amsfonts}
\usepackage{algorithmic}
\usepackage{graphicx}
\usepackage{textcomp}
\usepackage{float}
\usepackage{soul}
\usepackage{subcaption}
\usepackage{multirow}
\usepackage{changepage}
\usepackage{hyperref}
\usepackage{ulem}
\usepackage{adjustbox}
\usepackage{changepage}
\makeatother
\usepackage{comment}
\usepackage{xcolor}
\usepackage{pict2e}
 
\newcommand\orcidicon[1]{\href{https://orcid.org/#1}{\usebox{\ORCIDlogo}}}
\newsavebox{\ORCIDlogo}
\savebox{\ORCIDlogo}{%
\setlength{\unitlength}{\dimexpr 1em/256\relax}%
\begin{picture}(256,256)%
  \color[HTML]{A6CE39}\put(128,128){\circle*{256}}%
  \color{white}%
  \put(78.6,199.2){\circle*{20}}%
  \moveto(70.9,176,9)\lineto(86.3,176,9)\lineto(86.3,69.8)\lineto(70.9,69.8)%
  \closepath\fillpath%
  \moveto(108.9,176.9)\lineto(150.5,176.9)%
  \curveto(190.1,176.9)(207.5,148.6)(207.5 ,123.3)%
  \curveto(207.5,95,8)(186,69.7)(150.7,69.7)%
  \lineto(108.9,69.7)%
  \closepath\fillpath%
  \color[HTML]{A6CE39}%
  \moveto(124.3,83.6)\lineto(148.8,83.6)%
  \curveto(183.7,83.6)(191.7,110.1)(191.7,123.3)%
  \curveto(191.7,144.8)(178,163)(148,163)%
  \lineto(124.3,163)%
  \closepath\fillpath%
\end{picture}%
}
\def\BibTeX{{\rm B\kern-.05em{\sc i\kern-.025em b}\kern-.08em
T\kern-.1667em\lower.7ex\hbox{E}\kern-.125emX}}
\usepackage{CJKutf8}
\def\BibTeX{{\rm B\kern-.05em{\sc i\kern-.025em b}\kern-.08em
T\kern-.1667em\lower.7ex\hbox{E}\kern-.125emX}}
\usepackage{xcolor}
\usepackage{pict2e}
\savebox{\ORCIDlogo}{%
\setlength{\unitlength}{\dimexpr 1em/256\relax}%
\begin{picture}(256,256)%
  \color[HTML]{A6CE39}\put(128,128){\circle*{256}}%
  \color{white}%
  \put(78.6,199.2){\circle*{20}}%
  \moveto(70.9,176,9)\lineto(86.3,176,9)\lineto(86.3,69.8)\lineto(70.9,69.8)%
  \closepath\fillpath%
  \moveto(108.9,176.9)\lineto(150.5,176.9)%
  \curveto(190.1,176.9)(207.5,148.6)(207.5 ,123.3)%
  \curveto(207.5,95,8)(186,69.7)(150.7,69.7)%
  \lineto(108.9,69.7)%
  \closepath\fillpath%
  \color[HTML]{A6CE39}%
  \moveto(124.3,83.6)\lineto(148.8,83.6)%
  \curveto(183.7,83.6)(191.7,110.1)(191.7,123.3)%
  \curveto(191.7,144.8)(178,163)(148,163)%
  \lineto(124.3,163)%
  \closepath\fillpath%
\end{picture}%
}
\usepackage{hyperref} 
\hypersetup{
  colorlinks=false,
  linkbordercolor=white,
 urlbordercolor=white,
pdfborder={0 0 0}
}
\usepackage{CJKutf8}

\begin{document}
\history{Date of publication xxxx 00, 0000, date of current version xxxx 00, 0000.}
\doi{10.1109/ACCESS.2017.DOI}
\title{A Methodological and Structural Review of Hand Gesture Recognition Across Diverse Data Modalities}
\author{
\uppercase{Abu Saleh Musa Miah\ \orcidicon{0000-0002-1238-0464} \authorrefmark{1}, (IEEE Member)}, \uppercase{Jungpil Shin \orcidicon{0000-0002-7476-2468}\authorrefmark{1},(Senior IEEE Member)}
  \uppercase{Md. Humaun Kabir\ \orcidicon{0000-0001-7225-0736} \authorrefmark{2}}, 
\uppercase{Md. Abdur Rahim\ \orcidicon{0000-0003-2300-1420} \authorrefmark{3}},
\uppercase{Abdullah Al Shiam\ \orcidicon{0000-0002-8787-5584} \authorrefmark{4}}}

 \address[1]{School of Computer Science and Engineering, The University of Aizu, Aizuwakamatsu, Japan (e-mail:jpshin@u-aizu.ac.jp, d8231105@u-aizu.ac.jp)}
\address[2]{Department of Computer Science and Engineering, Bangamata Sheikh Fojilatunnesa Mujib Science \& Technology University, Jamalpur 2012, Bangladesh (e-mail: humaun@bsfmstu.ac.bd)}
\address[3]{Department of Computer Science and Engineering, Pabna University of Science and Technology, Rajapur, Pabna 6600, Bangladesh (e-mail: rahim@pust.ac.bd)}
\address[4]{Department of Computer Science and Engineering, Sheikh Hasina University, Netrokona 2400, Bangladesh, (e-mail: shiam.cse@shu.edu.bd)}
\markboth
{....}
{This paper is currently under review for possible publication in IEEE Access.}
\corresp{Corresponding author: Jungpil Shin (e-mail: jpshin@u-aizu.ac.jp).}
\begin{abstract}
Researchers have been developing Hand Gesture Recognition (HGR) systems to enhance natural, efficient, and authentic human-computer interaction, especially benefiting those who rely solely on hand gestures for communication. Despite significant progress, the automatic and precise identification of hand gestures remains a considerable challenge in computer vision. Recent studies have focused on specific modalities like RGB images, skeleton data, and spatiotemporal interest points. This paper provides a comprehensive review of HGR techniques and data modalities from 2014 to 2024, exploring advancements in sensor technology and computer vision. We highlight accomplishments using various modalities, including RGB, Skeleton, Depth, Audio, EMG, EEG, and Multimodal approaches and identify areas needing further research. We reviewed over 200 articles from prominent databases, focusing on data collection, data settings, and gesture representation. Our review assesses the efficacy of HGR systems through their recognition accuracy and identifies a gap in research on continuous gesture recognition, indicating the need for improved vision-based gesture systems. The field has experienced steady research progress, including advancements in hand-crafted features and deep learning (DL) techniques. Additionally, we report on the promising developments in HGR methods and the area of multimodal approaches. We hope this survey will serve as a potential guideline for diverse data modality-based HGR research.
 \end{abstract}
\titlepgskip=-15pt
\maketitle
\section{Introduction}
\label{sec1}
In our daily interactions, non-verbal communication plays a crucial role, conveying approximately 65\% of human messages, compared to verbal communication, which accounts for only 35\%   \cite{han2021sign,miah2024hand_multi}. Nowadays, people increasingly use gestures to control daily devices such as televisions, computers, fans, and air-conditioning systems. Non-verbal communication includes body gestures like head movements, facial expressions, nodding, shaking the head, mouth movements, winking, eye gaze direction, body movements, arm gestures, and hand gestures. Effective gesture recognition methods are essential for ensuring robust human-computer interaction (HCI), offering alternatives to traditional tools like mice and keyboards \cite{sarma2021methods, lu2020driver}. 
Among these, hand gestures are particularly useful in daily activities. Automatic recognition of hand gestures is crucial for establishing natural non-verbal communication. SLR, in particular, is a vital application that bridges the communication gap between deaf and non-deaf communities. Systems that translate hand movements into speech or text are invaluable for communication, education, and rehabilitation, especially when a human interpreter is unavailable.

Despite extensive research on static and dynamic HGR systems for various applications, several challenges remain. These challenges include adapting to diverse inputs like environmental noise, signer variations, and language differences \cite{gao2020two}. Constraints during development often involve the signer’s environment to mitigate segmentation and tracking issues. Managing gesture transitions, especially in continuous sign language \cite{kakizaki2024dynamic,islam2024multilingual}, is difficult and can lead to incorrect recognition results. This complexity limits the practicality of vision-based gesture recognition in real-world settings \cite{mohamed2021review,lee2021electromyogram}. Creating robust signer-independent HGR systems is another significant challenge. Such systems should work with users who were not part of the training phase, enabling broad applicability without requiring individual training for each new user \cite{alhamazani2021using,gao2020listen}. Previous research has extensively covered both device-based and vision-based HGR systems. While vision-based systems are ideal for diverse real-world applications, existing reviews often provide broad overviews without focusing on specific advancements or future directions. To address this gap, this paper conducts a thorough review of current and historical literature, analyzing advancements in vision-based HGR systems and exploring potential future research directions.

\subsection{Background}
HGR is a technology that translates hand movements in sign language into text or speech \cite{mallik2024virtual,rahim2020hand}. It can be divided into two types: vision-based systems and device-based systems, based on how they capture hand gestures. Vision-based HGR systems offer a more natural interaction experience as users do not need to wear any cumbersome devices. These systems find wider applications in outdoor settings due to their ease of use. However, challenges arise in handling dynamic sign language datasets containing both isolated and continuous gestures. Existing research tends to focus on recognizing isolated gestures, limiting their practical applicability. More robust feature extraction and discrimination methods are necessary to enhance vision-based systems. It also needs a temporal modelling-based HGR system. Due to numerous applications, HGR has sparked significant research interest, as highlighted in numerous review papers \cite{6835606_review_2013, wadhawan2021sign,aloysius2020understanding,rastgoo2021sign}.
In 2013, Chen et al. surveyed the HGR methodology, vision-based, depth-based and glove-based approach \cite{6835606_review_2013}. Check et al. surveyed the state-of-the-art approach used in the hand gesture-based recognition system in 2019. They summarized the preprocessing, segmentation, augmentation and classification technique of the hand feature recognition system \cite{cheok2019review}. Aloysius et al. surveyed only vision-based video or continuous sign language recognition (CSLR system) in 2020 \cite{aloysius2020understanding}. By including the dynamic dataset-based HGR, Wadhawan et al. surveyed academic literature spanning from 2007 to 2017, where they included six key dimensions such as dataset collection approach, different types of signs based on time, mode of sign, one-hand or two-hand sign,  classification approach and rates of recognition in 2021\cite{wadhawan2021sign}.  Ratsgoo et al. also surveyed vision-based HGR for SLR in 2021 \cite{rastgoo2021sign}.  
Recent surveys have further expanded the field, such as Jain et al. provided a comprehensive review of DL approaches for HGR in 2022, focusing on advancements in model architectures and applications in human-computer interaction \cite{jain2022literature_survey_2022}. In 2024, Jing et al. reviewed vision-based various data modalities for HGR, where they mainly focused on the monocular and RGB-D cameras, discussing data acquisition, hand gesture detection and segmentation, feature extraction, and gesture classification. It also reviews experimental evaluations and highlights advances required to improve current HGR systems for more effective and efficient HRI. \cite{qi2024computer_survey_RGBD}. Taw et al. provided a comprehensive review of gesture recognition based on DL, highlighting the roles of CNN, LSTM networks, and transfer learning. The paper discusses the strengths and limitations of each technique and their applications in various fields, including human-computer interaction, virtual reality, healthcare, and smart homes \cite{10526274_survey_2024_tradition_deep}. By integrating these recent surveys, we found that no existing paper has studied the various modality-based HGR research work based on recent publication articles. 
\begin{figure}[htp]
    \centering
    \includegraphics[width=8cm]{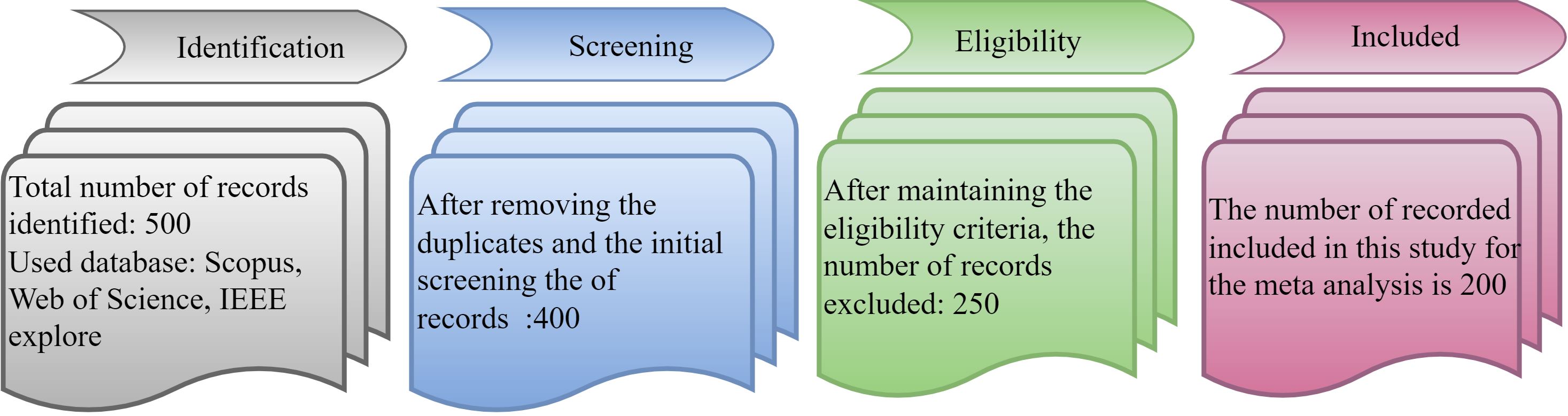}
    \caption{Article selection process procedure.}
    \label{fig:article_slection_pro}
    \end{figure}

\begin{figure}[htp]
    \centering
    \includegraphics[width=6cm]{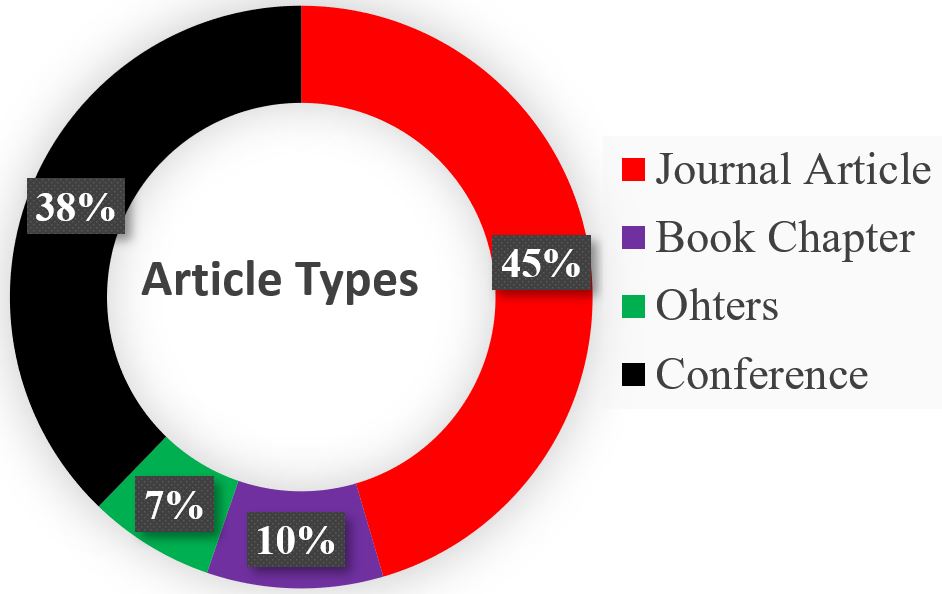}
    \caption{Articles type journal, conference, book chapters, and others.}
    \label{fig:intro_reference}
\end{figure}

\begin{figure}[htp]
    \centering
    \includegraphics[width=9cm]{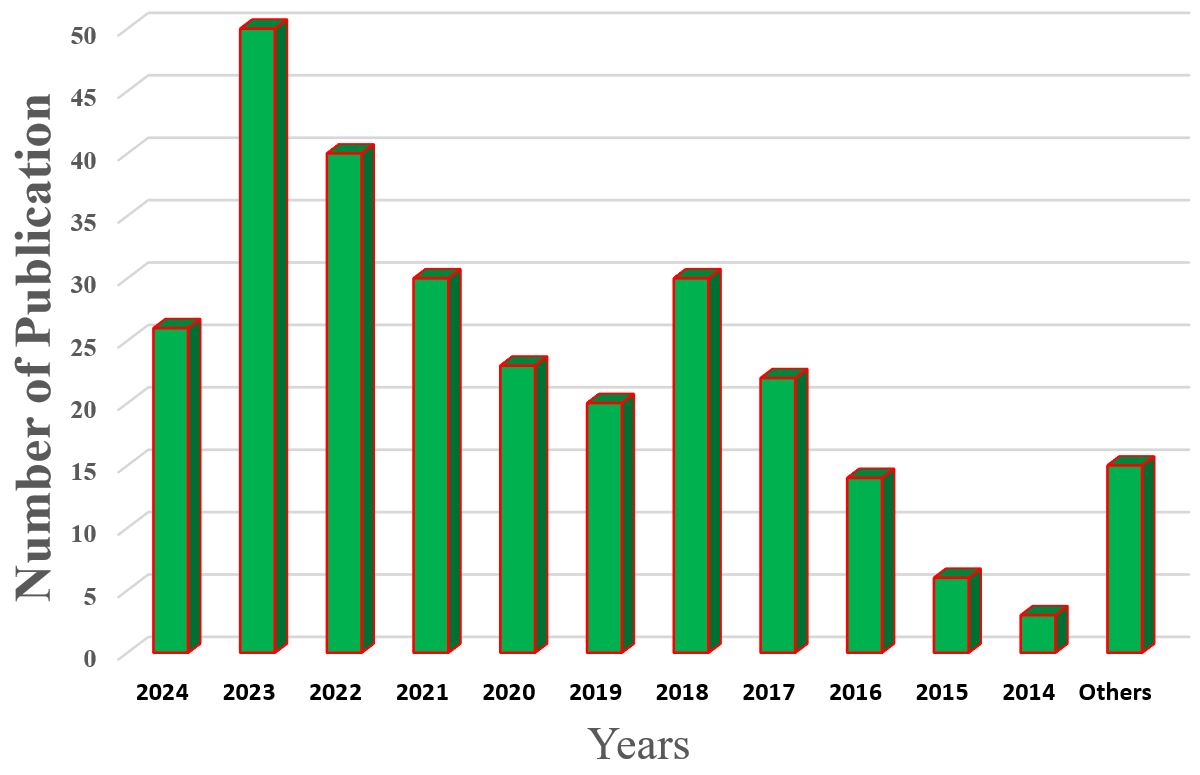}
    \caption{Year-wise peer-reviewed publications used in the study.}
    \label{fig:year_wise_reference}
\end{figure}
 \begin{figure}[htp]
    \centering
    \includegraphics[width=9cm]{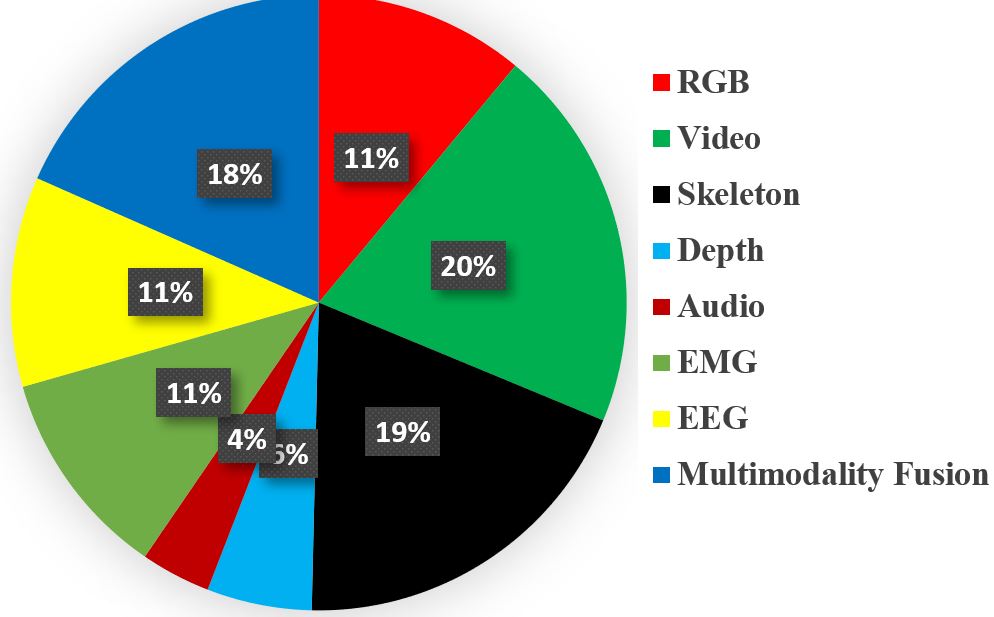}
    \caption{Various data modalities based references.}
    \label{fig:data_modalities_reference}
\end{figure}

\subsection{Article Search and Survey Methodology}\label{sec3.1}
To identify relevant articles on multimodality-based HGR, we employed a targeted search strategy using specific keywords. The focus was on the following terms:
\begin{itemize}
\item Vision and sensor-based recognition of static and dynamic hand gestures and sign language
\item Skeleton-based HGR and SLR
\item Multimodal dataset fusion-based HGR and SLR
\end{itemize}

We sourced articles from esteemed databases to ensure a comprehensive review of pertinent literature. The databases included IEEE Xplore Digital Library, MDPI, ScienceDirect, Springer Link, ResearchGate, and Google Scholar.

To refine and ensure relevance in our initial search results, we applied the following criteria:
\begin{itemize}
\item Publication date between 2014 and 2024
\item Inclusion of journals, proceedings, and book chapters
\item Focus on various data modality-based HGR systems, including RGB image, video, skeleton, depth, EMG, EEG, audio, and multimodal fusion modalities
\item Exclusion of papers where HGR in SLR are only mentioned tangentially
\item Exclusion of literature primarily consisting of reviews of other researchers' work
\item Exclusion of studies lacking in-depth information about their experimental procedures
\item Exclusion of research articles where the complete text isn't accessible, both in physical and digital formats
\item Exclusion of research articles that include opinions, keynote speeches, discussions, editorials, tutorials, remarks, introductions, viewpoints, and slide presentations
\end{itemize}

Using the search keywords outlined in this methodology, we identified 200 articles that met our inclusion and exclusion criteria. 
Figure \ref{fig:article_slection_pro} demonstrates the article selection procedure. 

Figures \ref{fig:intro_reference}, \ref{fig:year_wise_reference}, and \ref{fig:data_modalities_reference} show the types of references, year-wise references, and data modalities based on the number of references, respectively.
In the survey methodology, we reviewed each article through a process involving abstract review, methodology analysis, discussion, and result evaluations. Most of the papers were sourced from the IEEE Xplore Digital Library, ensuring a high standard of research quality. Different modalities used in HGR (HGR) have unique features, each with its own set of advantages and disadvantages, as summarized in Table \ref{tab:modality_pros_cons}.


\subsection{Research Gap and New Research Challenges}
While existing reviews have provided comprehensive overviews of HGR (HGR) research, there remains a significant gap in the literature focusing specifically on the advancements and potential avenues for multimodality-based HGR systems. Prior studies have largely overlooked the unique challenges and future directions necessary for developing robust and efficient multimodal HGR systems.
Our research aims to address this gap by thoroughly reviewing the existing literature to identify the progress made in multimodal-based HGR systems. By exploring the trajectory of multimodal HGR research, we aim to provide valuable insights and guidance for future research directions.

\subsection{Contribution}
\begin{figure*}[htp]
    \centering
    \includegraphics[width=18cm]{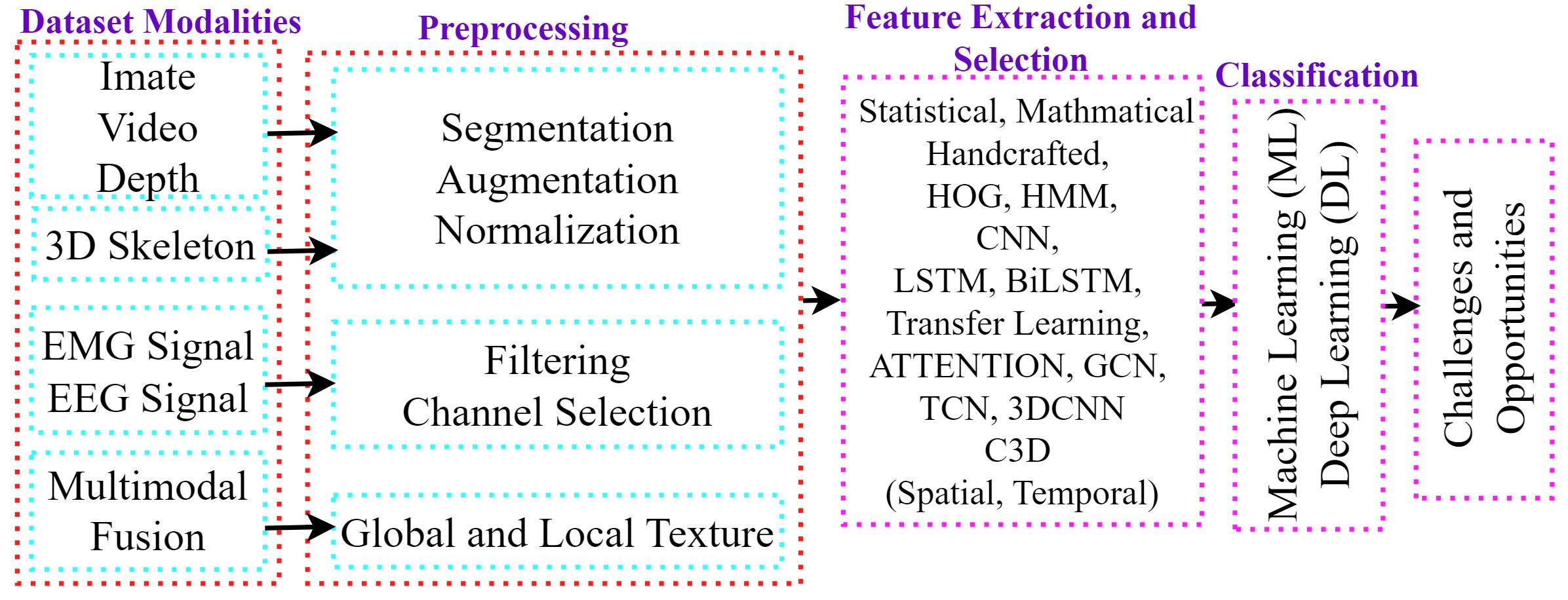}
    \caption{The basic flowgraph of the HGR research work.}
    \label{fig:basic_hgr_recognition_model}
\end{figure*}

Figure \ref{fig:basic_hgr_recognition_model} demonstrated the proposed methodology flowgraph, which was basically followed by HGR researchers. Figure \ref{fig:structure} demonstrated the structure of the study. The main contributions of this research are given below:
\begin{itemize}
    \item 
    Comprehensive Review: We provide an extensive review of HGR systems, focusing on the evolution of data acquisition, data environments, and hand gesture portrayals from 2014 to 2024.
    \item 
    Multimodal Analysis: Our study is the first to systematically examine the advancements in various data modalities based on HGR systems, which cover RGB, skeleton, depth, audio, EMG, EEG, and multimodal fusion.
    \item 
    Identification of Gaps and Future Directions: We created the datasets table for each modality with the latest performance beside the performance table. Then, we identify significant gaps in current research and propose potential future research directions.
    \item 
    Evaluation of System Efficacy: We assess the effectiveness of existing HGR systems by analyzing their recognition accuracy, providing a benchmark for future developments.
    \item 
    Guidance for Practitioners: The insights and findings from our review offer practical guidance for researchers and practitioners aiming to develop more robust and accurate HGR systems. Our comprehensive survey study will serve as a valuable resource for researchers and practitioners in the field. It will offer insights into the current state-of-the-art techniques, highlight the challenges, and suggest potential future research directions. This will facilitate informed decision-making and advance the development of innovative HGR systems.

\end{itemize}
By addressing these contributions, our paper not only fills existing gaps in the literature but also paves the way for future advancements in the field of HGR.  

\subsection{Research Question and Our Contribution}
To achieve this, we have formulated two primary research questions:
\begin{itemize}
    \item 
\textbf{Research Question 1 (RQ1):}  How have vision, sensor-based, and multimodal HGR systems evolved in areas such as data acquisition, data environment, and hand gesture portrayals from 2014 to 2024? \\\
To answer RQ1, we have thoroughly examined articles on the various data modalities used in HGR during this period. This examination aims to pinpoint the challenges faced and the solutions proposed in the field. Our findings indicate significant advancements in sensor technology, improved data acquisition techniques, and richer datasets that have enhanced the portrayal of hand gestures.
\item 
\textbf{Research Question 2 (RQ2): }How effective have the current various data modality-based HGR systems been, and what could be the prospective trajectories in this domain?  \\
To address RQ2, we have scrutinized the efficacy of the various data modality-based gesture recognition systems, particularly focusing on recognition accuracy. This analysis will help shed light on the current effectiveness and potential future directions of these systems.
\end{itemize}

\begin{figure*}[htp]
    \centering
    \includegraphics[width=18cm]{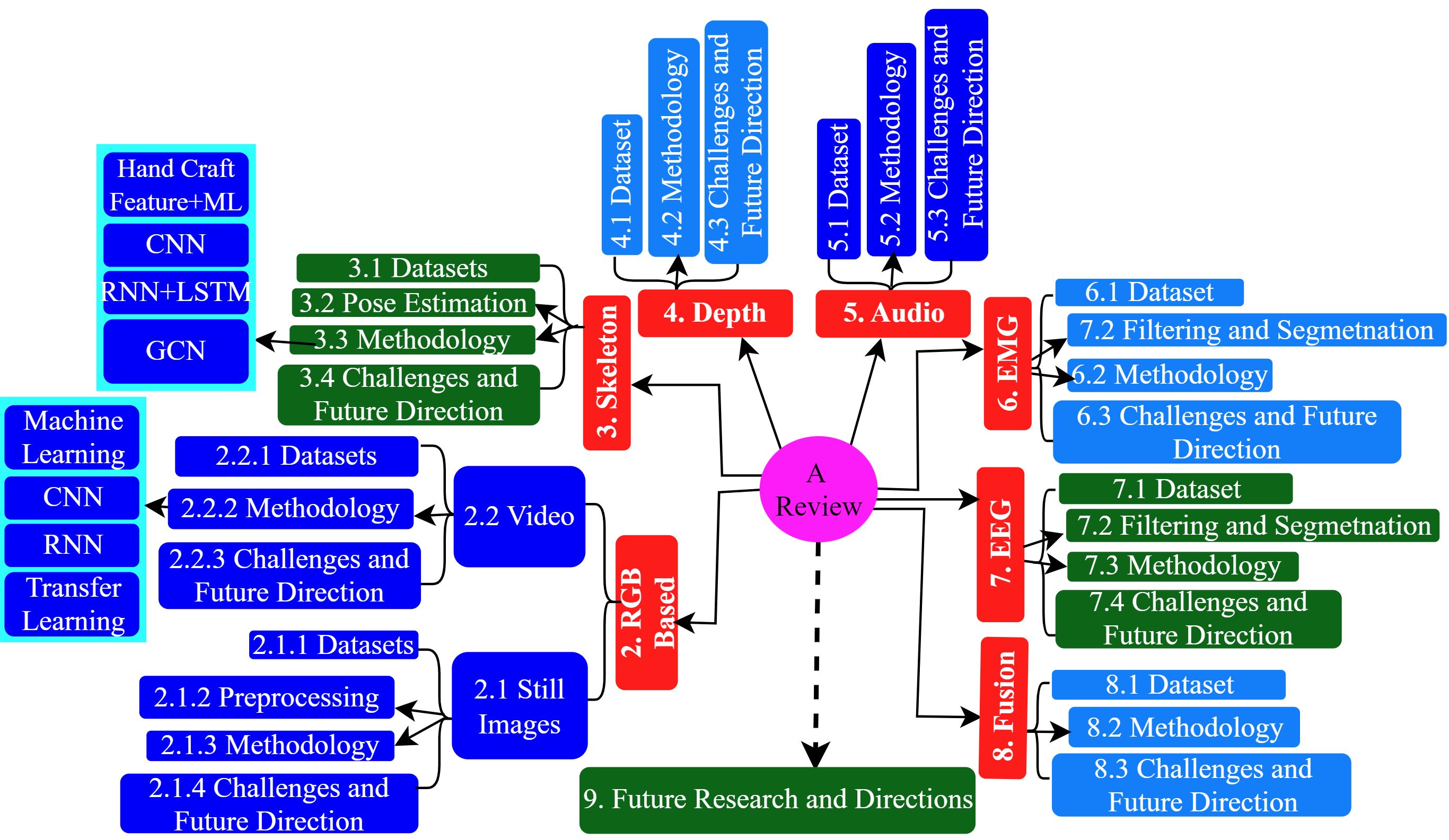}
    \caption{Structure of this paper.}
    \label{fig:structure}
\end{figure*}

\begin{table*}[ht]
\begin{adjustwidth}{-0cm}{0cm}
\setlength{\tabcolsep}{5pt}
\centering
\caption{Characteristics, advantages and disadvantages of different modalities.} \label{tab:modality_pros_cons}

\begin{tabular}{|c|c|c|c|}
\hline
\textbf{Modality}           & \textbf{Example}                  & \textbf{Pros}                                                                                                                                                                                 & \textbf{Cons}                                                                                                            \\ \hline
RGB   Single image  & \begin{tabular}[c]{@{}c@{}}Hand Gesture still image \cite{miah2024hand_multi}.\\ {\includegraphics[width=.12\textwidth, height=.12\textwidth]{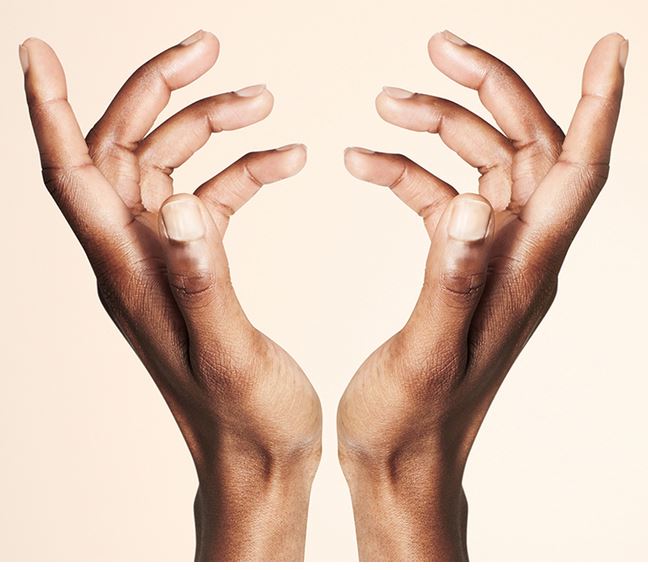}} \end{tabular} 
& \begin{tabular}[c]{@{}c@{}}Provide static data \\from static images.\\ Provide color distribution, \\color dominant, overall texture\end{tabular}                                                   & \begin{tabular}[c]{@{}c@{}} Only be able to look \\ at static information   \end{tabular}                                                                            \\ \hline
RGB   Video       & \begin{tabular}[c]{@{}c@{}}Hand Gesture \cite{zerrouki2024deep}.\\ {\includegraphics[width=.12\textwidth, height=.12\textwidth]{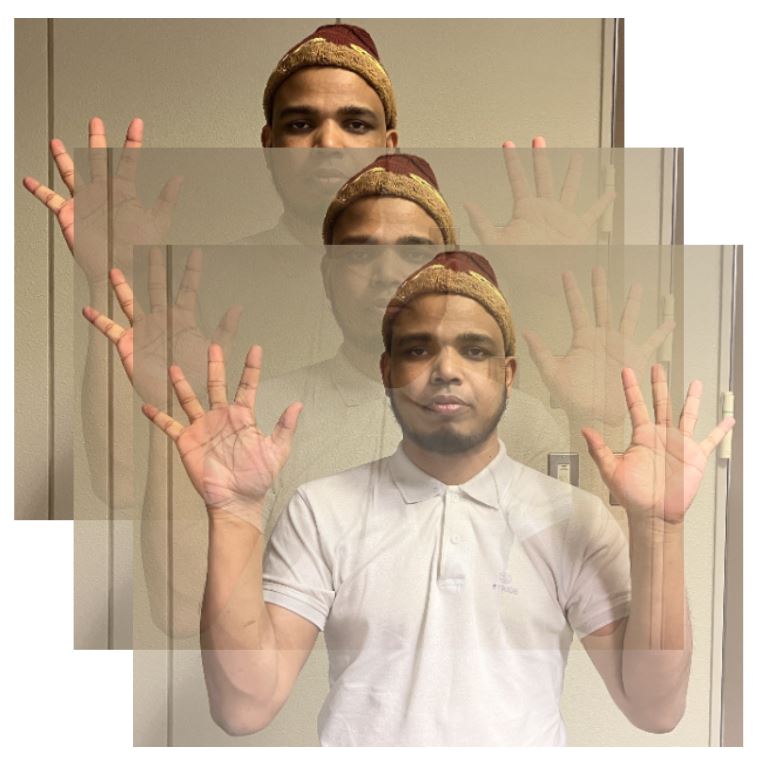}} \end{tabular}              & \begin{tabular}[c]{@{}c@{}}Provide motion information. \\ Give details about the rich appearance.\\ Simple to use and \\obtain numerous uses\end{tabular}                                         & \begin{tabular}[c]{@{}c@{}}Receptive to opinions.\\ Sensitive to context. \\Light sensitivity.\end{tabular}                \\ \hline
3D   Skelton                & \begin{tabular}[c]{@{}c@{}}Hand palm skeleton \cite{miah2023dynamic}.\\ {\includegraphics[width=.12\textwidth, height=.12\textwidth]{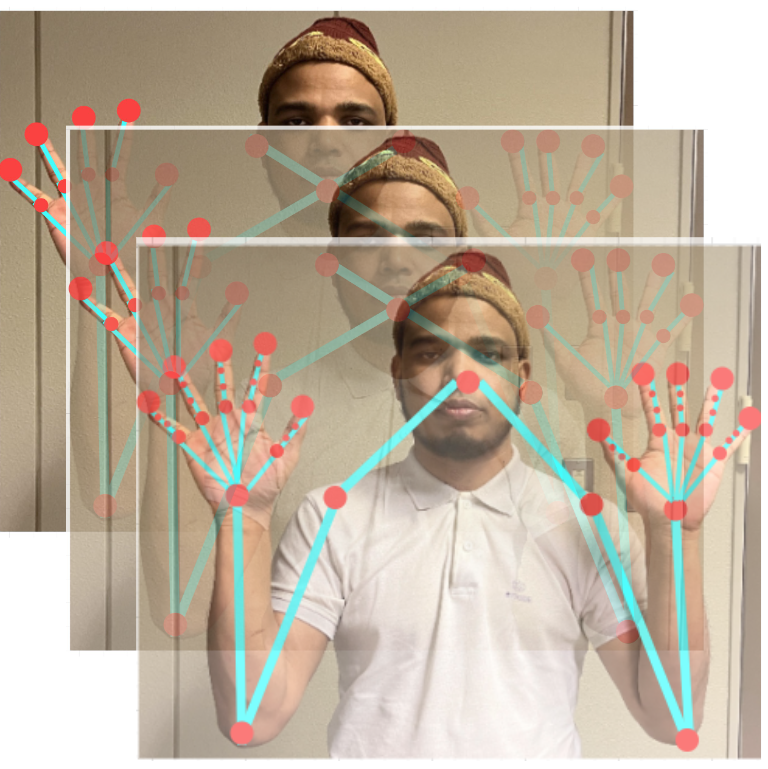}} \end{tabular}        & \begin{tabular}[c]{@{}c@{}}Provide the subject’s \\pose’s 3D structural information.\\ straightforward but instructive.\\ Inconsiderate of the perspective.\\ Disregard for the context.\end{tabular} & \begin{tabular}[c]{@{}c@{}}Absence of details \\regarding appearance.\\ Absence of comprehensive \\shape data.\end{tabular}  \\ \hline
Depth                       & \begin{tabular}[c]{@{}c@{}}Mopping floor \cite{chang2023exploration}.\\ {\includegraphics[width=.12\textwidth, height=.12\textwidth]{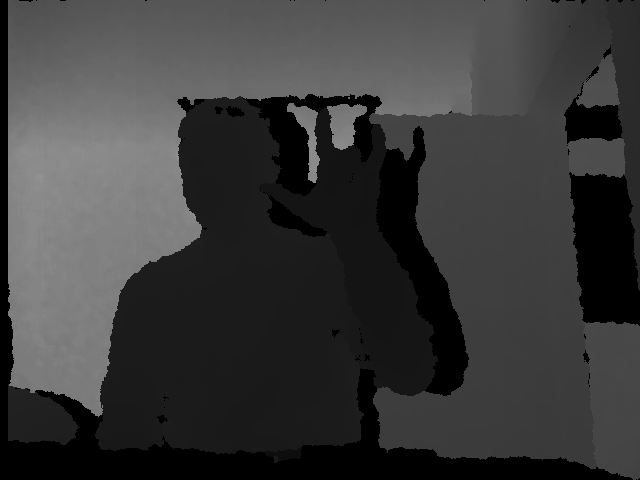}} \end{tabular}            &  \begin{tabular}[c]{@{}c@{}}Provide details on the \\3D structure. Give details\\ about geometric shapes.    \end{tabular}                                                                                                                  & \begin{tabular}[c]{@{}c@{}}Insufficient details about \\texture and colour.\\ Restricted practical distance.\end{tabular}   \\ \hline
EMG   Signal                & \begin{tabular}[c]{@{}c@{}}EMG Hand Gesture \cite{vasconez2023comparison}.\\ {\includegraphics[width=.12\textwidth, height=.12\textwidth]{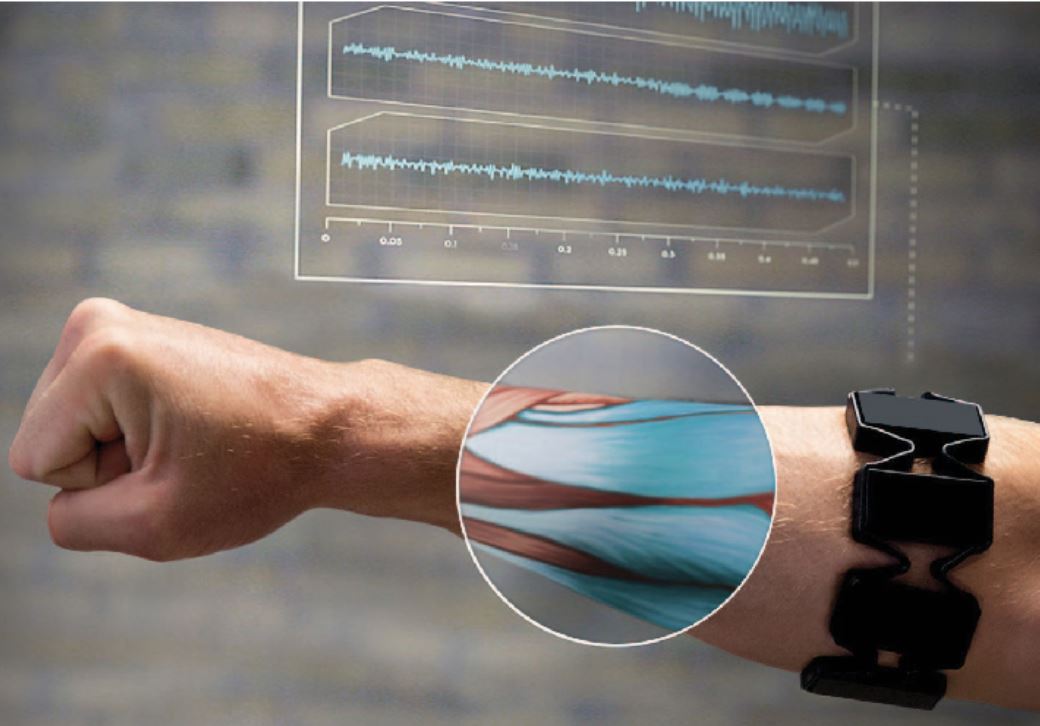}} \end{tabular}         & \begin{tabular}[c]{@{}c@{}}EMG provides muscle \\contractor information.\\ No need for light \\color or background.\end{tabular}                                                                    & \begin{tabular}[c]{@{}c@{}}Difficult to control \\muscle power.\\ Sometimes appropriate\\ information is costly\end{tabular} \\ \hline
EEG   Signal                & \begin{tabular}[c]{@{}c@{}}EEG Hand gesture \cite{BaronaLopez2024}.\\ {\includegraphics[width=.12\textwidth, height=.12\textwidth]{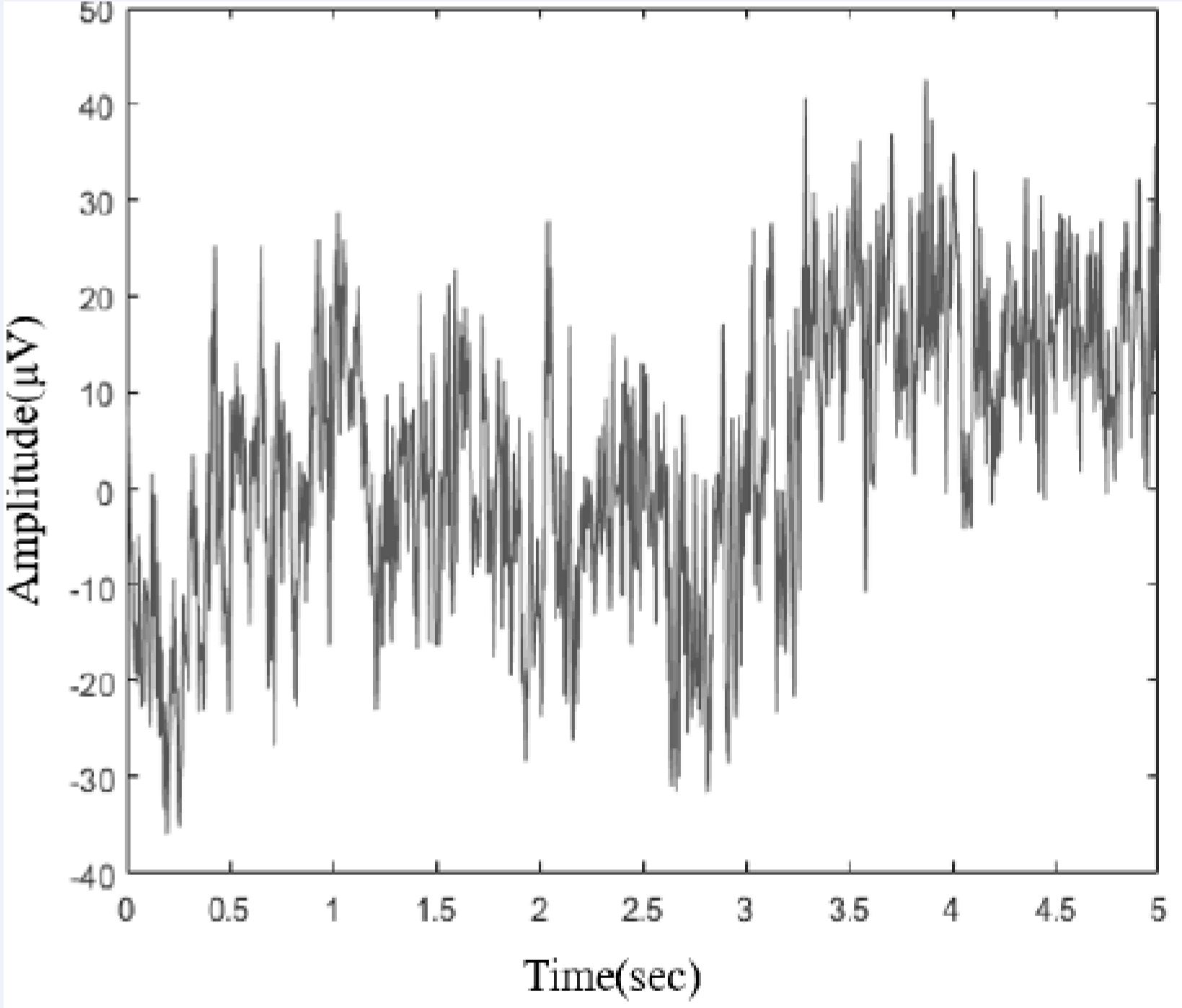}} \end{tabular}          & \begin{tabular}[c]{@{}c@{}}EEG provide muscle \\contractor information.\\ No need for light \\color or background.\end{tabular}                                                                     & \begin{tabular}[c]{@{}c@{}}Difficult to control\\ muscle power.\\ No need for light color or background\end{tabular}       \\ \hline
Audio                       & \begin{tabular}[c]{@{}c@{}}Audio wave of jumping \cite{ling2020ultragesture}.\\ {\includegraphics[width=.12\textwidth, height=.12\textwidth]{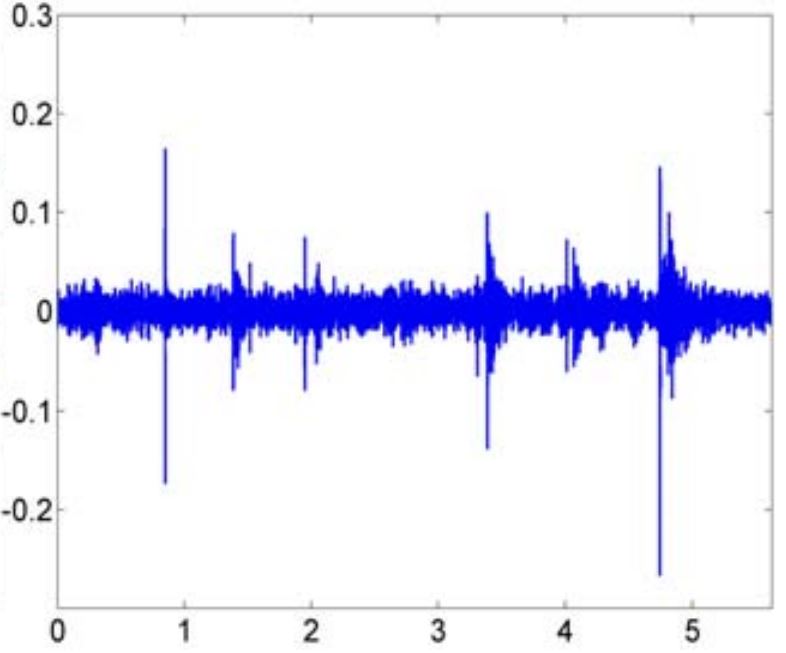}} \end{tabular}    &  \begin{tabular}[c]{@{}c@{}} Easy to locate \\actions in temporal sequence   \end{tabular}                                                                                            & \begin{tabular}[c]{@{}c@{}}Lack of appearance \\information    \end{tabular}                                                                                         \\ \hline
\end{tabular}
\end{adjustwidth}
\end{table*}

\subsection{Organization of the Study}
The rest of the paper is organized as follows: Section \ref{sec1} RGB-modality-based static and dynamic HGR work. Section \ref{sect2} described the skeleton-modality based HGR. Section \ref{sect3} described the depth data modalities based on HGR. Section \ref{sec4} describes the EMG-modality-based HGR. Section \ref{sect5} demonstrated the EEG signal-modality-based HGR, Section \ref{sec6} describes the audio signal-modality-based HGR, and Section \ref{sect7} demonstrates the multimodal fusion-based HGR. Conclusion described in the Section \ref{sect8}.


\section{RGB-Modal Based HGR}\label{sec1}
RGB modal-based HGR involves processing still images or videos to identify static and dynamic gesture recognition, respectively. RGB dataset, mainly collected using cameras, is relatively simple to gather and provides detailed appearance information about the scene context \cite{bilen2017action, soomro2015action, jain2023addsl, guney2022real,trujillo2023mexican, simchowitz2023language} 
An example of this is the gesture for the word "WELCOME" in Chinese Sign Language \cite{zhang2023multimodal}. 
RGB-based data modalities can be either static or dynamic, where a single image can express the full meaning of the gesture and sequence of frames needed to completely express the dynamic gesture \cite{zhang2023multimodal, aloysius2020understanding, amin2023assistive,rehman2023real}. Mobile cameras \cite{takayama2018sign, athira2022signer}, web cameras \cite{islam2018hand}, and specialized cameras such as HP Pavilion dv6 \cite{mahmood2018dynamic} are mainly used to record the dataset \cite{tao2018american}. RGB data modality-based SL datasets have become popular due to their portability and cost-effectiveness over sensor-based systems \cite{miah2022bensignnet}. Below, we included static and dynamic HGR using vision-based data modalities.

\subsection{RGB Still Image}
Static images belong to the RGB modality, where a single image can express the full meaning of the gesture, and this data modality is usually recorded by the camera. Figure \ref{fig:basic_hgr_recognition_model} demonstrated the overview of the still image-based HGR steps, which mainly included preprocessing, feature extraction, and classification. 


\subsubsection{Dataset}
Benchmark dataset is crucial for developing and evaluating the static HGR systems as demonstrated in Table \ref{tab:rgb_stil_image_dataset}. Table \ref{tab:rgb_stil_image_dataset} included various benchmark datasets' names of this modality, creation year, number of classes, dataset types, sample size and latest performance accuracy. The key datasets include KSL-77, which includes 77 classes containing 112564 frames from 20 individuals \cite{yang2020korean}, and ASL-10, featuring ten gestures from 14 individuals, each with ten instances \cite{marin2016hand}. The ASL-20 dataset comprises 20 ASL words with 18,000 frames \cite{marin2016hand}, while the dataset created by Islam et al. consists of 26 signs from three persons, totalling 9,360 images \cite{islam2018hand}. BSL-38 includes 38 classes with 320 images per class, created with both deaf and non-deaf students \cite{rafi2019image}. The Irish Sign Language (ISL) dataset contains 58,114 images for 23 common hand shapes from ISL \cite{oliveira2017dataset}. The ISL dataset includes 720 images, with 600 for testing and 120 for training \cite{dixit2013automatic}. The JSL Word Dataset features videos of JSL word motions, with frames converted to grayscale and analyzed \cite{ito2020japanese}. Lastly, the LSA64 dataset, designed for Argentinian sign language research, comprises recordings of 64 distinct signs performed by 10 individuals, standardized to 48 frames per video \cite{konstantinidis2018sign}. These datasets provide a comprehensive foundation for advancing HGR technology.

\begin{table*}[ht]
\caption{RGB still image modality based dataset description.} \label{tab:rgb_stil_image_dataset}
\setlength{\tabcolsep}{5pt}
\begin{tabular}{|l|l|l|l|l|l|l|l|}
\hline
\begin{tabular}[c]{@{}l@{}}Dataset\\ Names\end{tabular} & Year & \begin{tabular}[c]{@{}l@{}}Dataset\\ Types\end{tabular} & Classes & Subject & \begin{tabular}[c]{@{}l@{}}Total \\ Sample \end{tabular} & \begin{tabular}[c]{@{}l@{}}Sample \\ Sign \end{tabular} & \begin{tabular}[c]{@{}l@{}}Latest \\ Performance\\ Accuracy \end{tabular} \\
\hline
OUHAND & 2016 & \begin{tabular}[c]{@{}l@{}} ASL \end{tabular} & 10 & 23 & 3000 & - & \begin{tabular}[c]{@{}l@{}} 98.56 \cite{zhang2023lightweight} \end{tabular} \\
\hline
\begin{tabular}[c]{@{}l@{}}NTU\\  Dataset\end{tabular} & 2018 & Digits & 10 & 10 & 1000 & 100 & \begin{tabular}[c]{@{}l@{}}99.00 \cite{miah2024hand_multi} \end{tabular} \\
\hline
ASL-10 & 2020 & SL & 10 & 22 & 2800 & 120 & 95.00 \cite{miah2024hand_multi} \\
\hline
MUD & 2021 & \begin{tabular}[c]{@{}l@{}} ASL \end{tabular} & 36 & - & 2520 & 70 & \begin{tabular}[c]{@{}l@{}} 99.23 \cite{aurangzeb2024deep} \\ \end{tabular} \\
\hline
ASLAD & 2021 & \begin{tabular}[c]{@{}l@{}} ASL \end{tabular} & 29 & - & 87000 & 3000 & \begin{tabular}[c]{@{}l@{}} 99.00 \cite{aurangzeb2024deep} \\ \end{tabular} \\
\hline
NUS II & 2021 & \begin{tabular}[c]{@{}l@{}} ASL \end{tabular} & 10 & 40 & 2000 & - & \begin{tabular}[c]{@{}l@{}} 96.5 \cite{eid2023visual} \\ \end{tabular} \\
\hline
Marcel & 2021 & \begin{tabular}[c]{@{}l@{}} ASL \end{tabular} & 6 & - & 5531 & - & \begin{tabular}[c]{@{}l@{}} 96.57 \cite{eid2023visual} \end{tabular} \\
\hline
HGR1 & 2021 & \begin{tabular}[c]{@{}l@{}} ASL \end{tabular} & 25 & 12 & 899 & - & \begin{tabular}[c]{@{}l@{}} 93.36 \cite{zhang2023lightweight} \end{tabular} \\
\hline
ArSL \cite{aly2020deeparslr} & 2022 & SL & 23 & - & - & - & 95.00 \cite{shin2024japanese_jsl1} \\
\hline
ASL-20 \cite{shin2023rotation} & 2022 & SL & 20 & 5 & 18000 & 900 & 99.00 \\
\hline
KSL-77 \cite{yang2020korean} & 2023 & SL & 77 & 22 & 112,564 & 1461 & 99.00 \\
\hline
KSL-20 \cite{shin2023korean} & 2023 & SL & 20 & 25 & 96200 & 4800 & 99.00 \\
\hline
BSL \cite{rafi2019image} & 2023 & SL & 38 & 35 & 12160 & 320 & 96.00 \\
\hline
JSL \cite{shin2024japanese_jsl1} & 2023 & SL & 41 & 20 & 7380 & 1800 & 91.00 \cite{shin2024japanese_jsl1} \\
\hline
\begin{tabular}[c]{@{}l@{}}Creative \\ Senz3d\end{tabular} & 2023 & SL & 11 & 10 & 1320 & 300 & 98.51 \cite{sharma2024spatiotemporal} \\
\hline
FMCW-SAR \cite{hao2023static} & 2023 & \begin{tabular}[c]{@{}l@{}} HGR \end{tabular} & 5 & Simulated & - & - & \begin{tabular}[c]{@{}l@{}} 97.00 \cite{hao2023static} \end{tabular} \\
\hline
Custom ASL \cite{byberi2023glovesense} & 2023 & \begin{tabular}[c]{@{}l@{}} HGR \end{tabular} & 10 & 7 & 1400 & - & \begin{tabular}[c]{@{}l@{}} 99.76 \end{tabular} \\
\hline
\end{tabular}
\end{table*}

\subsubsection{Methodology of the RGB Still Image Modality}
\begin{figure}[htp]
    \centering
    \includegraphics[width=9cm]{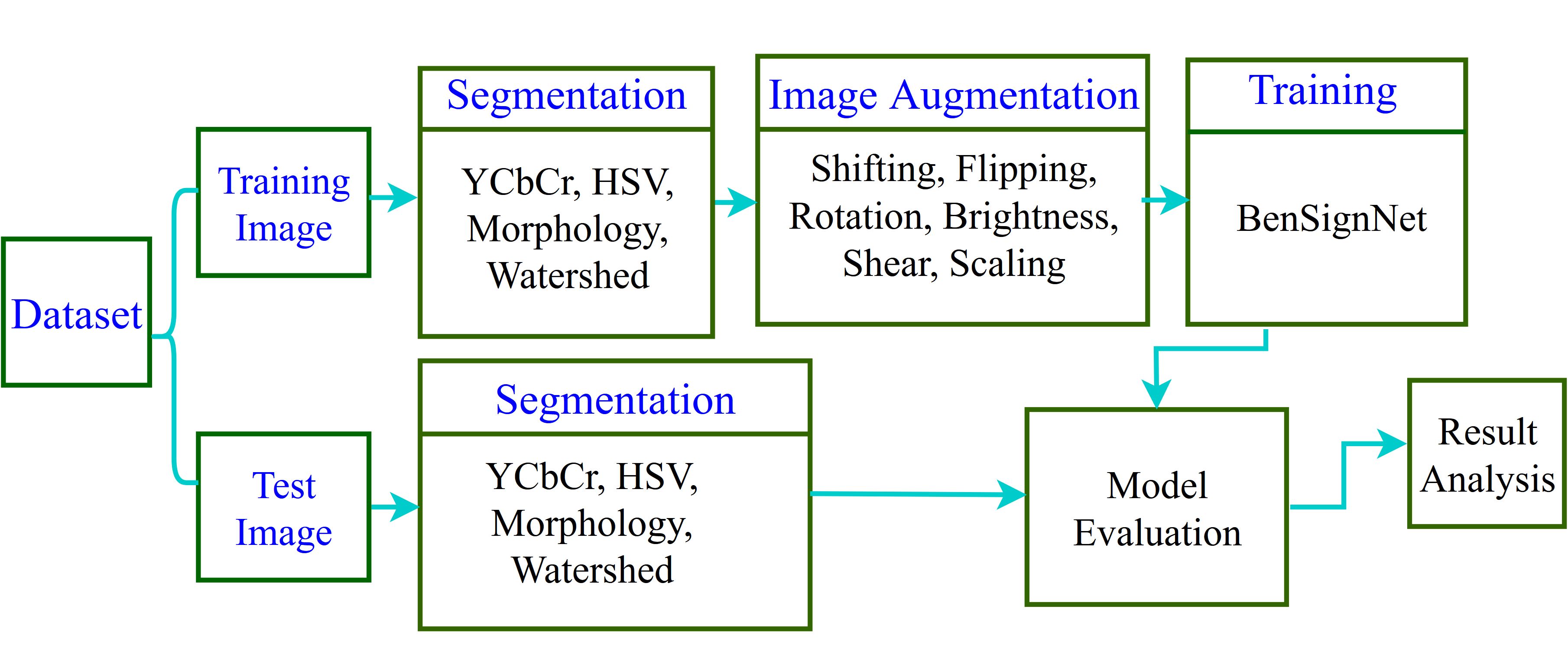}
    \caption{Working procedure of the RGB still-image-based gesture recognition\cite{miah2022bensignnet} }
    \label{fig:rgb_bensignnet}
\end{figure}

Figure \ref{fig:basic_hgr_recognition_model} demonstrated a common workflow diagram of the RGB still-image-based HGR. Table \ref{tab:performance_rgb_stilimage} demonstrated the performance of this data modality, including year, feature extraction and classification method performance. The preprocessing, feature extraction, machine learning (ML) and diverse explanations of the DL system are included below: 

\paragraph{Preprocessing}
Image segmentation and augmentation, including rotation, translation, and scaling, are vital preprocessing techniques. Segmentation enhances image relevance by dividing them into multiple segments or regions and labelling each pixel to outline objects and edges. For instance, Figure \ref{fig:rgb_bensignnet} demonstrates an RGB still-image-based HGR system for BSL, which utilizes segmentation and augmentation techniques \cite{miah2022bensignnet}. While skin-color models are commonly used to differentiate hand motions from the background, they may struggle with objects of similar skin tones, such as faces. Segmentation is further improved by training classifiers to distinguish hand regions from non-hand areas using attributes from extensive datasets. CNN-based segmentation methods, particularly Fully Convolutional Networks (FCNs), are increasingly popular. FCNs optimize motion segmentation by employing deconvolution layers and upsampling images to their original size through pixel prediction, handling images of any size without requiring uniform dimensions. Despite challenges like occlusion and varying light conditions, CNN-based segmentation effectively addresses these issues, enabling robust gesture segmentation.
 
\paragraph{Hand Crafted Feature and ML Approach}
Researchers have extensively employed hand-crafted feature extraction coupled with ML algorithms for SLR systems \cite{tao2018american, kwok2002real}. Various techniques, such as the HMM and Pattern Trees (SP-Tree), have been utilized by researchers to develop static HGR recognition. For instance, Ren et al. achieved 93.00\% accuracy for Greek Sign Language (GSL) using the Hidden Markov Models (HMM) approach, and Ong et al. reported 88.00\% accuracy for German Sign Language (GSL) by using SP-Tree method \cite{ren2013robust}. Additionally, Linear Discriminant Analysis (LDA), k-nearest Neighbors (KNN), and Random Decision Forest (RDF) have shown efficiency across various SL datasets \cite{kwok2002real}
Moreover, several prominent techniques have been utilized for this purpose, including histogram of oriented gradient (HOG), CNN, and PCA. An interesting approach described in \cite{dixit2013automatic} creates a unique feature vector by combining the Hu invariant moment with structural shape descriptors. This feature vector is then derived from the input image during the testing phase, following the application of preprocessing techniques. In another study presented in \cite{ito2020japanese}, CNNs are employed to extract features from the images collected for JSL word recognition.

Takayama et al. \cite{takayama2018sign} utilized HMM for word classification, effectively managing time and amplitude variances in time series signals. They designed left-to-right HMM models with 5 states for pauses and 20 states for each sign word. Athira et al. \cite{athira2022signer} employed an SVM for SLR, developing three models for Zernike moments, trajectory-based recognition, and shape-based recognition. The SVM used a multi-class C-SVC with a radial basis function kernel and a one-against-all strategy. Oliveira et al. \cite{oliveira2017dataset} used Principal Component Analysis (PCA) for Irish Sign Language (ISL) recognition. Ibrahim et al. \cite{ibrahim2018automatic} employed the Euclidean distance classifier for Arabic Sign Language (ArSL), achieving a 97.00\% recognition rate. Dixit et al. \cite{dixit2013automatic} used a multi-class Support Vector Machine (MSVM) for ISL recognition, reporting a 96.00\% accuracy. Ito et al. \cite{ito2020japanese} used SVMs for JSL classification, achieving mean accuracy rates of 99.2\%, 94.3\%, and 86.2\% for different numbers of JSL words. Islam et al. \cite{islam2018potent} reported a training model accuracy of approximately 95.00\% for recognizing BSL digits.
\begin{figure}[htp]
    \centering
    \includegraphics[width=9cm]{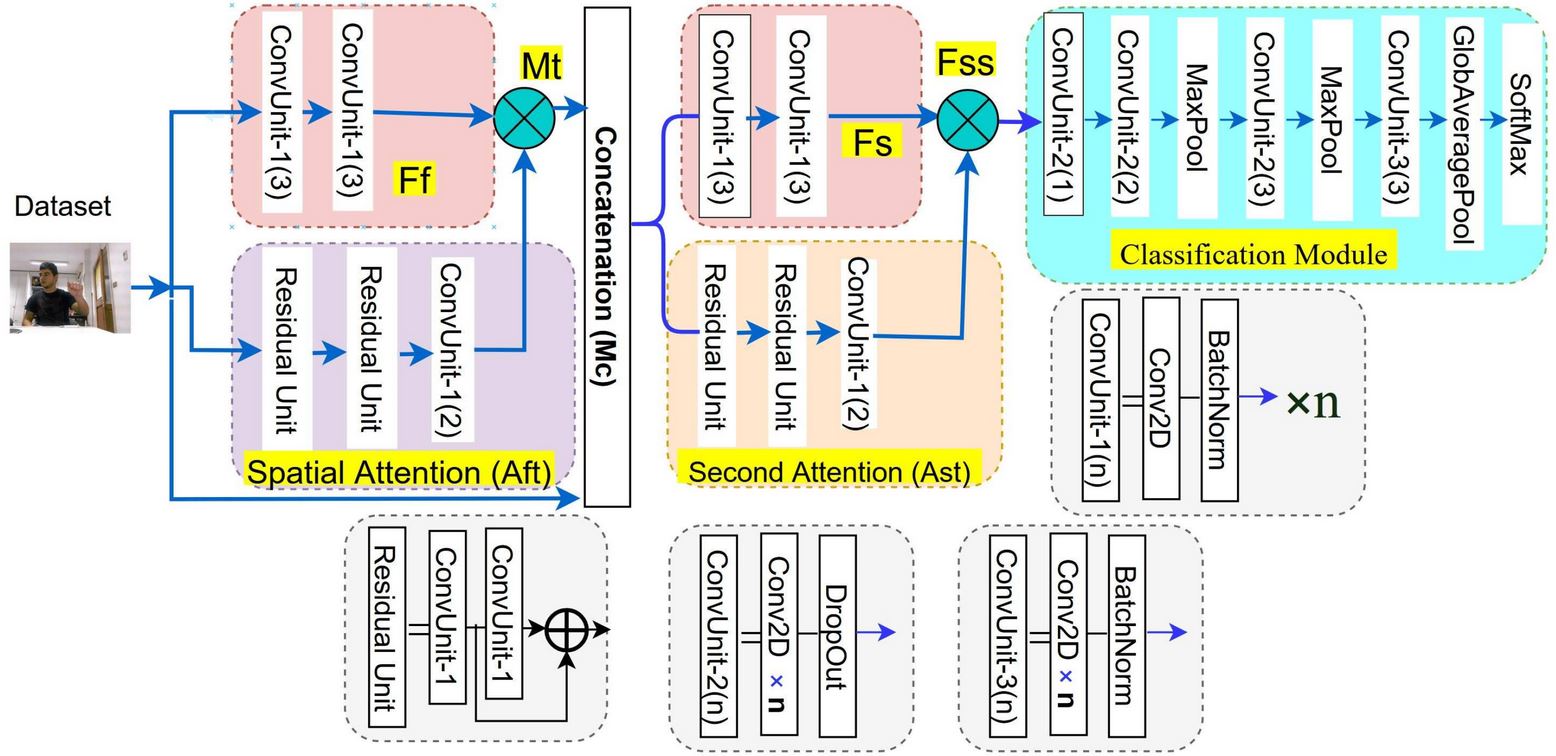}
    \caption{ RGB still-image based gesture recognition using multi-stage DL module\cite{computers12010013_multistage_musa} }
    \label{fig:rgb_multistage}
\end{figure}

\begin{table*}[ht]
\setlength{\tabcolsep}{8pt}
\caption{Databases for RGB Still Image Modality.} \label{tab:performance_rgb_stilimage}
\begin{center}
\begin{tabular}{|l|l|l|l|l|l|l|l|}
\hline
Authors & Year & \begin{tabular}[c]{@{}l@{}}Supported \\ Language\end{tabular} & Classes & Sample & \begin{tabular}[c]{@{}l@{}}Feature\\  Extraction\end{tabular} & Classifier & Performance \\
\hline
Tao et. al \cite{tao2018american} & 2018 & American SL & 24 & 500-600 & CNN & CNN & 84.80 \\ \hline
Ibrahim et. al \cite{ibrahim2018automatic} & 2018 & Arabic SL & 30 & 450 & N/A & HMM & 97.00 \\ \hline
Ito et. al \cite{ito2020japanese} & 2020 & Japanese SL & 20 & 13200 & CNN & MSVM & 94.30 \\ \hline
Selda et. al \cite{guney2022real} & 2022 & Turkish SL & 32 & 19200 & CNN & LSTM & 99.75 \\ \hline
Musa et. al \cite{miah2022bensignnet} & 2022 & Bangla SL & 36 & 19200 & BenSignNet & Softmax & 99.00 \\ \hline
Wang et. al \cite{wang2023ultrasonicgs} & 2023 & China SL & 15 & 720 & CRNN & BiLSTM & 92.40 \\ \hline
Jain et. al \cite{jain2023addsl} & 2023 & Danish SL & 36 & 252 & YOLOv5 & CSP Darknet & 92.00 \\ \hline
Musa et. al \cite{computers12010013_multistage_musa} & 2023 & Hand Gesture/ASL & 11 & 11000 & Multi-stage CNN & CNN & 99.00 \\ \hline
Shin et. al \cite{shin2023korean} & 2023 & KSL & 77 & 20000 & Transformer & CNN & 89.00 \\ \hline
Noble et al. \cite{noble2023static} & 2023 & \begin{tabular}[c]{@{}c@{}}Custom\\ capacitive\\ sensing\\ dataset\end{tabular} & 5 & 5000 & -&\begin{tabular}[c]{@{}c@{}}Decision Tree, \\ Naïve Bayes, \\ MLP, CNN\end{tabular} & \begin{tabular}[c]{@{}c@{}}MLP: 96.87\\ CNN: 95.94 \\ DT: 91.18 \\ NB: 88.34\end{tabular} \\ \hline
Sahoo et al. \cite{sahoo2023hand} & 2023 & \begin{tabular}[c]{@{}c@{}} I: Custom, \\  II: ASL-FS\end{tabular} & 6, 24 & \begin{tabular}[c]{@{}c@{}}  15,000\\ (Dataset I), \\60,000+\\ (Dataset II)\end{tabular} & DRCAM &Softmax& \begin{tabular}[c]{@{}c@{}}95.09 \\ 93.44\end{tabular} \\ \hline
Baptista et al. \cite{baptista2023domain} & 2023 & HGR & 4 & \begin{tabular}[c]{@{}c@{}}Tr:24611\\ T:17051\end{tabular} & SMLT & Softmax & \begin{tabular}[c]{@{}c@{}}SLT: 84.23 \\ MSMLT: 79.24 \\ SMLT: 90.03 \end{tabular} \\ \hline
\begin{tabular}[c]{@{}c@{}} Mohyuddin\\ et al. \cite{mohyuddin2023comprehensive} \end{tabular} & 2023 & Leap motion dataset & 8 & 264 per & SSC-DNN & Softmax & 98.7\% \\ \hline
Garg et al. \cite{garg2023multiscaled} & 2023 & \begin{tabular}[c]{@{}c@{}}NVGesture, \\ Briareo\end{tabular} & - & - & MsMHA-VTN & Softmax & \begin{tabular}[c]{@{}c@{}}88.22\\ 99.10\end{tabular} \\ \hline
Bhartiet al. \cite{bharti2023novel} & 2023 & MINDS Libras & - & - & Key frame & 3D CNN & 98.00 \\ \hline
Eid et al. \cite{eid2023visual} & 2023 & \begin{tabular}[c]{@{}c@{}}NUS II\\ Marcel\end{tabular} & \begin{tabular}[c]{@{}c@{}}10 \\ 6\end{tabular} & \begin{tabular}[c]{@{}c@{}}2000 \\ 5531\end{tabular} & \begin{tabular}[c]{@{}c@{}}Skin Segmentation\\ Data Augmentation\end{tabular} & CNN & \begin{tabular}[c]{@{}c@{}}NUS II: 96.5, \\ Marcel: 96.57\end{tabular} \\ \hline
Zhang et al. \cite{zhang2023lightweight} & 2023 & \begin{tabular}[c]{@{}c@{}}HGR1, \\ OUHANDS\end{tabular} & \begin{tabular}[c]{@{}c@{}}25 \\ 10\end{tabular} & \begin{tabular}[c]{@{}c@{}}899 \\ 3,000\end{tabular} & \begin{tabular}[c]{@{}c@{}}BaseNet, \\ MSS,LAS\end{tabular} & LHGR-Net & \begin{tabular}[c]{@{}c@{}}93.36,  98.57\end{tabular} \\ \hline
Hao et al. \cite{hao2023static} & 2023 & \begin{tabular}[c]{@{}c@{}} FMCW-SAR\\ Imaging\end{tabular} & 5 & - & HOG-PCA & Random Forest & \begin{tabular}[c]{@{}c@{}}Unob:97 \\ Ob:93\end{tabular} \\ \hline
Byberi et al. \cite{byberi2023glovesense} & 2023 & Custom ASL & 10 & 1400 & Inductive Sensing & Random Forest & \begin{tabular}[c]{@{}c@{}}CV: 99.76\end{tabular} \\ \hline
Musa et. al \cite{miah2024hand_multi} & 2024 & SL & 10-77 & 1100-20000 & GmTC & CNN & 95.00 \\ \hline
\begin{tabular}[c]{@{}c@{}} Aurangzeb \\et al. \cite{aurangzeb2024deep} \end{tabular} & 2024 & \begin{tabular}[c]{@{}c@{}} MUD\\ ASLAD \end{tabular} & \begin{tabular}[c]{@{}c@{}} 36\\ 29\end{tabular} & \begin{tabular}[c]{@{}c@{}}2520 \\ 87,000 \end{tabular} & \begin{tabular}[c]{@{}c@{}}Image-\\based\\ features\end{tabular} & HVCNNM & \begin{tabular}[c]{@{}c@{}} 99.23, \\ 99.00\end{tabular} \\ \hline
Sharma et al. \cite{sharma2024spatiotemporal} & 2024 & \begin{tabular}[c]{@{}c@{}}Ego hand \\ ASL \\ Senz3D \end{tabular} & - & - & - & \begin{tabular}[c]{@{}c@{}}ResNet18 \\ DMD\end{tabular} & \begin{tabular}[c]{@{}c@{}}97.85 \\ 98.49 \\ 98.51 \end{tabular} \\ \hline
Bose et al. \cite{bose2024precise} & 2024 & \begin{tabular}[c]{@{}c@{}}NUSHP-II, \\ SENZ-3D, \\ MITI-HD\end{tabular} & - & - & - & \begin{tabular}[c]{@{}c@{}}Yolo-v2 (DarkNet-19), \\ Yolo-v3 (DarkNet-53)\end{tabular} & \begin{tabular}[c]{@{}c@{}} 99.10(MI-HD) \\  99.18(MI-HD)\end{tabular} \\ \hline
Alonazi \cite{alonazi2023smart} & 2024 & \begin{tabular}[c]{@{}c@{}}Egogesture, \\ Jester\end{tabular} & 15, 15 & \begin{tabular}[c]{@{}c@{}}2081 v, 148092 \end{tabular} & \begin{tabular}[c]{@{}c@{}}CNN, \\ Neural gas, \\ Thermal mapping\end{tabular} & DBN & \begin{tabular}[c]{@{}c@{}}90.73 \\ 89.33\end{tabular} \\ \hline
\begin{tabular}[c]{@{}c@{}} Montazerin \\et al. \cite{montazerin2023transformer} \end{tabular} & 2023 & \begin{tabular}[c]{@{}c@{}}Egogesture, \\ Jester\end{tabular} & 15, 27 & \begin{tabular}[c]{@{}c@{}}2081 RGB , \\ 148,092 \end{tabular} & \begin{tabular}[c]{@{}c@{}}CNN-based\\ detector \end{tabular} & DBN & \begin{tabular}[c]{@{}c@{}}90.73 \\ 89.33 \end{tabular} \\ \hline
\begin{tabular}[c]{@{}c@{}} Damaneh \\et al. \cite{damaneh2023static} \end{tabular} & 2024 & \begin{tabular}[c]{@{}c@{}}Massey \\ ASL Alphabet, \\ ASL\end{tabular} & - & \begin{tabular}[c]{@{}c@{}}2520 \\ 87,000 \\ 23,400\end{tabular} & - & \begin{tabular}[c]{@{}c@{}}CNN, Gabor filter, \\ ORB feature descriptor\end{tabular} & \begin{tabular}[c]{@{}c@{}}99.92 \\ 99.80 \\ 99.80\end{tabular} \\ \hline
\end{tabular}
\end{center}
\end{table*}

\begin{figure}[htp]
    \centering
    \includegraphics[width=9cm]{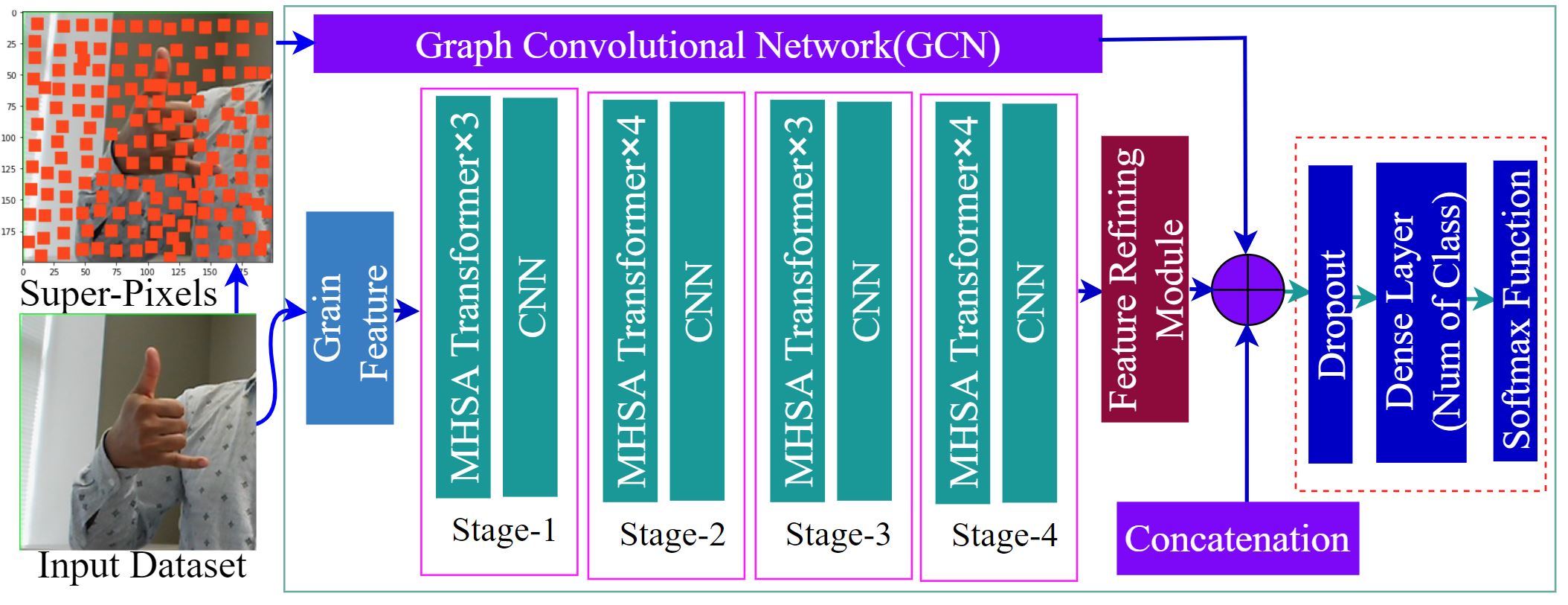}
    \caption{ RGB still-image based gesture recognition using GmTC module\cite{miah2024hand_multi} }
    \label{fig:rgb_GmTC}
\end{figure}

\paragraph{CNN-Based Methods}
Researchers focus on DL models for effective, generalized HGR with large-scale datasets that face difficulties in ML algorithms. Miah et al. utilized a CNN-based model, BenSignNet, after preprocessing with segmentation and augmentation, achieving impressive accuracy rates of 93.00\% for the BdSL38 dataset and 99.00\% for the ASL dataset \cite{miah2022bensignnet}. Similarly, DL models have shown significant improvements in recognition accuracy for Chinese and Arabic sign languages \cite{yuan2020hand, aly2020deeparslr}.
Wang et al. \cite{wang2023ultrasonicgs} discuss the CRNN (Convolutional Recurrent Neural Network) architecture, which combines convolutional layers for feature extraction and recurrent layers for sequence modelling. Then, the Bi-LSTM network captures semantic dependencies in both forward and backward directions and reported 98.80\% for single gestures.
Tao et al. \cite{tao2018american} applied a DL algorithm for RGB-based HGR, achieving an exceptional accuracy of 99.9\% with the half-half evaluation technique on 32,831 testing samples. The benchmark dataset's leave-one-out evaluations showed strong overall mean F-scores for the 24 indicators. The YOLOv5 model proposed by Jain et al. \cite{jain2023addsl} achieved a commendable accuracy of 92.00\% for Danish SLR.
To enhance feature effectiveness, Miah et al. \cite{computers12010013_multistage_musa} applied a multi-stage deep neural network, showing good performance accuracy for various hand gesture datasets. Figure \ref{fig:rgb_multistage} demonstrates the model architecture of each stage of this work.
\begin{figure}[htp]
    \centering
    \includegraphics[width=6cm]{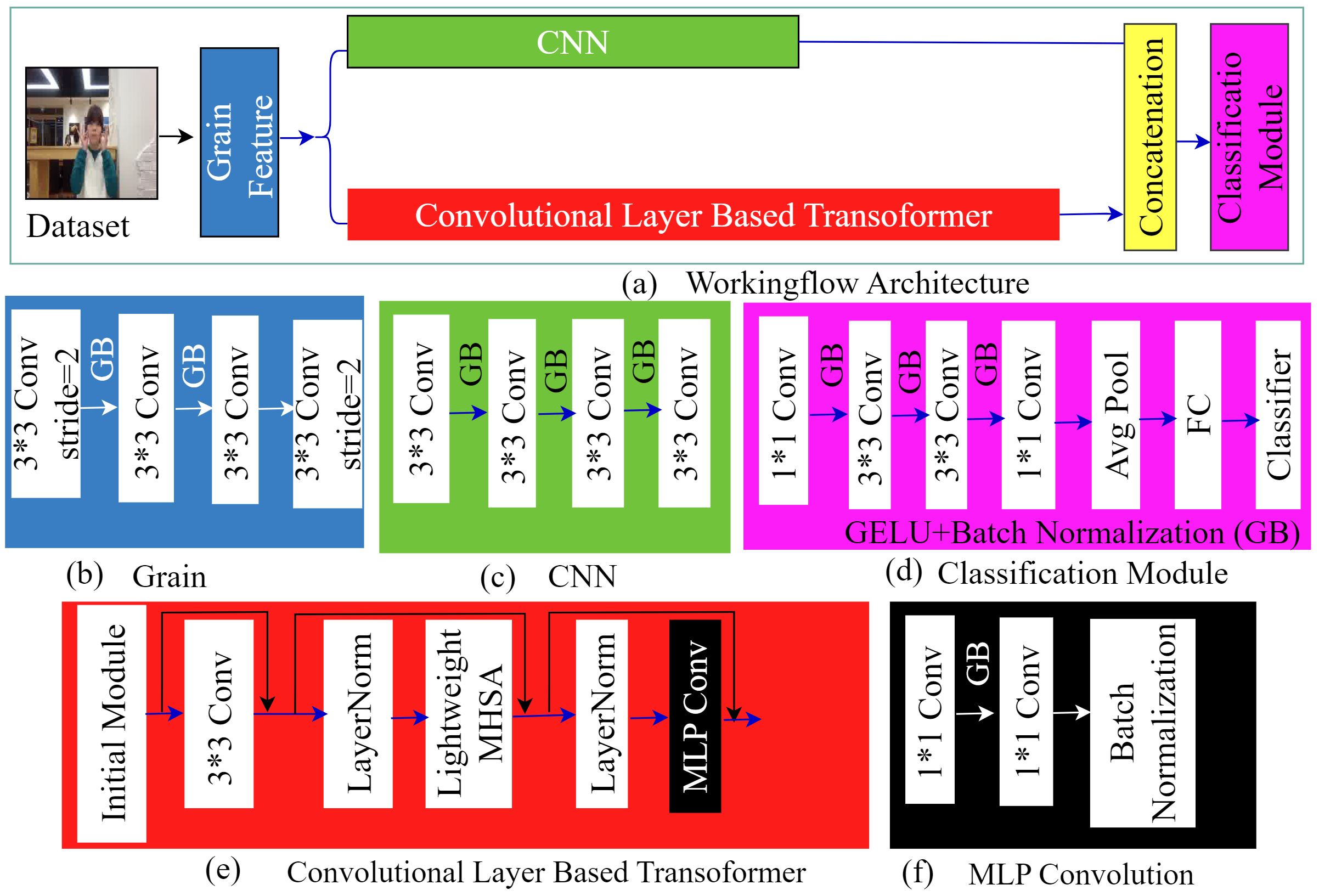}
    \caption{ RGB still-image based gesture recognition using transformer and CNN module\cite{shin2023korean} }
    \label{fig:rgb_cmt_modified}
\end{figure}
\paragraph{Transformer and GCN-Based Methods}
Recent advancements in vision-based HGR have seen the emergence of techniques leveraging GNNs or GCNs. The Vision Transformer (ViT) has gained prominence for SL applications \cite{vaswani2017attention}. However, concerns about potential information loss with ViT have led to the development of transformer models like CNN meets Transformer (CMT). Guo et al. introduced CMT, incorporating self-attention with CNN layers to extract multi-scale features efficiently \cite{guo2022cmt}. Subsequent optimizations by Shin et al. further improved CMT's performance, achieving impressive accuracy rates for KSL datasets \cite{shin2023korean}. This system may encounter difficulties with BdSL, JSL, or ASL datasets and vice versa. To address these challenges, researchers have been working to develop automatic multi-cultural SLR systems using ML and DL approaches \cite{nurnoby2023multi}. Figure \ref{fig:rgb_cmt_modified} visualizes the architecture of the model. More recently, Miah et al. employed a Graph meet with CNN and Transformer model (GmTC) module to enhance multi-culture SLR, and they achieved good performance accuracy \cite{miah2024hand_multi}. Figure \ref{fig:rgb_GmTC} demonstrated the working architecture of the GmTC. 

\subsubsection{Challenges and Future Direction}
Current challenges in RGB still image-based HGR involve limited model effectiveness in handling orientation variations, partial occlusion, and the inability to capture depth and spatial information accurately. The lack of diverse datasets and the need for computationally efficient models are additional hurdles.
Future research directions can address these challenges by exploring multiview augmentation techniques, integrating 3D approaches, incorporating supplementary features like temporal information and DL, expanding datasets, and developing signer-independent systems for SLR tasks. Improved image-gathering techniques are also crucial for better motion capture. Benchmarking against state-of-the-art methods, exploring 3D approaches, and incorporating additional features or techniques for various SLRs are also important. 

\subsection{RGB Video Modality Based Dynamic Gesture Recognition} 
Figure \ref{fig:basic_hgr_recognition_model} demonstrated the overview of HGR recognition architecture, including video modality. In dynamic HGR, signers articulate signs in a sequence of frames which need to express the full meaning of a single hand gesture \cite{zhang2019two, ullah2017action}. The main challenge of dynamic HGR is to extract temporal dependencies and relationships among the consecutive frames. An illustration of this can be seen in the continuous gestures used to convey the sentence "IT IS CLOSED TODAY" in ISL as highlighted in\cite{tripathi2015continuous, mohamed2021review}. 


\subsubsection{Dataset}
Dynamic HGR is crucial to express the meaning of the sign and hand gesture, as in the RGB still image. Table \ref{tab:rgb_video_dataset} summarizes these datasets, listing details such as name, release year, number of gesture classes, subjects involved, total samples, samples per class, modality (RGB or RGB+D), and primary performance metrics, typically accuracy.

Key datasets include Montalbano (v2), 20BN-jester, ChaLearn LAP IsoGD, DVS Gesture, and SKIG, spanning from 2013 to 2022 and covering a wide range of classes and sample sizes. These datasets are essential for advancing gesture recognition research and providing benchmarks for model development and performance evaluation. Notable datasets include KSL-77 from Korea with 112,564 samples \cite{8299578_d_vid_EgoGesture}, Jester with 148,092 samples \cite{kopuklu2018motion_d_vid_Jesture}, LSFB-CONT from Belgium with 85,000 samples \cite{fink2021lsfb_d_rgb_LSFB-CONT}, and LSA64 capturing 3200 samples \cite{ronchetti2023lsa64_d_rgb_LSA64}.
Moreover, the Kurdish sign language dataset is a video dataset collected using a high-quality webcam with a resolution of $640 \times 480$ at 30 frames per second \cite{mahmood2018dynamic}. The JSL video dataset includes 92 JSL words signed by ten native signers. The videos, recorded in office environments using smartphone cameras, consist of 3,900 usable videos for 78 words \cite{takayama2018sign}. The ISL dataset comprises approximately 900 static images and 700 videos of alphabets and dynamic words collected from seven participants using an external webcam \cite{athira2022signer}.

The German sign language dataset RKS-PERSIANSIGN, collected by \cite{rastgoo2020hand}, includes three combined datasets, while RWTH-PHOENIX-Weather-2014 \cite{pu2018dilated} consists of RGB videos with approximately 1 million frames covering 6,841 sentences in German Sign Language. Jain et al. \cite{jain2023addsl} utilized the Annotated Dataset for Danish Sign Language (ADDSL) for hand gesture detection and recognition in Danish Sign Language. Arabic Sign Language (ArSL) videos, consisting of 450 videos captured at 30 fps, were used for daily school life evaluations \cite{ibrahim2018automatic}. Additionally, a large-scale dataset of 25,000 annotated videos was obtained from public ASL videos on video-sharing platforms, captured by ASL students and teachers \cite{joze2018ms}.

\begin{table*}[ht]
\caption{ RGB video modality-based public dataset that is commonly used for dynamic HGR. } \label{tab:rgb_video_dataset}
\setlength{\tabcolsep}{5pt}
\begin{tabular}{|l|l|l|l|l|l|l|l|l|}
\hline
\begin{tabular}[c]{@{}l@{}}Dataset\ Names\end{tabular} Dataset Name& Year & Lang. &Classes & Subject & \begin{tabular}[c]{@{}l@{}}Total \ Sample \end{tabular} & \begin{tabular}[c]{@{}l@{}}Samples/C  \end{tabular} &\begin{tabular}[c]{@{}l@{}}Input \\ Device \end{tabular} & Latest Performance\\
\hline
SKIG \cite{liu2013learning} & 2013 & Hand Gesture & 10 & 6 & 2160 & - & RGB/D & \begin{tabular}[c]{@{}l@{}}Accuracy \end{tabular} \\\hline
Montalbano (v2) \cite{escalera2015chalearn} & 2014 & Hand Gesture & 20 & 27 & 14000 & - & RGB,D & \begin{tabular}[c]{@{}l@{}}Accuracy \end{tabular} \\\hline
RWTH-PHOENIX-Weather \cite{koller2015continuous_d_vid_RW_GERMANy_PHOENIX14-T} & 2015 & Germany & 1200 & - & 45760 & - & RGB & \begin{tabular}[c]{@{}l@{}}Accuracy \end{tabular} \\\hline
SIGNUM \cite{koller2015continuous_d_vid_RW_GERMANy_PHOENIX14-T} & 2015 & Germany & 450 & - & 33210 & - & RGB & \begin{tabular}[c]{@{}l@{}}Accuracy \end{tabular} \\\hline
ChaLearn LAP IsoGD \cite{wan2016chalearn} & 2016 & Hand Gesture & 249 & 21 & 47933 & - & RGB,D & \begin{tabular}[c]{@{}l@{}}Accuracy \end{tabular} \\ \hline
DVS Gesture \cite{amir2017low} & 2016 & Hand Gesture & 11 & 29 & 1342 & - & RGB & \begin{tabular}[c]{@{}l@{}}Accuracy \end{tabular} \\
NVGesture \cite{molchanov2016online_d_vid_NVgesture} & 2016 & Hand Gesture & 25 & - & 1532 & - & RGB & \begin{tabular}[c]{@{}l@{}}Accuracy \end{tabular} \\\hline
IsoGD & 2017 & Various & 13 & - & 47933 & - & RGB & \begin{tabular}[c]{@{}l@{}}67.14 \end{tabular} \\\hline
PHOENIX14 \cite{Camgoz2018_PHOENIX14} & 2018 & Germany & 1081 & 9 & 6841 & - & RGB & \begin{tabular}[c]{@{}l@{}}Accuracy \end{tabular} \\\hline
PHOENIX14-T \cite{koller2015continuous_d_vid_RW_GERMANy_PHOENIX14-T} & 2015 & Germany & 1085 & 9 & 8227 & 9 & RGB & \begin{tabular}[c]{@{}l@{}}Accuracy \end{tabular} \\ \hline
20BN-jester \cite{materzynska2019jester} & 2019 & Hand Gesture & 20 & 27 & 13858 & - & RGB,D & \begin{tabular}[c]{@{}l@{}}Accuracy \end{tabular} \\\hline
Jester \cite{kopuklu2018motion_d_vid_Jesture} & 2018 & Various & 27 & - & 148092 & - & RGB & \begin{tabular}[c]{@{}l@{}}Accuracy \end{tabular} \\\hline
IPN \cite{benitez2021ipn_d_vid_NPN} & 2021 & Various & 13 & - & 4000 & - & RGB & \begin{tabular}[c]{@{}l@{}}Accuracy \end{tabular} \\\hline
CSL-Daily \cite{Zhou2021} & 2021 & Chinese & 2000 & 10 & 20654 & - & RGB & \begin{tabular}[c]{@{}l@{}}32.2\\ (WER) \end{tabular} \\\hline
SIGNUM \cite{VonAgris2008_signum} & 2021 & German & 1230 & 25 & 15075 & - & RGB & \begin{tabular}[c]{@{}l@{}}- \end{tabular} \\\hline
BOBSL \cite{albanie2021bbc_BOBSL} & 2021 & British & 395 & 85 & 47551 & - & RGB & \begin{tabular}[c]{@{}l@{}}50.5 \cite{albanie2021bbc_BOBSL} \end{tabular} \\\hline
LSFB \cite{Fink2021_LSFB} & 2021 & \begin{tabular}[c]{@{}l@{}}French,\\ Belgian\end{tabular} & 6883 & 100 & 85132 & - & RGB & \begin{tabular}[c]{@{}l@{}}51.5 \cite{Fink2021_LSFB} \end{tabular} \\\hline
KSL-77 \cite{10360810_ksl2} & 2022 & Hand Gesture & 77 & 25 & 112564 & 1461 & RGB & Accuracy \\\hline
EgoGesture \cite{8299578_d_vid_EgoGesture} & 2022 & Hand Gesture & 83 & 50 & 2081 & - & RGB & \begin{tabular}[c]{@{}l@{}}91.64 \cite{karsh2024mxception} \end{tabular} \\\hline
MSRGesture \cite{8299578_d_vid_EgoGesture} & 2022 & Hand Gesture & 9 & - & - & - & RGB & \begin{tabular}[c]{@{}l@{}}99.41 \cite{karsh2024mxception} \end{tabular} \\\hline
LSA-T \cite{dal2022lsa_v_rgb_LSAT} & 2022 & Argentinian & 103 & - & 14880 & - & RGB & \begin{tabular}[c]{@{}l@{}}Accuracy \end{tabular} \\\hline
27 ASL \cite{mavi2022new} & 2022 & American & 27 & 173 & - & - & RGB & \begin{tabular}[c]{@{}l@{}}- \end{tabular} \\\hline
LSE-Sign \cite{Gutierrez-Sigut2016} & 2022 & Spanish & 5100 & - & - & - & RGB & \begin{tabular}[c]{@{}l@{}}- \end{tabular} \\\hline
ASLG-PC12 \cite{Yin2020} & 2023 & American & - & - & 87709 & - & RGB & \begin{tabular}[c]{@{}l@{}}Accuracy \end{tabular} \\\hline
ASL-LEX 2.0 \cite{Sehyr2021} & 2023 & American & - & - & 87709 & - & RGB & \begin{tabular}[c]{@{}l@{}}- \end{tabular} \\\hline
LSA64 \cite{ronchetti2023lsa64_d_rgb_LSA64} & 2023 & Argentinian & 64 & - & 3200 & - & RGB & \begin{tabular}[c]{@{}l@{}}Accuracy \end{tabular} \\\hline
\end{tabular}
\end{table*}

\begin{table*}[]
\centering
\caption{Methodological summary of the dynamic HGR using video modality}
\label{tab:accuracy_rgb_video}
\begin{tabular}{|c|c|c|c|c|c|c|c|}
\hline
\textbf{Author}     & Year     & \textbf{Dataset Name} & \textbf{No Class} & \textbf{No Sample} & \textbf{Feature Model}                                  & \textbf{Classifier} & \textbf{Accuracy/J.I.} \\ \hline
Devi et al. \cite{Devi2015} &2015& ChaLearn LAP IsoGD &  249& 21 & C3D & SoftMax & 57.40 \\ \hline
Szegedy et al. \cite{Szegedy2016}&2016 & Montalbano &  20  &27 & Two-stream+RNN & SoftMax & 91.70 \\ \hline
Cheng et al.  \cite{Cheng2016}&2016 & ChaLearn LAP IsoGD & 249& 21 & C3D+LSTM & Softmax & 68.14 \\ \hline
Oliveira et al.  \cite{oliveira2017dataset} &2017   & Iris SL                   & 23             & 58,114            & PCA                                                     & PCA              & 95.00\%                 \\ \hline
Ibrahimet et al.   \cite{ibrahim2018automatic}   &2018 & Arabic SL                   & 222             & 450           & Geometric features                                                     & Euclidean distance              & 97.00\%                 \\ \hline
Sun et al.  \cite{Sun2018} &2018& SKIG &  10 & 6& C3D+LSTM &  SoftMax& 98.60 \\ \hline
Zhao et al. X \cite{Zhao2018}&2018 & Montalbano &  20  &27 & DNN+DCNN & SoftMax & 81.62 \\ \hline
Pigou et al. \cite{Pigou2018}&2018 & Montalbano &  20  &27 & RNN &  SoftMax& 67.71 \\ \hline
Islam et al.  \cite{islam2018hand} &2018    & American SL                 & 26             & N/A             & DCNN                                                         & MCSVM               & 94.57\%              \\ \hline
Mahmood   et al. \cite{mahmood2018dynamic} &2018  & Kurdish SL                  & 10             & N/A             & N/A                                                          & ANN                 & 98.00\%              \\ \hline

Takayama   et al.  \cite{takayama2018sign}& 2018& Japanese SL                 & 92             & N/A             & Z-Score, PCA                                                 & HMM                 & 93.35\%              \\ \hline
Pu   et al.  \cite{pu2018dilated}&  2018    & German SL  & 9    & N/A             & 3D-ResNet     & 3D-ResNet    \\ \hline
Pu   et al. \cite{Pu2019} &  2019    &  \begin{tabular}[c]{@{}c@{}}PHOENIX-Weather\\ CSL\end{tabular}                    & -             & N/A             & 3D-ResNet+BiLSTM                                                    & \begin{tabular}[c]{@{}c@{}}Attention\\ Soft DTW \end{tabular} 
 & \begin{tabular}[c]{@{}c@{}}36.7/wer\\ 67.00 \end{tabular}  ,                  \\ \hline
Guo et al. \cite{Guo2019}&2019 & SKIG & 10 & 6 & ResC3D+Attention & SoftMax & 90.60 \\ \hline
Jiang et al \cite{Jiang2019} &2019& ChaLearn LAP IsoGD &  249& 21 & ResC3D &  SoftMax& 50.93 \\ \hline
Gao et al. \cite{gao2020two} &2020& ASL &  &  & 2S-CNN &  SoftMax& 92.00 \\ \hline
Sharma et al. \cite{Sharma2020}&2020& ChaLearn LAP IsoGD &  249& 21 & C3D+Pyramid & SoftMax & 49.20 \\ \hline
Gao et al.  \cite{Gao2020}& 2020& SKIG & 10 & 6 & R3DCNN+RNN &  SoftMax& 100.0 \\ \hline
Rastgoo   et al.  \cite{rastgoo2020hand} &2020 & Persian SL                  & 100            & N/A             & \begin{tabular}[c]{@{}c@{}}3DCNN,\\    ResNet50\end{tabular} & LSTM                & 99.80\%              \\ \hline
Cheng et al.  \cite{Cheng2020} & 2020  & Arabic dataset                   & 120           & -           & CNN                                                    & Bi-LSTM              & 97.3\%                 \\ \hline

Li et al. \cite{Li2020} &2020   & SLT (RPWT)dataset                    & 120           & 3000           & TSPNet                                                    & Encoder/decoder              & 97.3\%                 \\ \hline

Camgoz  et al.  \cite{Camgoz2020} &  2020 & \begin{tabular}[c]{@{}c@{}}PHOENIX14\\PHOENIX14-T\end{tabular}       & \begin{tabular}[c]{@{}c@{}}1081\\6841\end{tabular}           &   \begin{tabular}[c]{@{}c@{}}1085\\8227\end{tabular}            & CNN                                                  & CTC             & -              \\ \hline

Niu  et al.  \cite{Niu2020} &  2020 & \begin{tabular}[c]{@{}c@{}}PHOENIX14\\PHOENIX14-T\end{tabular}       & \begin{tabular}[c]{@{}c@{}}1081\\6841\end{tabular}           &   \begin{tabular}[c]{@{}c@{}}1085\\8227\end{tabular}        & CNN                                               & CTC           & 26.8(WER(               \\ \hline

Niu  et al.  \cite{Niu2020} &  2020 & \begin{tabular}[c]{@{}c@{}}PHOENIX14\\PHOENIX14-T\end{tabular}       & \begin{tabular}[c]{@{}c@{}}1081\\6841\end{tabular}           &   \begin{tabular}[c]{@{}c@{}}1085\\8227\end{tabular}        & CNN                                               & CTC           & 26.8(WER(               \\ \hline
Yin et al.  \cite{Yin2020} &  2020 & \begin{tabular}[c]{@{}c@{}}PHOENIX14-T\\ ASLG-PC12\end{tabular}       & \begin{tabular}[c]{@{}c@{}}6841\end{tabular}           &   \begin{tabular}[c]{@{}c@{}}8227\\ 87709\end{tabular}        & STMC-Transformer                                               & Bi-LSTM, CTC   & 96.60               \\ \hline

Hoao et al.  \cite{Hao2021} &  2021 & \begin{tabular}[c]{@{}c@{}}PHOENIX14\\PHOENIX14-T\end{tabular}       & \begin{tabular}[c]{@{}c@{}}1081\\6841\end{tabular}           &   \begin{tabular}[c]{@{}c@{}}1085\\8227\end{tabular}      & 3D CNN                                                   & Bi-LSTM+CTC              & 22.00,22.40               \\ \hline

Zhou et al.  \cite{Zhou2021} &  2021 & \begin{tabular}[c]{@{}c@{}}Kinetics-400\\ HSL\end{tabular}      & 8           & -          & HOG                                                  & (3+2+1)D CNN              & 94.60               \\ \hline

Hu  et al.  \cite{Hu2023} &  2023 & \begin{tabular}[c]{@{}c@{}}Kinetics-400\\ HSL\end{tabular}      & 8           & -          & HOG                                                  & (3+2+1)D CNN              & 94.60               \\ \hline
Athira et al. \cite{athira2022signer} &  2022   & Indian SL                   & N/A            & N/A             & HOG                                                          & SVM                 & 91\%                 \\ \hline

Jain   et al.  \cite{jain2023addsl} &2023   & Danish SL                   & 36             & N/A             & Deep CNN                                                     & YOLOv5              & 92\%                 \\ \hline 

Shanableh et al.  \cite{Shanableh2023}  &2023  & Arabic dataset                   & 120           & -           & CNN                                                    & Bi-LSTM              & 97.3\%                 \\ \hline
Guo et al.  \cite{Guo2023} &  2023 & \begin{tabular}[c]{@{}c@{}}PHOENIX14-T\\ CSL\end{tabular}       & \begin{tabular}[c]{@{}c@{}}100 \\ \end{tabular}           &   \begin{tabular}[c]{@{}c@{}}8227\\25000\end{tabular}        & CNN + LSTM +
HMM                                              & Transformer   & \begin{tabular}[c]{@{}c@{}}21.10 (WEN)\\ 98.25\end{tabular}              \\ \hline

DU et al.  \cite{Du2022} &  2023 & \begin{tabular}[c]{@{}c@{}}WLASL\\ NMFs-CSL. \end{tabular}       & \begin{tabular}[c]{@{}c@{}}2000\\1067 \end{tabular}           &   \begin{tabular}[c]{@{}c@{}}21083\\32,010\end{tabular}        & Transformer-spatial& temporal-softmax  & \begin{tabular}[c]{@{}c@{}}57.13\\ 72.4\end{tabular}              \\ \hline

Zhou et al.  \cite{Zhou2021b} &  2023 & \begin{tabular}[c]{@{}c@{}}PHOENIX14-T\\ CSL\\HKSL \end{tabular}            & \begin{tabular}[c]{@{}c@{}}8227\\100\\ 50\end{tabular}           &   \begin{tabular}[c]{@{}c@{}}8227\\25000\\ 2400\end{tabular}        & \begin{tabular}[c]{@{}c@{}}(3+2+1)D ResNet\\ +
BiLSTM +\\ BERT\end{tabular}  & CTC  & \begin{tabular}[c]{@{}c@{}}20.2\\ 23.30\\12.45 WEN\end{tabular}            
\\ \hline

Karsh et al.  \cite{karsh2024miv3net} & 2024 & \begin{tabular}[c]{@{}c@{}}MUGD\\ ISL, ArSL \\ NUS-I, NUS-II \end{tabular}            & \begin{tabular}[c]{@{}c@{}}-\end{tabular}           &   \begin{tabular}[c]{@{}c@{}}-\end{tabular}        & \begin{tabular}[c]{@{}c@{}}inception V3 \\ \end{tabular}  & mIV3Net  & \begin{tabular}[c]{@{}c@{}}97.14,\\ 99.3 , 97.4 \\99, 99.8 \end{tabular}            
\\ \hline

Karsh et al.  \cite{karsh2024mxception} & 2024 & \begin{tabular}[c]{@{}c@{}}EgoGesture\\ MSR Gesture \end{tabular}            & \begin{tabular}[c]{@{}c@{}}-\end{tabular}           &   \begin{tabular}[c]{@{}c@{}}-\end{tabular}        & \begin{tabular}[c]{@{}c@{}}Xception \\ \end{tabular}  & CNN  & \begin{tabular}[c]{@{}c@{}}91.64\\99.41\end{tabular}            
\\ \hline
Hax et al. \cite{hax2024novel} & 2024 & Depth\_Camera\_Dataset & 6 & 662 & Inception-v3 (CNN) & LSTM (RNN) & 83.66\% \\ \hline

Zhou et al. \cite{zhou2024fgdsnet}& 2024 & \begin{tabular}[c]{@{}c@{}}OUHANDS\\ HGR1\\ EgoHands\\NUS-II  \end{tabular}            & \begin{tabular}[c]{@{}c@{}} 10\\ 25\\ 4\\ 10\end{tabular}           &   \begin{tabular}[c]{@{}c@{}}-\end{tabular}        & \begin{tabular}[c]{@{}c@{}}FGDSNet \\ \end{tabular}  & Softmax  & \begin{tabular}[c]{@{}c@{}}89.57\\96.89\\ 97.69\\ 99.80\\ \end{tabular}            
\\ \hline

Zhou et al. \cite{zhou2024fgdsnet}& 2024 & \begin{tabular}[c]{@{}c@{}}OUHANDS\\ HGR1\\ EgoHands\\NUS-II  \end{tabular}            & \begin{tabular}[c]{@{}c@{}} 10\\ 25\\ 4\\ 10\end{tabular}           &   \begin{tabular}[c]{@{}c@{}}-\end{tabular}        & \begin{tabular}[c]{@{}c@{}}FGDSNet \\ \end{tabular}  & Softmax  & \begin{tabular}[c]{@{}c@{}}89.57\\96.89\\ 97.69\\ 99.80\\ \end{tabular}            
\\ \hline
Zerrouki et al. \cite{zerrouki2024deep} & 2024 & \begin{tabular}[c]{@{}c@{}} Interactive Museum\\ Maramotti \end{tabular} &  7 &  700  &  \begin{tabular}[c]{@{}c@{}}Image-based\\ feature extraction\end{tabular} & Bi-LSTM & 99.86\% \\ \hline
Farid et al. \cite{al2024single_faridi_one_shot} & 2024 & \begin{tabular}[c]{@{}c@{}} SKIG\\ DCOG \end{tabular}  & 10 &  \begin{tabular}[c]{@{}c@{}}1080 \\ -\end{tabular} & - & \begin{tabular}[c]{@{}c@{}} SSD-CNN \\with deep\\ dilated masks\end{tabular} & \begin{tabular}[c]{@{}c@{}} 90.61 \\ 88.56\end{tabular} \\ \hline
\end{tabular}
\end{table*}

%
\subsubsection{Methodology of the RGB Video Modality for Dynamic HGR}
Table \ref{tab:accuracy_rgb_video} demonstrated the summary of the existing Video modality HGR system including year, dataset information,  feature extraction and classification method performance.  Dynamic gestures involve both temporal and spatial information, requiring each dynamic HGR system to extract these features to enhance performance, accuracy, and efficiency. Below, we describe the methodology for dynamic HGR, encompassing both ML and DL approaches. In DL, approaches can be categorized based on feature types, including single-stream CNN, two-stream CNN (which includes spatial and temporal streams), 3D CNN, LSTM, and transfer learning. These methods leverage various network architectures to capture and process the intricate details in dynamic gestures.

\paragraph{ML Based Feature Extraction Classification Approaches}
Many researchers have been working to develop dynamic HGR systems using various methodologies, such as the Dynamic Time Warping (DTW) algorithm and the HMM \cite{plouffe2016static,fine1998hierarchical,haid2019inertial}.
Dynamic HGR involves analyzing both temporal and spatial information, which presents unique challenges. Two classical methods for recognizing dynamic gestures are the Dynamic Time Warping (DTW) algorithm and the HMM.
The DTW approach calculates the similarity between two different-length time series signals. Dynamic HGR can measure the similarity between two videos that consist of hand gestures. The concept is two video modalities need to be aligned before calculating the similarities. 
Corradini et al. \cite{corradini2001dynamic} employed DTW for dynamic HGR. Their methodology involved preprocessing the video, extracting features, and normalizing them into a sequence template. The resulting sequences were matched with templates in the training set, and the gesture recognition result corresponded to the category with the smallest difference from the template. However, the DTW approach does not utilize statistical formulas during training and faces challenges with large data volumes and complex gestures.

To overcome these challenges, many researchers use HMM for state sequence or time series problems. HMM employs forward and backward algorithms to train each gesture category. During testing, all samples traverse HMM models using the forward algorithm to compute the probability of each trial.
Saha et al. developed a dynamic HGR system using HMM, achieving 90\% accuracy with 12 dynamic gesture datasets \cite{saha2017hmm}. Similarly, Yang et al. employed HMM for dynamic HGR by converting images to HSV color models, segmenting with threshold values, and using a state-based positioning algorithm to determine gesture start and end positions in continuous video \cite{yang2012dynamic}. They extracted hand-crafted features such as size, speed, and shape, as well as trained and tested their model with HMM. Later, Kevin P. Murphy expanded Yang's algorithm using the HMM toolset and an additional library for dynamic HGR.
Takayama et al. \cite{takayama2018sign} employed HMM for JSL word classification, significantly reducing annotation workload and achieving 38.01\% accuracy. HMM and DTW-based systems generally include three steps: preprocessing, feature extraction, and model construction. Preprocessing is tailored to scenarios like skin color and hand region. Feature extraction involves parameters such as edge changes, wrist angle, finger angle, cross-section area, movement centre, number of fingers, and radius area. The DTW and HMM models use a k-means algorithm for clustering feature sets. Recently, Neverova et al. proposed gesture recognition using a multimodal dataset, incorporating multimodal densities from the training data, but large datasets still face performance challenges \cite{neverova2016moddrop}.

\paragraph{Single Stream and Two Stream CNN} 
DL-based CNN has proven its excellence in various fields, including computer vision and HGR, achieving significant improvements. Researchers have explored diverse applications using various data. For instance, Xu applied a CNN to recognize gestures using a complex dataset recorded with a monocular camera, reporting good performance accuracy \cite{xu2017real}.
\begin{figure}[htp]
    \centering
    \includegraphics[width=9cm]{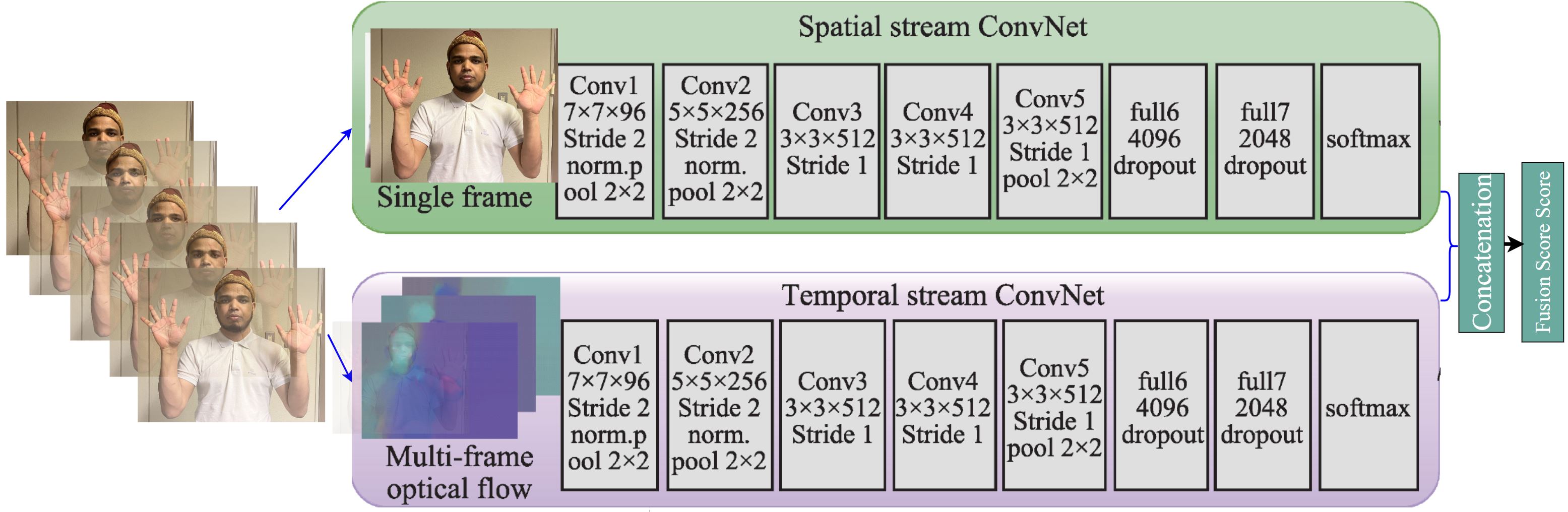}
    \caption{Dynamic Gesture Recognition Using Two-stream CNN}
    \label{fig:vid_two_stream}
\end{figure}

Islam et al. \cite{islam2018hand} use Deep Convolutional Neural Networks (DCNNs) for efficient feature extraction in ASL recognition. Athira et al. \cite{athira2022signer} employ a Histogram of Orientation Gradient (HOG) and Histogram of Edge Frequency (HOEF) with CNNs for static and dynamic gesture recognition in ISL, achieving 91\% and 89\% accuracy for finger spelling alphabets and single-handed dynamic words, respectively. Traditional 2D CNNs struggle with video data due to their inability to capture temporal features. To address this, researchers use two-stream CNN networks, integrating spatial and temporal feature extraction streams, often constructed with 2D CNNs like VGG, AlexNet, InceptionNet, and ResNet.

The spatial-temporal-based two-stream network, first introduced by Simonyan in 2014, demonstrated impressive results in video classification tasks using Inception-based DL \cite{simonyan2014two}. This network uses five sequences of frames for the spatial stream input and the remaining five frames for the temporal stream. The final features are produced by fusing these streams to determine the gesture category. However, this method struggles to extract long-range information, which is crucial for dynamic HGR. To address this, Wang et al. applied a temporal segment network (TSN) as a sparse sampling-based method to capture long-range dependencies \cite{wang2016temporal}. The TSN method, constructed with Inception v2, divides the video into segments, produces probability vectors for each, and averages these vectors for the final long-term video prediction. To further improve performance accuracy, Feichtenhofer et al. fused spatial and temporal information in TSN, aiming for higher layer fusion to generate category score fusion \cite{feichtenhofer2016convolutional}. Zhu et al. applied a CNN-based motion net to enhance spatial-temporal features, achieving good performance accuracy with a dynamic hand gesture dataset \cite{zhu2019hidden}.

Kopuklu et al. applied a two-stream approach using frames and a single RGB image in the TSN's temporal network \cite{kopuklu2018motion}. Wang et al. developed a two-stream RNN for skeleton-based action recognition, effectively modeling temporal dynamics and spatial configurations with skeletal data \cite{wang2017modeling}. Gao et al. proposed a two-stream-based ASL recognition system using CNN, named 2S-CNN, achieving 92\% accuracy with an ASL dataset \cite{gao2020two}.
While 2D CNNs are effective for extracting spatial contextual features, they often overlook temporal contextual information, which is crucial for dynamic HGR involving both images and videos. To address this, it is essential to switch to 3D CNNs, which can capture timing information from frame sequences more effectively.

\begin{figure}[htp]
    \centering
    \includegraphics[width=8cm]{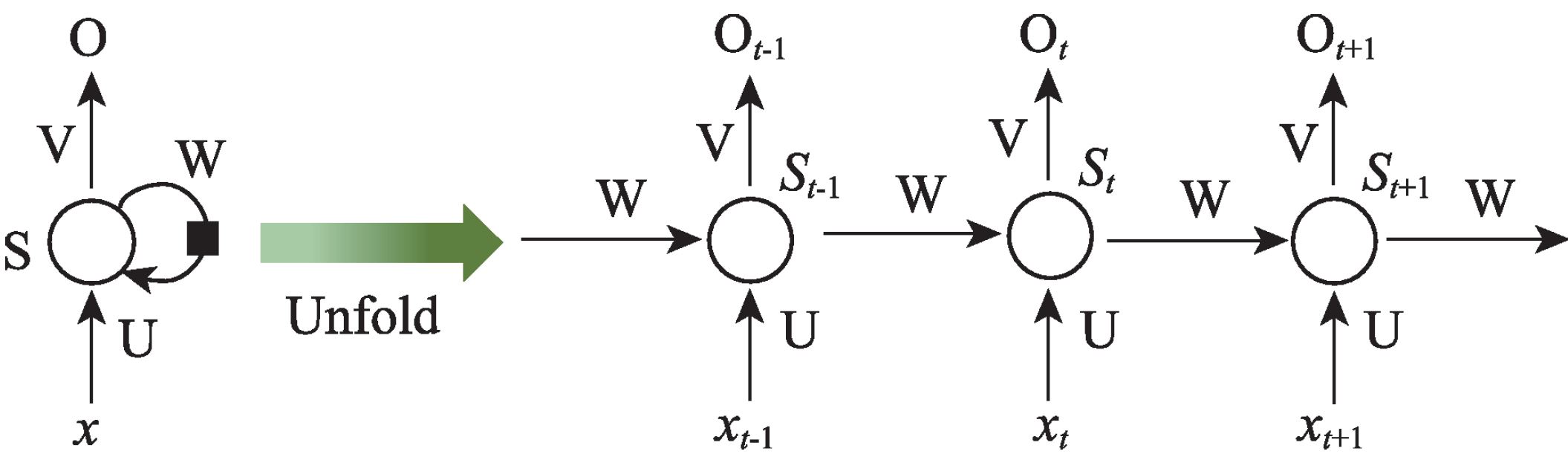}
    \caption{Tomporal domain analysis with RNN}
    \label{fig:vid_RNN}
\end{figure}

\paragraph{Recurrent Neural Networks(RNN) Based Method}
GPT
For dynamic gesture recognition, capturing the chronological sequence is crucial, and traditional networks struggle with this task. Recurrent Neural Networks (RNNs) address this challenge by processing sequential data and generating effective features. They link lower and upper frames to represent the time dimension, with LSTM being a type of RNN that enhances this capability \cite{Pigou2018}. Figure \ref{fig:vid_RNN} illustrates the RNN structure, where \(x_t\) is the input at time \(t\), the hidden layer \(S\), and the output \(O\). This structure captures the relationship among consecutive frames, considering the current frame (\(t\)), previous frame (\(t-1\)), and next frame (\(t+1\)).

\begin{figure}[htp]
    \centering
    \includegraphics[width=8cm]{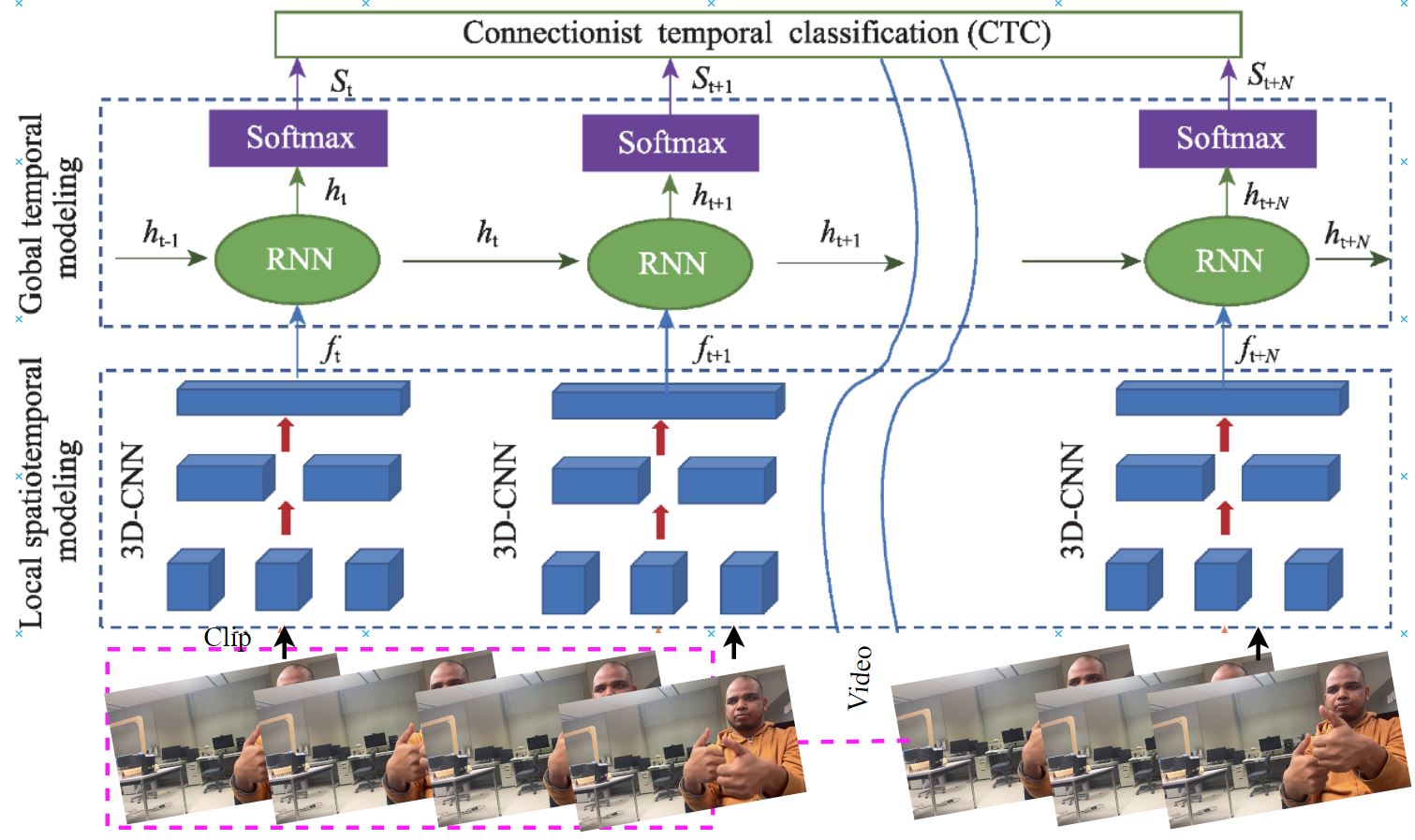}
    \caption{Dynamic gestures classification using R3D CNN}
    \label{fig:vid_R3d_cnn}
\end{figure}

Pigou et al. \cite{Pigou2018} applied an end-to-end neural network incorporating bidirectional recurrence and temporal convolution to enhance performance. More recently, Molchanov et al. proposed the R3DCNN method for gesture recognition, as shown in Figure \ref{fig:vid_R3d_cnn} \cite{molchanov2016online_d_vid_NVgesture}. This method combines a 3D CNN for short-term video sequences with an RNN for long-term video sequences, effectively capturing both spatial and temporal features.
The main drawback of RNN models is their susceptibility to the vanishing gradient problem. This issue occurs because RNNs use backpropagation through time, which can cause gradients to become very small, making it difficult for the network to learn long-term dependencies effectively. Consequently, RNNs often struggle to maintain the "history" effect over extended sequences, leading to challenges in accurately recognizing dynamic gestures that depend on long-term contextual information. To address this problem, Zhu et al. developed a LSTM-based DL system \cite{zhu2019hidden}.
Additionally, they utilized 2D CNNs to further enhance the effectiveness of the feature extraction process.

Mahmood et al. \cite{mahmood2018dynamic} employed an Artificial Neural Network (ANN) for real-time dynamic HGR in Kurdish Sign Language, achieving a performance accuracy of 98\%. Jain et al. \cite{jain2023addsl} utilized sequence modelling techniques with CNNs for Danish SLR, attaining high accuracy with a YOLOv5-based model. Rastgoo et al. \cite{rastgoo2020hand} demonstrated over 90\% recognition accuracy in hand SLR using LSTM on the RKS-PERSIANSIGN dataset.

\paragraph{3D CNN Based Dynamic HGR}
The 3D CNN consists of 3D convolutional, activation, and pooling layers, extracting both spatial and temporal information from frame sequences or videos. While the working procedure is similar to 2D CNN, the key difference is that 2D layers operate on a single feature map with height and width dimensions, whereas 3D layers include height, width, and time dimensions. This temporal information helps 3D CNNs capture temporal dynamics in addition to spatial data.


Many researchers have developed 3D CNN-based systems, often referred to as the C3D model \cite{tran2015learning}. This model consists of eight convolutional layers, five pooling layers, a softmax layer, and two fully connected layers. Each convolutional layer uses the ReLU activation function, and some convolutional layers are integrated with pooling layers to enhance the network without increasing parameters. Both fully connected layers employ a dropout rate of 0.3 to prevent overfitting.

The C3D model \cite{tran2015learning} is recognized as an efficient spatiotemporal feature extractor and is widely employed in dynamic gesture recognition algorithms. Liu et al. proposed a dynamic gesture recognition method based on C3D \cite{liu2013learning}, while Zhu et al. enhanced its capabilities with pyramid input and fusion strategies \cite{zhu2019hidden}. Zhang et al. proposed using C3D and bidirectional convolutional long short-term memory networks (BLCTMNs) to learn 2D temporal feature maps for HGR \cite{zhang2023multimodal}. 
However, the simple architecture of C3D limits its feature expression. To address this, Tran et al. introduced residual connections, resulting in the ResC3D model \cite{tran2017convnet}, which allows for deeper networks without performance degradation. Li et al. further improved dynamic gesture recognition with ResC3D by incorporating an attention mechanism to focus on relevant video frames and motion regions \cite{li2019skeleton}.

Rastgoo et al. \cite{rastgoo2020hand} utilized a 3D Convolutional Neural Network (3DCNN), specifically ResNet50, for pixel-level feature extraction in hand sign recognition from RGB videos. Pu et al. \cite{pu2018dilated} achieved effective continuous SLR on the RWTH-PHOENIX-Weather dataset, with an inference time of around 1 second for a 140-frame sign video. They employed a stacked dilated convolutional network and the Connectionist Temporal Classification (CTC). Their study evaluated several state-of-the-art approaches as baselines for SLR, including 2D-CNN-LSTM, body key-point, CNN-LSTM-HMM, and 3D-CNN \cite{joze2018ms}.

\paragraph{Transfer Learning}
Pu et al. \cite{pu2018dilated} achieved effective continuous SLR using a stacked dilated convolutional network and Connectionist Temporal Classification (CTC) on the RWTH-PHOENIX-Weather dataset.

\paragraph{Attention and Recent Technologies}
Joze et al. \cite{joze2018ms} proposed an I3D architecture for SLR, surpassing existing state-of-the-art methods with an accuracy of 95.16\%. Jain et al. \cite{jain2023addsl} outperformed existing methods in Danish SLR with a YOLOv5-based model, achieving an accuracy of 92\% and an average inference time of 9.02ms per image. The YOLOv5 model, trained with a CSP-DarkNet53 backbone and YOLOv3 head, enabled efficient detection and recognition of signs.

\subsubsection{Current Challenges and Future Direction of the RGB Video Modality}

In video-based HGR, several significant challenges persist:
\textbf{ Temporal Dynamics: }Accurately interpreting the temporal dynamics of gestures, especially in longer sequences, remains difficult.\\
\textbf{Class Variability: }Substantial variability within and between gesture classes complicates the development of robust recognition models.\\
\textbf{Real-time Processing:} Ensuring efficient real-time processing while maintaining high accuracy is a technical hurdle.\\
\textbf{Generalization:} Developing models that generalize well across different environments and users is challenging.

To address these challenges, attention mechanisms and transformer networks can be explored in the future to better capture intricate temporal dynamics and long-range dependencies. Strategies to transfer knowledge from controlled settings to diverse real-world scenarios will be crucial. Combining information from multiple sources (e.g., RGB, depth, and motion sensors) will enhance recognition accuracy and robustness. Advancements in edge computing will enable efficient real-time processing, while privacy-preserving techniques will ensure user data integrity. Incorporating human pose estimation algorithms to improve gesture recognition accuracy. Developing methods to effectively transition models trained on synthetic data to real-world applications. By pursuing these directions, HGR research aims to significantly improve performance and expand applications in human-computer interaction, virtual reality, healthcare, and beyond.

\section{3D Skeleton Modality} \label{sect2}
Skeleton-based HGR addresses challenges faced by image-based SLR systems, such as background clutter, hand occlusion, lighting variations, and enhancing hand movement representation. By extracting joint skeleton key points from RGB videos, researchers have enhanced system accuracy and efficiency while tackling computational complexity. Advancements in 3D camera technology and tools like Mediapipe, Alphapose, and OpenPose have facilitated precise skeleton point collection, creating a more robust gesture recognition framework.
Skeleton-based approaches improve communication for the deaf and hearing impaired by accurately recognizing sign language gestures. 
There are many researchers who have been working to develop skeleton-based HGR systems using handcrafted features with ML or end-to-end DL.

\subsection{Dataset}
There are few skeleton-based datasets available online. Also, we easily extracted skeleton-based datasets from the RGB video.  Table \ref{Tab:d_skeleton} included various benchmark datasets' names of this skeleton-based modality, creation year, number of classes, dataset types, sample size and latest performance accuracy. DHG \cite{de2019spatial_d_skel_ASLLVD}, SHREC \cite{de2019spatial_d_skel_ASLLVD} are the most used skeleton datasets which are publicly available for the HGR domain. Among them, the WLASL dataset comprises 2000 ASL signs from 119 subjects across 21089 videos, averaging 10.5 videos per sign with 67 joints per frame \cite{li2020word_d_skel_WLASL}. MSL dataset includes 30 MSL signs by 20 subjects with 3000 videos, each sign having 20 videos and 67 joints per frame \cite{mejia2022automatic_d_skel_MSL}. ASLLVD dataset features 2745 ASL signs with 9748 videos, averaging 3/4 videos per sign and 67 joints per frame \cite{de2019spatial_d_skel_ASLLVD}. DHG dataset covers 14/28 general signs by 20 subjects, totalling 2800 videos, with 22 joints per frame \cite{de2019spatial_d_skel_ASLLVD}. SHREC dataset also includes 14/28 general signs by 27 subjects in 2800 videos, with 22 joints per frame \cite{de2019spatial_d_skel_ASLLVD}. MSRA dataset consists of 17 MSRA signs with 76500 videos (500 videos per sign) and 21 joints per frame \cite{de2019spatial_d_skel_ASLLVD}. The PSL dataset involves 19 PSL signs with 2700 images (55 images per sign) and 67 joints per frame. Also, Table \ref{tab:skeleton-performance_accuracy} demonstrated information about some datasets of hand gestures, such as Devineau et al. \cite{devineau2018deep} use the DHG dataset for HGR with around 3000 samples. De Smedt et al. \cite{de2016skeleton} employ a dataset of 14 gestures by 20 participants for evaluating their skeleton-based 3D HGR approach. Liu et al. \cite{liu20203d} utilize the ChaLearn 2014 dataset, featuring RGB frames, depth maps, user body masks, and 3D skeletal joint positions for Italian cultural gesture analysis. These datasets are vital for advancing 3D HGR research.

\begin{table*}[]
\setlength{\tabcolsep}{5pt}
\centering
\caption{Skeleton modality-based dataset description.} \label{Tab:d_skeleton}
\begin{adjustwidth}{0cm}{0cm}
\begin{tabular}{|l|l|l|l|l|l|l|l|l|}
\hline
\begin{tabular}[c]{@{}l@{}}Dataset \\ Names\end{tabular} &Year& \begin{tabular}[c]{@{}l@{}}Language\end{tabular} &  Signs & Sub. & \begin{tabular}[c]{@{}l@{}}Total\\ videos\end{tabular} & \begin{tabular}[c]{@{}l@{}}Videos \\ Per\\Sign\end{tabular} & \begin{tabular}[c]{@{}l@{}}Joint\\ Per \\Frame\end{tabular} &\begin{tabular}[c]{@{}l@{}} Latest \\ Performance \\Accuracy\end{tabular}\\ \hline

ASLLVD \cite{de2019spatial_d_skel_ASLLVD}&2019 & ASL &  2745 & n/a & 9748 & 3/4 & 67&- \\\hline
DHG \cite{de2019spatial_d_skel_ASLLVD} &2019& General &  14/28 & 20 & 2800 & - & 22 &-\\\hline
SHREC \cite{de2019spatial_d_skel_ASLLVD} &2019& General &  14/28 & 27 & 2800 & - & 22&- \\\hline
MSRA \cite{de2019spatial_d_skel_ASLLVD} & 2019& General &  17 & n/a & 76500 (i) & 500 & 21&- \\\hline
WLASL \cite{li2020word_d_skel_WLASL}&2020 & ASL &  2000 & 119 & 21089 & 10.5 & 67 &-\\\hline
MSL\cite{mejia2022automatic_d_skel_MSL} &2022& MSL&  30 & 20 & 3000 & 20 & 67 &-\\\hline
PSL & 2020 & pakistani & 19&n/a & \begin{tabular}[c]{@{}l@{}}2700\\(img)\end{tabular} & \begin{tabular}[c]{@{}l@{}}55\\(img)\end{tabular} & 67&- \\ \hline

AUTSL \cite{sincan2020autsl_autsl}& 2020 & Turkey & 226 & \begin{tabular}[c]{@{}l@{}}43\end{tabular} & \begin{tabular}[c]{@{}l@{}}38336\\\end{tabular} &-& 67& 98.00\cite{Jiang2021} \\\hline

Briareo & 2020 & HGR & 12 & \begin{tabular}[c]{@{}l@{}}40\end{tabular} & \begin{tabular}[c]{@{}l@{}}1440\\\end{tabular} &-& -& 96.64 \cite{slama2023str} \\\hline
FPHA& 2020 & HGR & 45 & \begin{tabular}[c]{@{}l@{}}6\end{tabular} & \begin{tabular}[c]{@{}l@{}}1175\\\end{tabular} &-& -& 91.16\cite{slama2023str} \\
\hline
\end{tabular}
\end{adjustwidth}
\end{table*}

\begin{table*}[]
\centering
\caption{Databases and performance of Skeleton modality for HGR}
\label{tab:skeleton-performance_accuracy}
\begin{tabular}{|c|c|c|c|c|c|c|c|}
\hline
\textbf{Author}    &Year           & \textbf{\begin{tabular}[c]{@{}c@{}}Dataset \\ Name\end{tabular}} & \textbf{No Class} & \textbf{ No Sample} & \textbf{\begin{tabular}[c]{@{}c@{}}Feature \\ Model\end{tabular}} & \textbf{Classifier}                                           & \textbf{Performance} \\ \hline
De Smedt et al.   \cite{smedt20163d}   &2016    & American SL                                                             & 14            & 2800            & Fisher Vectors                                                                  & SVM                                                    & 88.00\%              \\ \hline
De   Smedt et al. \cite{smedt20163d} &2016     & American SL                                                            & 10             & N/A             & HOG                                                                    & \begin{tabular}[c]{@{}c@{}}Fisher kernel, \\ SVM\end{tabular} & 86.86\%              \\ \hline

Smedt et al. {\cite{smedt20163d}}& 2016 & DHGD & 28 & 2800 & ASJT & Softmax & 80.11 \\ \hline
De et al. {\cite{de2016skeleton}}&2016 &  DHGD & 28 & 2800 & SoCJ + HoHD + HoWR & Softmax & 80.00 \\ \hline
Chen et al. {\cite{chen2017motion}}& 2017 & DHGD & 28 & 2800 & MARNN & Softmax & 80.32 \\ \hline
Boulahia {\cite{boulahia2017dynamic}}&2017  & DHGD & 28 & 2800 & Boulahia & Softmax & 80.48 \\ \hline
Liu et al.   \cite{liu2017skeleton}& 2017 &      American SL                                                             & 20            & 13,585            & LSTM                                                                 & LSTM                                                    & 96.30\%              \\ \hline
Konstantinidis et al. \cite{konstantinidis2018deep}  &2018&  Argentinian SL                                                         & 64             & N/A             & VGG-19 Network                                                         & \begin{tabular}[c]{@{}c@{}}CNN, \\ RNN, LSTM\end{tabular}     & 98.09\%              \\ \hline
Devineau   et al. \cite{devineau2018deep}&  2018&   N/A                                                                    & 14             & N/A             & DL                                                          & DL                                                 & 91.28\%              \\ \hline

Nunez et al.   \cite{nunez2018convolutional} &2018       & American SL                                                             &60            & 14,000            & CNN, LSTM                                                                 & CNN, LSTM                                                    & 99.00\%              \\ \hline
 Ma et al. {\cite{ma2018hand}} &2018 & DHGD& 28&2800& GREN &-& 82.03 \\ \hline
 Ma et al. {\cite{ma2018hand}} & 2018 &  DHGD & 28 & 2800 & NIUKF-LSTM & Softmax & 80.44 \\ \hline
CNN+LSTM {\cite{nunez2018convolutional}}& 2018 & DHGD & 28 & 2800 & CNN+LSTM & Softmax & 74.19 \\ \hline
Yan et al {\cite{yan2018spatial}} &2018 & DHGD & 28 & 2800 & STA-GCN & Softmax & 87.10 \\ \hline
Hou et al. {\cite{hou2018spatial}}& 2018 & DHGD & 28 & 2800 & Res-TCN & Softmax & 83.60 \\ \hline
Hou et al. {\cite{hou2018spatial}} & 2018& DHGD & 28 & 2800 & STA-Res-TCN & Softmax & 85.00 \\ \hline
Si et al.  {\cite{si2019attention}} &2019 & DHGD & 28 & 2800 & CNN+RNN & Softmax & 74.19 \\ \hline
Res-C3D {\cite{si2019attention}}& 2019 & Shrec & 28 & 2800 & CNN+LSTM & Softmax & 89.52 \\ \hline
Chen et al. \cite{chen2019mfa} &2019&  DHGD&  28 & 2800 & MFA-Net & Softmax & 81.04 \\ \hline
Chen et al.  {\cite{chen2019construct}} &2019 & DHGD & 28 & 2800 & DG-STA & Softmax & 88.00 \\ \hline
Musa et al. {\cite{wei2019multistream}} & 2019& AUSTL & - & - & Multi-Stream GCN & CNN & 99.00 \\ \hline
Rastgooet   al. \cite{rastgoo2021sign}& 2021 &      Persian SL                                                             & 100            & N/A             & 3DCNN                                                                  & \begin{tabular}[c]{@{}c@{}}2DCNN,\\ 3DCNN, LSTM\end{tabular}  & 99.80\%              \\ \hline
Jiang   et al.  \cite{jiang2021skeleton}& 2021 &   Chinese SL                                                             & N/A            & N/A             & \begin{tabular}[c]{@{}c@{}}SL-GCN,\\ SSTCN,  3DCNN\end{tabular}        & GEM                                                           & 99.81                \\ \hline
Han   et al.  \cite{han2021sign}& 2021  &      Chinese SL                                                             & 22             & N/A             & ResNet and  LSTM                                                       & ResNet, LSTM                                                  & 88.6\%               \\ \hline
Jiang   et al.  \cite{jiang2021sign} & 2021 &   Turkish SL                                                             & 226            & N/A             & SSTCN                                                                  & CNN, LSTM                                                     & 98.53\%              \\ \hline
Musa et al {\cite{miah2023dynamic}}&2023  & DHGD & 28 & 2800 & DG-STA & Softmax & 88.00 \\ \hline

Shin et al {\cite{10360810_ksl2}}& 2023 & KSL & 77 & 20000 & Dynami GCNN & Softmax & 99.00 \\ \hline

Hinrichs et al.  \cite{Hinrichs2023} &  2023 & \begin{tabular}[c]{@{}c@{}}PHOENIX14\\PHOENIX14-T\end{tabular}       & \begin{tabular}[c]{@{}c@{}}1081\\6841\end{tabular}           &   \begin{tabular}[c]{@{}c@{}}1085\\8227\end{tabular}        & Transformer                                              &Softmax        &\begin{tabular}[c]{@{}c@{}}18.55,\\ 18.59  (WER)  \end{tabular}     
\\ \hline

Slama et al. \cite{slama2023str} & 2023 & \begin{tabular}[c]{@{}c@{}}SHREC'17 Track \\ Briareo \\ FPHA\end{tabular} & \begin{tabular}[c]{@{}c@{}}14 , 28  \\ 12, \\ 45\end{tabular} &  \begin{tabular}[c]{@{}c@{}}2800 \\ 1440 \\ 1175\end{tabular} & \begin{tabular}[c]{@{}c@{}}  STr-GCN \end{tabular} & Softmax & \begin{tabular}[c]{@{}c@{}} 93.39 \\ 96.64 \\ 91.16\end{tabular} \\ \hline

Peng et al. \cite{peng2023efficient} & 2023 & \begin{tabular}[c]{@{}c@{}}SHREC'17 \\ FPHA\end{tabular} & \begin{tabular}[c]{@{}c@{}}14,28  \\  45\end{tabular} &  \begin{tabular}[c]{@{}c@{}}2800\\ 1175 \end{tabular} & EffGCN & ResGCNeXt & \begin{tabular}[c]{@{}c@{}}95.36, 93.45\end{tabular} \\ \hline
Musa et al. {\cite{miah2024sign_largescale}}&2024  & WLASL & 2000 & - &  \begin{tabular}[c]{@{}c@{}}Dynamic\\ SepTCN GCN\end{tabular}  & Softmax & 90.00\% \\ \hline

Khanna et al.  \cite{khanna2024hand} &  2024 & \begin{tabular}[c]{@{}c@{}}15 gestures\end{tabular}       & \begin{tabular}[c]{@{}c@{}}15\end{tabular}           &   \begin{tabular}[c]{@{}c@{}}\end{tabular}        & Geometric                                               &multi-view CNN        &\begin{tabular}[c]{@{}c@{}}91.00 (WER)  \end{tabular}       
\\  \hline
Mahmud et al.  \cite{mahmud2024quantized} &  2024 & \begin{tabular}[c]{@{}c@{}}DHG-14/28, \\ SHREC-2017\end{tabular}      & \begin{tabular}[c]{@{}c@{}}14\\ 28 \end{tabular}           &   \begin{tabular}[c]{@{}c@{}}\end{tabular}        &   \begin{tabular}[c]{@{}c@{}} Multimodal\\fusion\\CRNN  \end{tabular}                                            &Softmax       &\begin{tabular}[c]{@{}c@{}}90.82:89.21, \\ 93.81:90.24\end{tabular}      
\\  \hline
Jinfu Liu et al. \cite{liu2023temporal} & 2024 & \begin{tabular}[c]{@{}c@{}}SHREC'17, DHG-14/28, \\ NTU RGB+D\\ NW-UCLA\end{tabular} & \begin{tabular}[c]{@{}c@{}}14, 28\\ 60, 10\end{tabular} &  \begin{tabular}[c]{@{}c@{}}2800, 2800 \\ 56880\\ 1494\end{tabular} &  \begin{tabular}[c]{@{}c@{}} \\ spatiotemporal\end{tabular} & TD-GCN & \begin{tabular}[c]{@{}c@{}}97.02:95.36, \\ 93.9:91.4 \\  96.8 \\  97.4\end{tabular} \\ \hline

\end{tabular}
\end{table*}

\subsection{Methodology of the 3D Skeleton Modality}
To overcome the redundant background, computational complexity, and partial occlusion issues, joint skeleton-based data modality become popular in the computer vision field, but it remains a challenging task. Table \ref{tab:skeleton-performance_accuracy} demonstrated the summary of the existing Video modality HGR system, including year, dataset information, feature extraction and classification method performance. Many researchers employed preprocessing, feature extraction with ML and end-to-end DL, which is given below~\cite{ohn2013joint,thakkar2018part,hussein2013human}.

\subsubsection{Preprocessing and Pose Estimation}
In 2D pose estimation, deformable part models are common but often lack detail and context \cite{yang2012articulated}. CNN-based methods improve on this by using detection-based techniques (which create heat maps to detect joints \cite{bulat2016human}) and regression-based techniques (which map joint coordinates using complex functions \cite{toshev2014deeppose}). Detection-based methods usually perform better than regression-based ones.
For 3D pose estimation, the goal is to create a 3D pose that matches the person’s position in the image. Although deep neural networks have improved 3D pose estimation, it remains difficult due to the larger pose space and more ambiguities \cite{zhou2018monocap,gan2013human,chen2015reduced}. Researchers have used media pipe, oppose, alpha pose, and impose to extract the joint keypoint, which constructed 2D key points or 3D key points based on the settings. 

\subsubsection{Hand Crafted Feature and ML Approach}
\begin{figure}[htp]
    \centering
    \includegraphics[width=8cm]{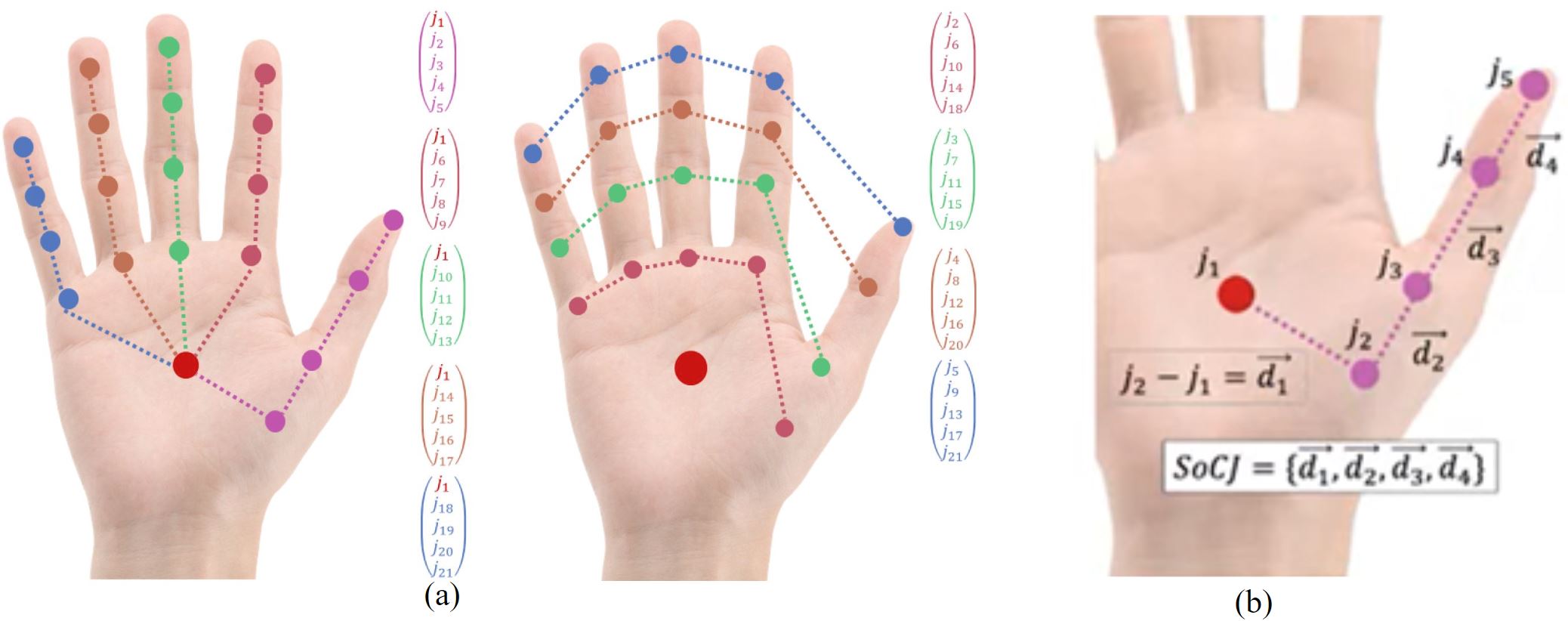}
    \caption{SoCJ handcrafted 9 feature descriptors \cite{de2017shrec}}
    \label{fig:skelton_socj_handcraftfeature}
\end{figure}

After extracting skeleton points from RGB-based data modalities, researchers often use joint skeleton-based feature extraction techniques. Shin et al. employed a geometrical approach to extract distance and angular features from 21 hand key points using the MediaPipe system for an ASL dataset \cite{shin2021american}. Ohn-Bar et al. proposed a feature generator using the Histogram of Oriented Gradients (HOG) algorithm with a linear SVM for classification \cite{ohn2013joint}. Other methods include using a covariance matrix for joint locations \cite{hussein2013human} and joint distance and angles to capture intraclass variance \cite{wang2012mining}. Smedt et al. introduced a hand geometric configuration method for spatial-temporal motion feature extraction, achieving high accuracy on the DHG dataset using SVM \cite{smedt20163d}. 
Smedt et al. extracted features based on Fisher vectors and skeleton-based geometric techniques, then applied SVM to the concatenated features, achieving 83.00\% accuracy for 14 gestures and 80.00\% for 28 gestures in the DHG dataset \cite{de2016skeleton}. They combined three features: the shape of connected joints (SoCJ), histogram of hand directions (HoHD), and histogram of wrist rotations (HoWR). Figure \ref{fig:skelton_socj_handcraftfeature} (a) illustrates the SoCJ features. They also applied Fisher vectors and connected features for the SHREC'17 dataset with an SVM classifier, achieving 88.24\% accuracy for 14 gestures and 81.90\% for 28 gestures \cite{de2017shrec}. Their work highlighted the superiority of 3D skeleton information over depth-based approaches but did not account for gesture amplitude, potentially losing temporal information. Chen et al. proposed a motion feature extractor combining articulated finger movements and global hand movement features to extract bone angles, using RNN for classification. Their model achieved 84.68\% accuracy for 14 classes and 80.32\% for 28 classes on the DHG dataset \cite{chen2017motion}.
Shao et al. proposed a method that integrates shape and motion information using feature descriptors like Motion History Images (MHI) and Predicted Gradients (PCOG) \cite{shao2012human}. De Smedt et al. introduced a methodology focused on extracting hand kinematic descriptors from gesture sequences, encoding them via a Fisher kernel and a multi-level temporal pyramid, and using a linear SVM classifier for recognition \cite{de2019heterogeneous}. Figure \ref{fig:skelton_socj_handcraftfeature} (b) demonstrates the feature calculation procedure. Additionally, Rastgoo et al. \cite{rastgoo2020hand} introduce heatmap images derived from detected key points, offering a new feature representation alongside pixel-level and multi-view hand skeleton features. Table \ref{tab:skeleton-performance_accuracy} shows various existing systems' methods and performance accuracy.

\subsubsection{CNN-Based Methods}
Handcrafted feature-based systems face limitations in efficiency and generalization. To address these, researchers have adopted end-to-end DL algorithms for hand gesture classification using raw skeleton data \cite{li2019skeleton,hou2018spatial}. Many studies have combined CNNs with technologies. For instance, Konstantinidis et al. proposed a skeleton-based SLR system using DL, achieving a 2.27\% increase in accuracy on the LSA64 dataset when using body features over hand features \cite{konstantinidis2018sign}. Nunez et al. \cite{nunez2018convolutional} showed extensive experimental study on publicly available data benchmarks, including the MSR Action3D dataset, MSRDailyActivity3D dataset, UTKinect-Action3D dataset, NTU RGBD dataset, and Montalbano V2 dataset.  Zhang et al. used Kinect-captured skeletal data for improved gesture recognition \cite{zhang2012microsoft}. Huynh-The et al. demonstrated efficient action appearance management by pairing skeleton data with CNNs \cite{huynh2019encoding}. Devineau et al. showed that DL algorithms significantly improved F1 scores for DHG HGR \cite{devineau2018deep}.

\subsubsection{RNN-LSTM Based Methods}
De Smedt et al. \cite{de2019heterogeneous} leverage the inherent structure of hand topology for skeleton-based HGR. Methods using RNNs, especially LSTM units, combined with CNNs are notably more effective than traditional ML and standalone CNN models. Konstantinidis et al. \cite{konstantinidis2018sign} demonstrated the enhanced precision of SLR by integrating HMM with CNNs and bidirectional RNNs incorporating LSTM units.

Lai et al. integrated a CNN with an RNN DL model for recognizing skeleton-based hand gestures, achieving 85.61\% accuracy on the DHG 14 gesture dataset \cite{lai2018cnn+}. Ma et al. combined LSTM with an unscented Kalman filter (UKF) \cite{ma2018hand}, while Han et al. proposed a two-stream method combining RGB and skeleton data using KLSTM-3D ResNet to improve recognition rates \cite{han2021sign}. Han et al. also highlighted the enhanced spatiotemporal feature extraction through the fusion of ResNet and LSTM networks. Nunez et al. integrated CNN and LSTM features for extracting temporal features from 3D pose estimation, reporting 99.00\% accuracy, underscoring the superiority of RNNs, especially LSTM \cite{nunez2018convolutional}.

Lin et al. combined a skeleton LSTM and a Res-C3D network for recognizing hand gestures and compared their result with the existing system\cite{si2019attention}. To improve the performance, Chen et al. extracted articulated finger movement and hand movement features, then concatenated the features and fed them into the RNN and reported 84.68\% accuracy for 14 classes and 80.32\% for 28 classes on the DHG dataset \cite{chen2017motion}. Rastgoo et al. explored DL-based approaches using RNN-LSTM and graph neural networks (GNN or GCN) to enhance gesture recognition performance, offering valuable insights into HGR \cite{rastgoo2020hand}.

\subsubsection{Attention and GCN-Based Methods}
\begin{figure*}[htp]
    \centering
    \includegraphics[width=14cm]{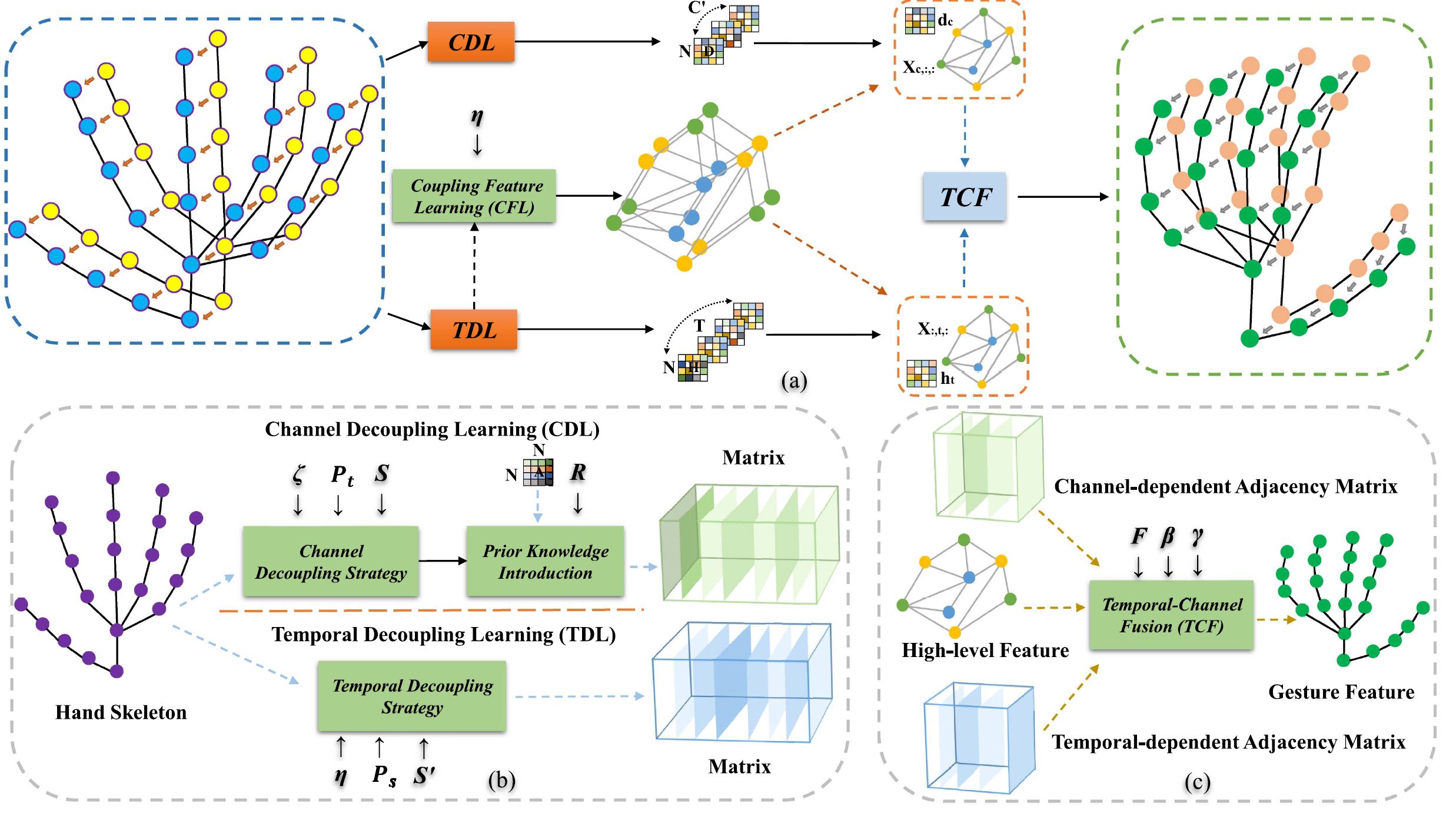}
    \caption{Temporal Decoupling Graph Convolution (TDCN). \cite{liu2023temporal}}
    \label{fig:skelton_TDCN}
\end{figure*}

Recent advancements in the various research domains by using self-attention mechanisms and transfer learning to enhance performance \cite{tian2018cr}. Self-attention networks establish semantic relationships among features \cite{vaswani2017attention}, and spatial-temporal attention has been integrated with various architectures like CNNs and RNNs \cite{si2019attention,hou2018spatial}. Transfer learning techniques have also been explored to leverage pre-trained models for gesture recognition tasks. More recently, Li et al. proposed a temporal GCN (TDCN) to develop a skeleton-based dynamic HGR \cite{liu2023temporal} and Figure \ref{fig:skelton_TDCN} demonstrated the working diagram of this model. The reported  SHREC'17 Track: 97.02\% (14 classes), 95.36\% (28 classes)
DHG-14/28: 93.9\% (14 classes), 91.4\% (28 classes), NTU RGB+D: 96.8\% (cross-view), 92.8\% (cross-subject), NW-UCLA: 97.4\% accuracy. 

 \begin{figure*}[htp]
    \centering
    \includegraphics[width=12cm]{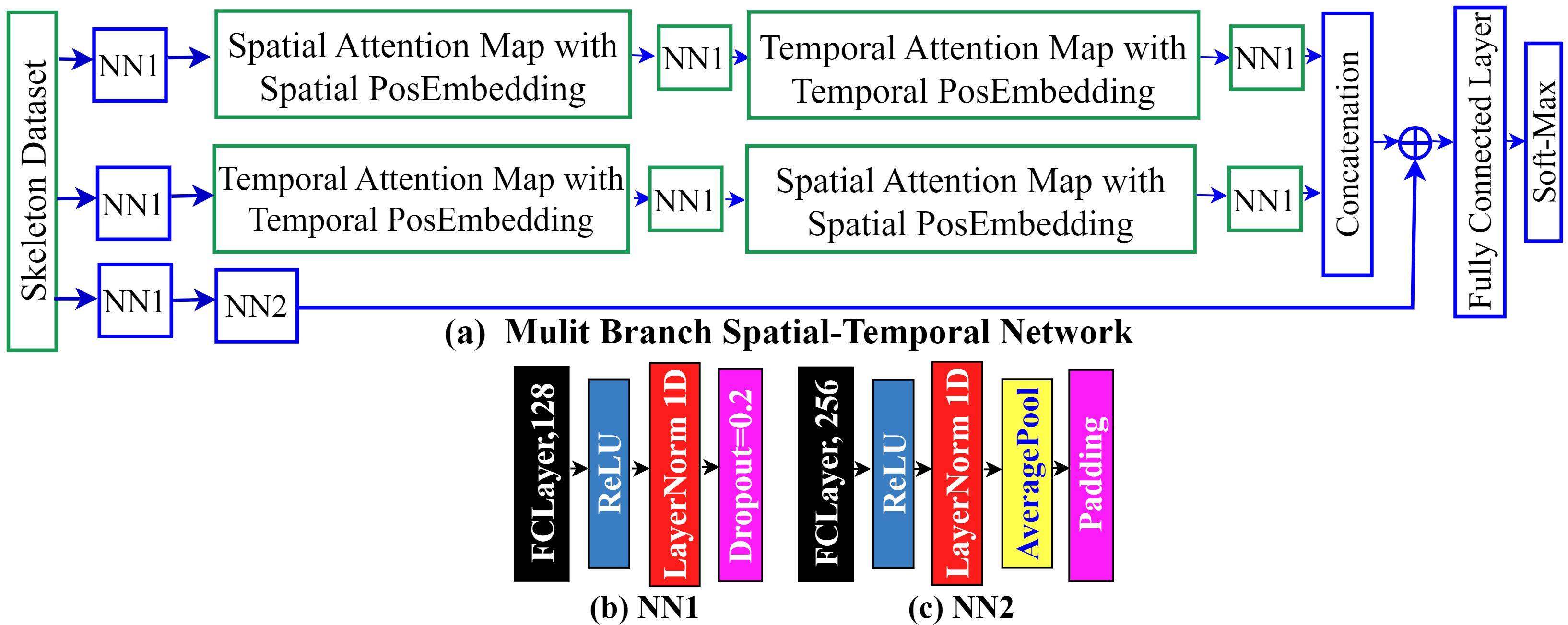}
    \caption{Spatial-temporal graph based dynamic HGR \cite{miah2023dynamic}}
    \label{fig:skelton_musa_gcndynamic}
\end{figure*}

\begin{figure*}[htp]
    \centering
    \includegraphics[width=12cm]{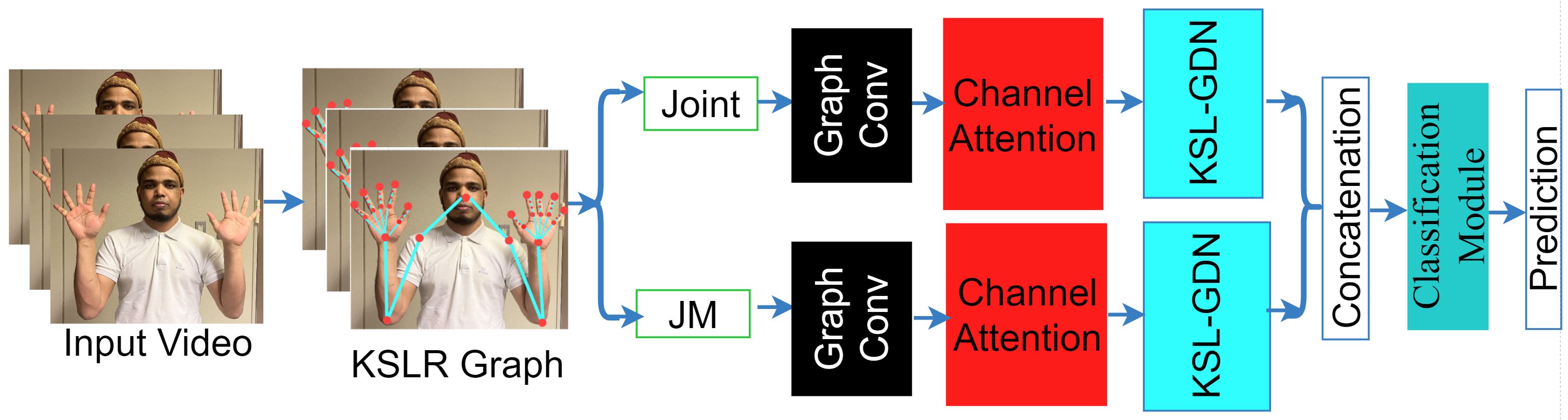}
    \caption{Graph based dynamic KSL recognition \cite{10360810_ksl2}}
    \label{fig:skelton_musa_gcndynamic}
\end{figure*}

\begin{figure}[htp]
    \centering
    \includegraphics[width=8cm]{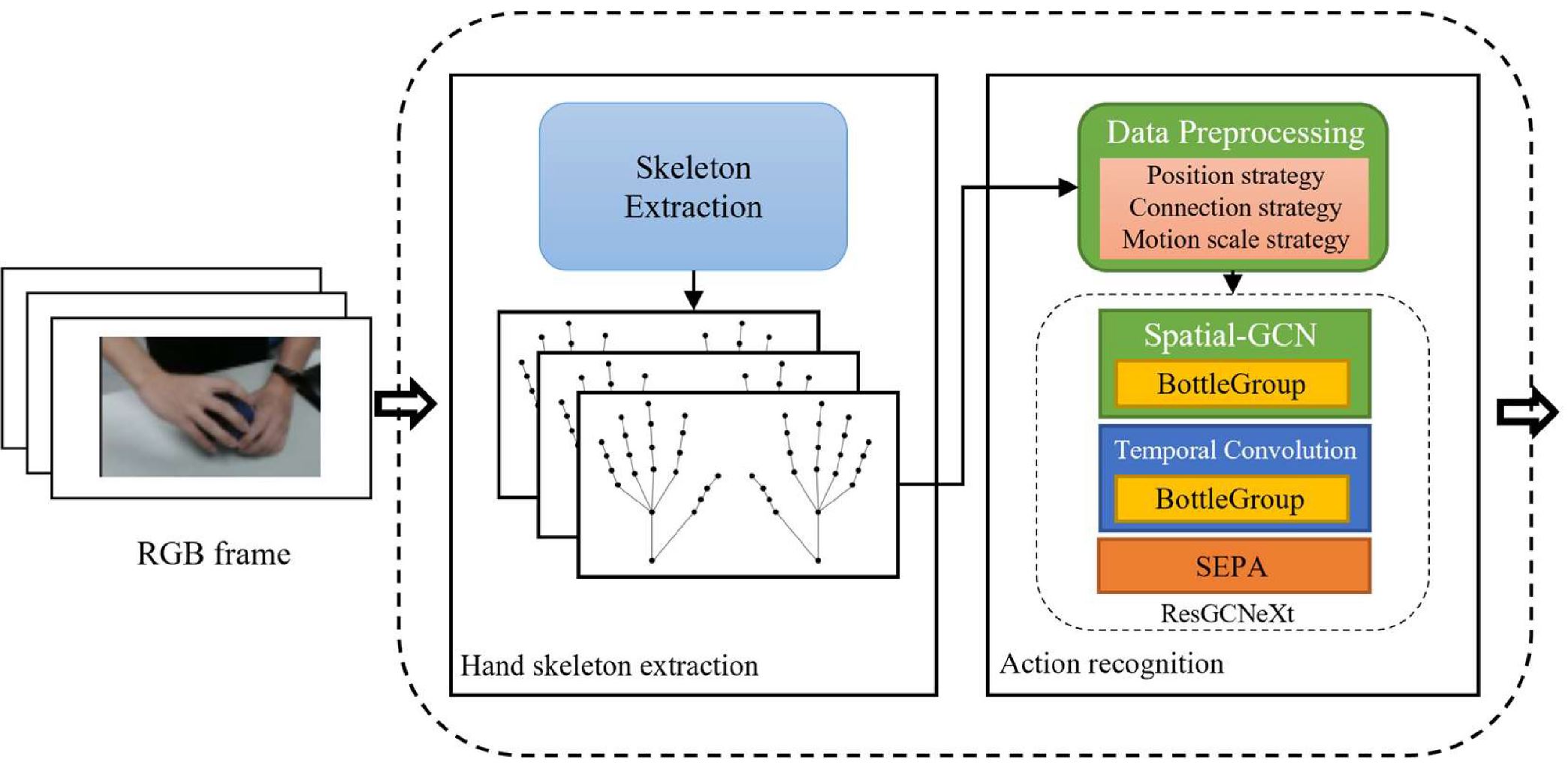}
    \caption{Extenstion GCN based model. \cite{peng2023efficient}}
    \label{fig:skelton_ResGCNeXt}
\end{figure}

\begin{figure}[htp]
    \centering
    \includegraphics[width=6cm]{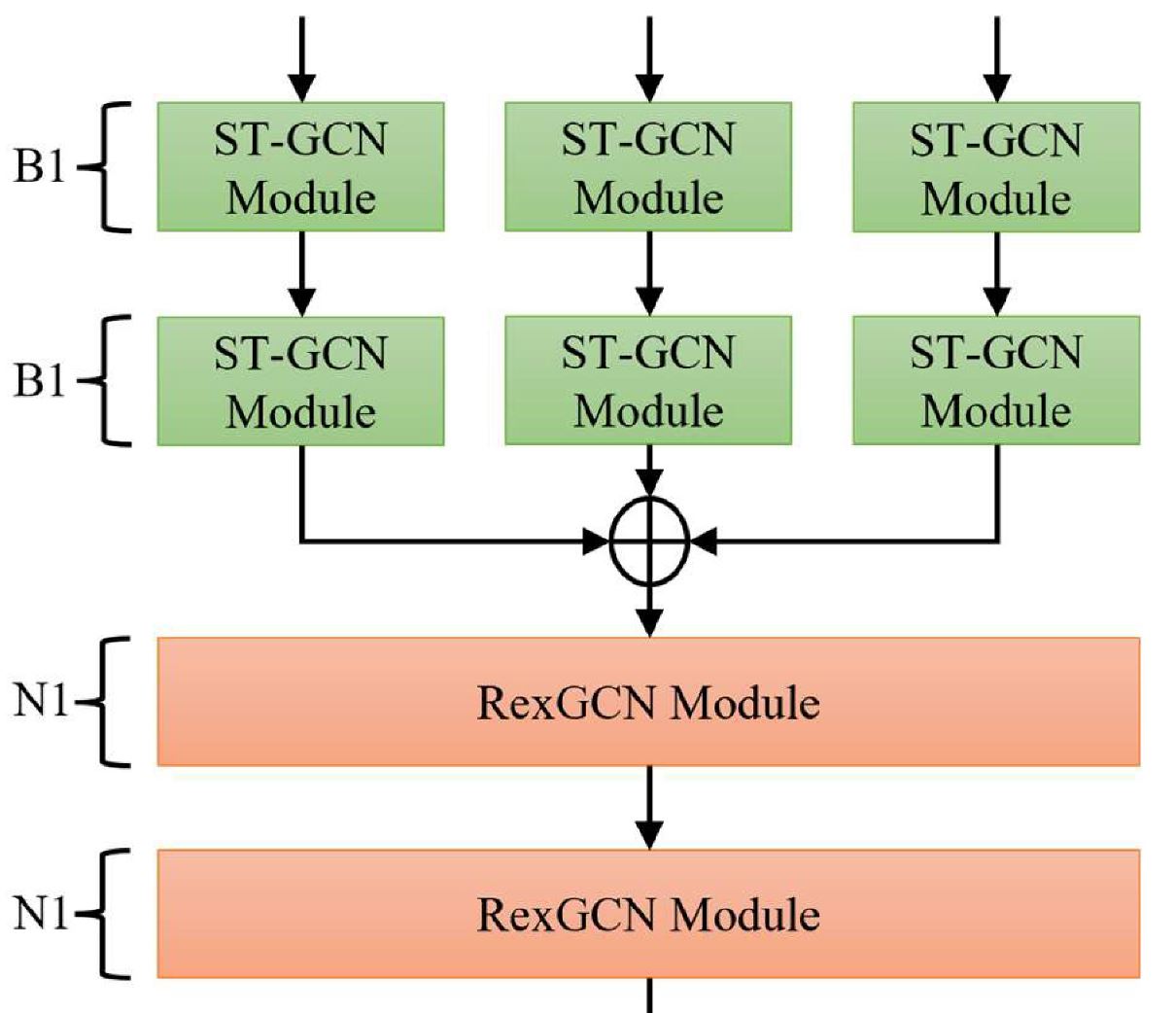}
    \caption{ResGCNeXt details architecture. \cite{peng2023efficient}}
    \label{fig:skelton_ResGCNeXt_details}
\end{figure}

\begin{figure}[htp]
    \centering
    \includegraphics[width=8cm]{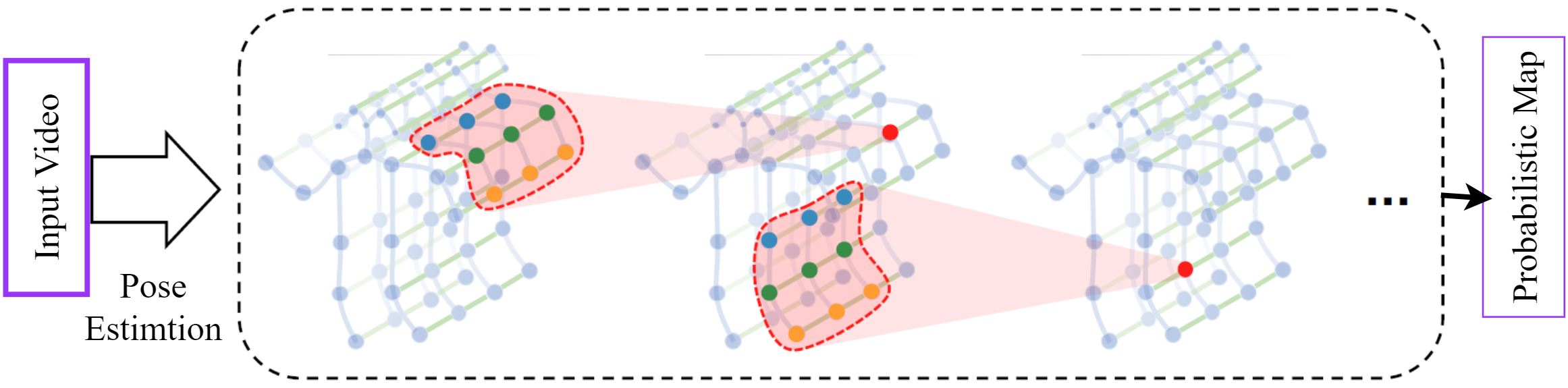}
    \caption{Spatial-temporal graph convolutional neural network (STGCN) \cite{yan2018spatial}}
    \label{fig:skelton_stgcn}
\end{figure}

\begin{figure}[htp]
    \centering
    \includegraphics[width=9cm]{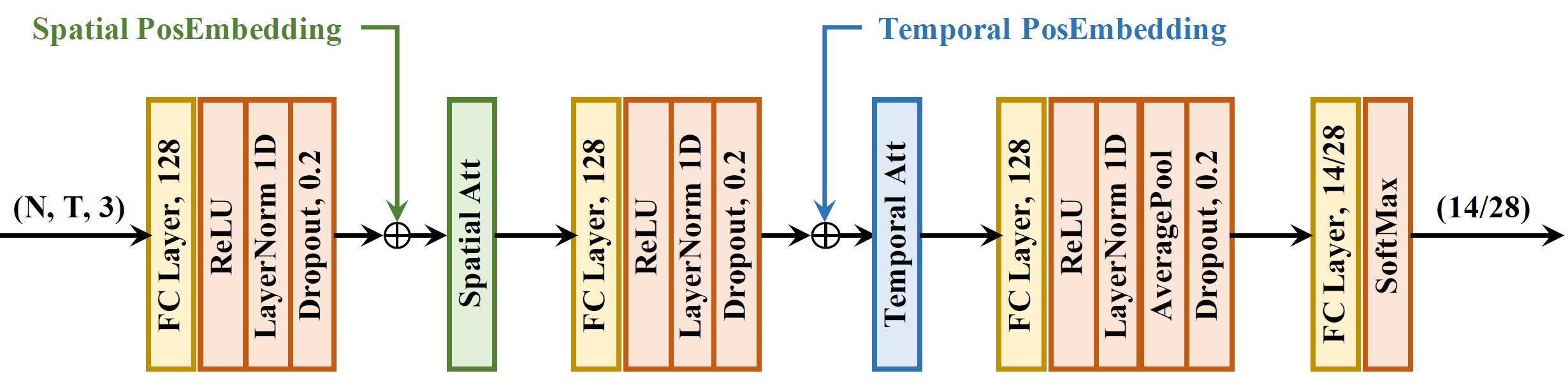}
    \caption{The demonstrated the dg-sta architecture. \cite{chen2019construct}}
    \label{fig:skelton_DG-sta}
\end{figure}

Existing systems often neglect motion and temporal features, failing to capture intricate joint relationships. Yan et al. used AlphaPose for hand skeleton joint extraction and applied Spatio-temporal GCN for gesture recognition \cite{yan2018spatial}. Recent research combines spatial-temporal attention with architectures like CNNs~\cite{si2019attention}, RNNs, and soft-attention, as well as memory attention networks (MANS) \cite{li2021memory}.
To enhance the HGR task, Song et al. utilized RNN and LSTM to extract the spatio-temporal featural features\cite{song2017end}. Hou et al. designed residual connections and a temporal convolutional neural network (STA-Res-TCN)  for improving the skeleton-based HGR performance~\cite{hou2018spatial}. Various feature levels were extracted based on the time information by using the CNN, and they reported 89.20\% and 85.00\% accuracy for 14 and 28 gestures of the DHG dataset. Moreover, 93.60\% and 90.70\% accuracy were reported for 14 and 28 gestures of the shared dataset. More recently, GCN has been used by many researchers to enhance the HGR in terms of accuracy and efficiency~\cite{yan2018spatial,chen2019construct,thakkar2018part}. While existing systems have shown success in specific cases, they still face generalization issues and often struggle to achieve high performance across diverse datasets. Yan et al. applied a temporal feature enhancement-based ST-GCN model, extracting both spatial and temporal features, as shown in Figure \ref{fig:skelton_stgcn} \cite{yan2018spatial}. More recently, Chen et al. proposed enhancing spatial and temporal features through various stages of GCN for HGR, as demonstrated in Figure \ref{fig:skelton_DG-sta} \cite{chen2019construct}.


Addressing this, Musa et al. employed a GCN combined with an attention-based spatial-temporal network to extract intricate hand skeleton relationships, achieving good performance on DHGD, SHREC, and MSRA datasets \cite{miah2023dynamic}. The working architecture of their method is illustrated in Figure \ref{fig:skelton_musa_gcndynamic}. Similarly, Shin et al. utilized a joint skeleton and joint motion-based GCN to recognize KSL, also achieving high accuracy \cite{10360810_ksl2}. Ping et al. applied a ResGCNeXt model to recognize skeleton-based dynamic HGR \cite{peng2023efficient} and reported 95.36\% and 93.45\% accuracy for Shrec and FPHA datasets, respectively, which is demonstrated in Figure \ref{fig:skelton_ResGCNeXt}. They mainly used various stages and parallel combinations of GCN demonstrated in Figure \ref{fig:skelton_ResGCNeXt_details}

 \begin{figure*}[htp]
    \centering
    \includegraphics[width=12cm]{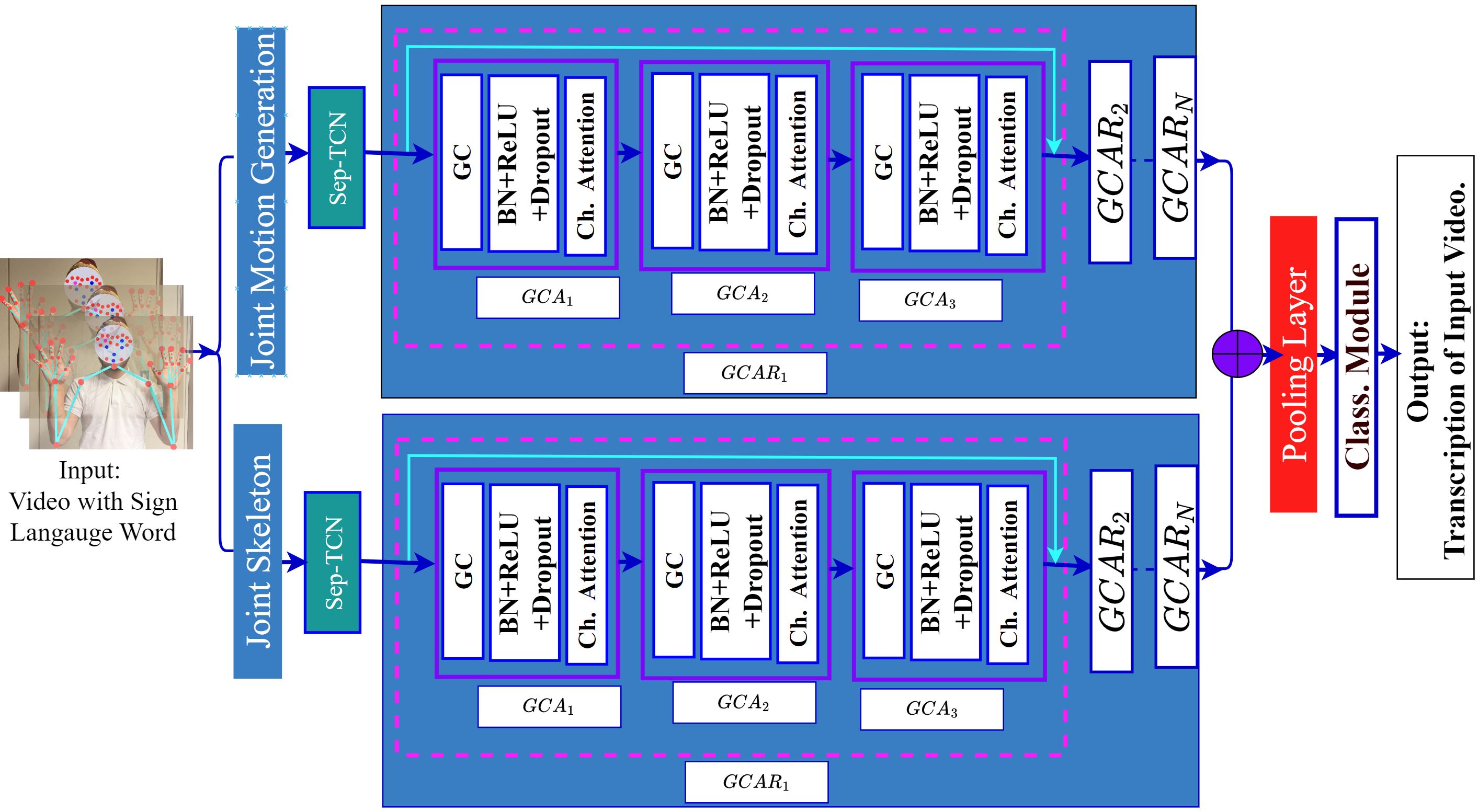}
    \caption{Sep-TCN with GCN based HGR \cite{miah2024sign_largescale}}
    \label{fig:skelton_musa_largescale}
\end{figure*}

  \begin{figure*}[htp]
    \centering
    \includegraphics[width=10cm]{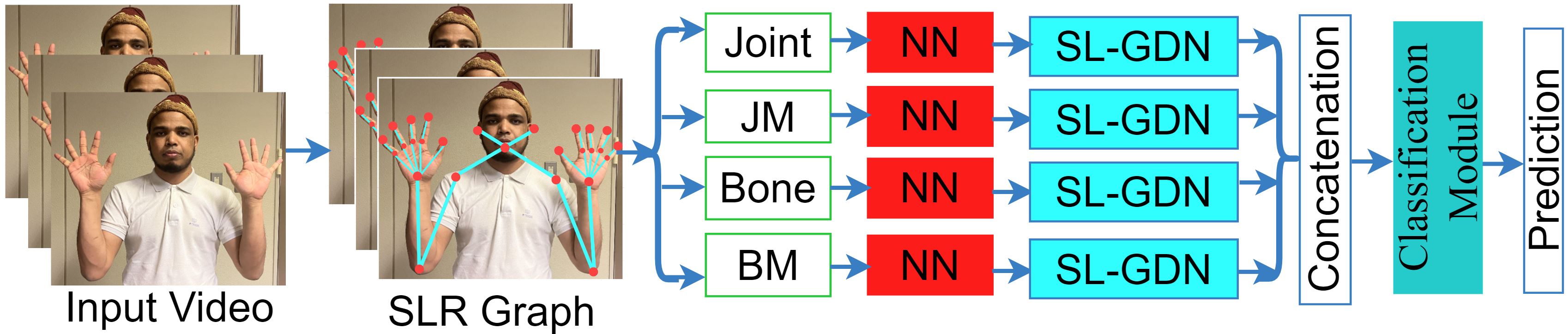}
    \caption{Fusion multi-stream based HGR \cite{electronics12132841_miah_multistream_4}}
    \label{fig:skelton_musa_multi_stream}
\end{figure*}

 Miah et al. extracted joint and joint motion information from video-based skeleton points using a two-stream GCN, achieving good performance accuracy \cite{miah2024sign_largescale}. Figure \ref{fig:skelton_musa_largescale} illustrates the architecture. To further improve performance, Miah et al. selected 27 key points out of 67 and included joint, bone, joint motion, and bone motion information from the video-based skeleton points. They employed a four-stream graph-based convolutional network, achieving enhanced accuracy. Figure \ref{fig:skelton_musa_multi_stream} shows the architecture of this method.

\subsection{Future Direction}
In 3D skeleton-based HGR, several key challenges hinder accurate and robust recognition.
Firstly, media pipe, oppose, alpha pose, and impose are are still facing challenges in achieving exact points from the RGB video. Many frames produce 0 value or skipped that make major issues to efficient work and making a good system.
Additionally, the limited availability of large-scale, diverse datasets tailored for 3D HGR impedes the development and evaluation of robust models. This scarcity restricts the training and testing of algorithms, limiting their generalizability and performance.
Furthermore, the computational complexity associated with processing and analyzing high-dimensional skeletal data in real-time applications presents practical challenges, particularly for deployment on resource-constrained devices. 

Future research in 3D skeleton-based HGR aims to overcome key challenges and advance the field. One crucial area is refining hand pose estimation algorithms to prevent skipping frames with no key point values.  Future methods would involve strong temporal contextual feature enhancement by extracting the strong relationship among the consecutive frame joints. The enhanced feature extraction algorithms with an effective classification approach would improve accuracy and reliability for precise gesture recognition in diverse scenarios. Additionally, efforts to enhance model generalization across different environments and user populations are essential, requiring the creation of larger and more diverse datasets tailored for 3D HGR.

\section{Depth Information Based Modality} \label{sect3}
Depth-based HGR utilizes depth data from sensors like Time-of-Flight (ToF) cameras, Microsoft Kinect, Intel RealSense, or Leap Motion Controller \cite{kim2015depth, de2017shrec, zengeler2018hand, alhamazani2021using}. Depth data offers several advantages over RGB images, providing 3D scene representations, robustness to lighting changes, and explicit spatial information for accurate hand tracking and gesture analysis \cite{mohaghegh2018aggregation, wang2018depth, sanchez20223dfcnn}. It captures fine details of hand movements, enhancing gesture precision \cite{chang2023exploration}. Researchers employ various ML and DL techniques for depth-based gesture recognition, as discussed below.

\subsection{Dataset}
Depth datasets are crucial for advancing Hand Gesture Recognition (HGR) and providing essential research resources. They typically contain depth images or maps captured by sensors such as ToF cameras, structured light sensors, depth cameras, and charge-coupled devices (CCD) \cite{ding2022cnn}. Many existing video RGB or skeleton datasets also contain depth information in Tables \ref{tab:rgb_video_dataset} and Table \ref{Tab:d_skeleton}. Notable depth datasets include MSR Gesture 3D, ChaLearn LAP, DHG, SHREC'17 Track, and REHAP. Specific examples include Kim et al.'s dataset \cite{kim2015depth} with gestures at different distances, Chang et al.'s BSL alphabets dataset \cite{chang2023exploration}, De Smedt et al.'s hand gestures sequences \cite{de2017shrec}, Zengeler et al.'s REHAP dataset \cite{zengeler2018hand}, and Alhamazani et al.'s infrared-based dataset \cite{alhamazani2021using}. These datasets aid in training and evaluating algorithms, driving advancements in computer vision and human-computer interaction research. The depth dataset and performance of these datasets are shown in Table \ref{tab:depth_performance}.

\begin{table*}[]
\centering
\caption{Methodological summary of the depth modality based HGR}
\label{tab:depth_performance}
\begin{tabular}{|c|c|c|c|c|c|c|c|}
\hline
\textbf{Author}   &Year  &  \textbf{ Lang.} & \textbf{Class} & \textbf{Sample} & \textbf{Feature Extraction}                                   & \textbf{Classifier} & \textbf{Perfromance[\%]} \\ \hline
Kim et al. \cite{kim2015depth}   &2015     & American SL                 & 4              & N/A             & MLBP                                                          & MLBP                & 95.63              \\ \hline
De Smedt et al. \cite{smedt20163d}  &2016   & N/A                         & 14             & 2800            & FV                                                            & SVM                 & 88.24              \\ \hline
Zengeler et al.  \cite{zengeler2018hand}  &2018  & N/A                         & 10             & 600,000         & \begin{tabular}[c]{@{}c@{}}ESF, PFH,\\  VFH\end{tabular} & LSTM                & 91.28            \\ \hline
Alhamazani et al. \cite{alhamazani2021using} &2021  & N/A                         & N/A            & 55              & Distance, Centroid                                            & SVM                 & N/A                  \\ \hline
Ding et al. \cite{ding2022cnn}  &2022      & depth-grayscale                  & 10            & N/A             & VGG-16 CNN                                                       & Softmax                & 83.88              \\ \hline
Chang et al. \cite{chang2023exploration}   &2023    & British SL                  & 450            & N/A             & CNN                                                           & CNN                 & 58.00             \\ \hline
\end{tabular}
\end{table*}

\subsection{Methodology of the Depth Modality}
There are many researchers who utilized preprocessing, feature extraction and classification using various ML and DL algorithms, which are described below: 
\subsubsection{Perprocessing approach}
 Depth images provide additional spatial information compared to RGB, aiding in gesture classification and recognition. The depth threshold method gauges the proximity of objects to the camera, determining the distance of each pixel. It then extracts an image within a designated range. This technique enhances gesture detection accuracy by pinpointing the depth range of the hands or considering the hand as the closest object. Consequently, it leads to more precise delineation of the hand area and improved preprocessing. However, this approach imposes constraints on the manner and extent of recognition.
\subsubsection{Feature Extraction and ML Approach}
Many researchers have developed HGR using feature extraction and ML algorithms with depth modality. Depth maps capture spatial information about hand shapes, movements, and configurations, enabling the extraction of key features such as hand contours, finger positions, and hand skeletons \cite{kim2015depth,de2017shrec}.
Kim et al. \cite{kim2015depth} analyzed depth images to extract hand-shape features using a modified local binary pattern (MLBP). De Smedt et al. \cite{de2017shrec} utilized 3D depth information to extract hand silhouettes, employing Fisher Vector (FV) for feature extraction. Alhamazani et al. \cite{alhamazani2021using} used hand contours and centroid location for gesture recognition. Chang et al. \cite{chang2023exploration} utilized Region of Interest (ROI) segmentation to enhance gesture detection accuracy.
Classification methods also varied: Kim et al. \cite{kim2015depth} used MLBP, De Smedt et al. \cite{de2017shrec} employed SVM with a linear kernel, and Alhamazani et al. \cite{alhamazani2021using} showed promising results with SVM. Despite successes, these ML approaches have drawbacks, relying on hand-crafted features that are time-consuming to design and may miss details in complex gestures. They often struggle with variations in hand poses, occlusions, and complex backgrounds.

\subsubsection{Features and End to END DL Based Approaches}
Recently, many researchers have been working to develop a depth modality based HGR system. Zengeler et al. \cite{zengeler2018hand} explored sensor data fusion within a DL framework for HGR, employing point cloud descriptors like Ensemble of Shape Functions (ESF), Point Feature Histograms (PFH), and Viewpoint Feature Histogram (VFH) implemented in the Point Cloud Library (PCL). Chang et al. \cite{chang2023exploration} used a CNN model for hand gesture categorization, achieving 58\% accuracy with a two-phase algorithm involving ROI segmentation and CNN categorization. This approach eliminates the need for hand-crafted feature extraction by automatically learning relevant features from raw data.
Zengeler et al. \cite{zengeler2018hand} also demonstrated that combining CNNs and LSTMs yields reliable results for HGR using depth data, with improved results when using a second ToF sensor. This combination effectively processes volumetric data and models temporal dynamics, capturing spatial and sequential information in hand gestures. Kim et al. \cite{kim2015depth} proposed an MLBP hand-tracking algorithm for real-time and accurate hand tracking, outperforming other methods in accuracy and tracking performance. De Smedt et al. \cite{de2017shrec} showed promising results with a skeleton-based approach, achieving accuracies of 88.24\% and 81.90\% for 14 and 28 gestures, respectively.
Alhamazani et al. \cite{alhamazani2021using} highlighted the efficacy of their depth camera and ML techniques in recognizing gestures with five fingers better than four. Ding et al. \cite{ding2022cnn} integrated CCD RGB-IR and depth-grayscale sensor data using CNNs for HGR, emphasizing the importance of multiple modalities. Li et al. \cite{li2023continuous} introduced an attentive 3D-Ghost module for dynamic HGR, showing the benefits of multimodal data. Gao et al. \cite{gao2021hand} developed a two-stream CNN framework for American SLR using RGB and depth data fusion, demonstrating the trend towards multimodal approaches for improved accuracy. Mahmud et al. \cite{mahmud2021deep} proposed a DL-based multimodal depth-aware system for dynamic HGR, underscoring the value of combining different input modalities for robust recognition. 

\subsection{Future Direction}
Hand gestures often involve complex movements and poses, resulting in parts of the hand being occluded from the depth sensor's view, which leads to incomplete or missing depth information. This complicates the accurate interpretation of gestures. Self-occlusions, where one part of the hand blocks another, present additional challenges, especially in real-time scenarios. HGR models need to be robust against variations in lighting conditions, cluttered backgrounds, and different viewpoints. Ensuring these models can generalize well to unseen scenarios while maintaining accurate recognition performance is crucial for practical deployment.
Future research should focus on developing a generalized HGR system based on comprehensive depth datasets. Exploring attention mechanisms and transformer architectures could improve the capture of spatial and temporal relationships within the depth data, leading to more effective feature representations.

\section{EMG Modality Based HGR} \label{sec4}
Recent advancements in HGR have leveraged electromyography (EMG) signals to overcome limitations associated with traditional RGB, skeleton, or depth datasets. The field of gesture recognition using surface Electromyography (sEMG) data has garnered significant interest across various domains such as medicine, exercise science, engineering, and prosthetic limb control \cite{jiang2012myoelectric}. Surface electromyography (sEMG) signals, which capture electrical activity generated by muscle contractions, offer a promising alternative for gesture recognition. Nevertheless, the ease of implementation remains of the EMG-based HGR a critical factor in ensuring the effectiveness of these assistive technologies \cite{li2023continuous, kim2023emg, montazerin2023transformer, kerdjidj2023implementing}.

\subsection{Dataset}
EMG modality datasets are demonstrated in Table \ref{tab:data_emg}. According to the table, several datasets are available for the EMG-based gesture domain, notably the Ninapro DB series, which consists of 9 datasets (Ninapro DB1 to Ninapro DB9) recorded using Cybergloves technology with integrated sensors \cite{stival2019quantitative_emg_data}. These datasets are commonly used by researchers to train and evaluate ML models for accurately recognizing hand and finger gestures. Ninapro DB1-DB9 are considered benchmark datasets in this field.
In the study by Kim et al. \cite{kim2023emg}, a dataset of 400 EMG data samples from 50 subjects was collected, with each sample consisting of 8 channels and spanning 1 second. The EMG data underwent min-max normalization and was reshaped into a 50 x 8 two-dimensional format for DL model training.
Côté-Allard et al. \cite{cote2017transfer} utilized two datasets: a pretraining dataset with recordings from 18 able-bodied subjects performing seven gestures for 20 seconds each, and an evaluation dataset involving 17 healthy subjects performing the same seven gestures in three rounds, each lasting 20 seconds. The study by Lee et al. \cite{lee2021electromyogram} involved ten healthy subjects performing ten hand/finger gestures, including seven individual finger (IF) gestures, resulting in a dataset for their research. Zhang et al. \cite{zhang2020novel} created a specialized dataset for model training, consisting of 30 repetitions of 21 short-term hand gestures by 13 subjects using a Myo armband. Colli Alfaro et al. \cite{colli2022user} used electromyography (EMG) and inertial measurement unit (IMU) data for user-independent gesture recognition, although specific details about the dataset size are not provided.

\begin{table*}[ht]
\centering
\caption{Summary datasets for EMG based HGR modality} \label{tab:data_emg}
\begin{tabular}{|l|l|l|l|l|l|l|l|l|}
\hline
Name &Year& Dataset Types&Sensor & No.Gesture &No. of Subjects& No. of Sample& No. Channel & \begin{tabular}[c]{@{}c@{}} Latest \\ Performance \\ Accuracy \end{tabular} \\ 
\hline
DB1 &2019 & HGR & EMG,Kinmatics& 52 & 27  & -&10 Otto Bock & 96.87 \\ \hline
DB2 &2014& HGR & EMG,Kinmatics&50& 40 & -& 12 Delsys Trigno & 92.28 \cite{duan2023alignment_aifusion}  \\ \hline
DB3&2014 & HGR  &EMG,Kinmatics& 50 & 11 & -& 12 Delsys Trigno & 91.11 \cite{duan2023alignment_aifusion} \\ \hline
DB4 &2020& HGR &EMG,Kinmatics& 52& 10 & -& 12 Cometa & 93.00 \cite{wang2024optimization} \\ \hline
DB5&2021 & HGR &sEMG, inertial& 41 & 10 & -& \begin{tabular}[c]{@{}c@{}} 16 electrodes \\2 Thalmic\\ Myo Armbands \end{tabular} &87.00 \cite{duan2023alignment_aifusion,wang2024optimization} \\ \hline
DB6 &2019& HGR &EMG,Kinmatics& 8 & 10 &-&14 Delsys Trigno & \begin{tabular}[c]{@{}c@{}}80.62 \cite{wang2024optimization}\end{tabular} \\ \hline
DB7 &2016& HGR &EMG,Inertial& 41 & 22& - & 12 Delsys Trigno & 96.76 \cite{wang2024optimization}\\ \hline
DB8&2020 & HGR & sEMG, kinematic & 9 & 12 & -& 16 Delsys Trigno & 90.00 \\ \hline
DB9&2018 & HGR & sEMG, kinematic & 40 & 77 & - & sEMG, kinematic &89.47\\ \hline
DB10&2017  & HGR &sEMG, Inertial& 10& 45 & -&12 Delsys Trigno & \begin{tabular}[c]{@{}c@{}}  80.00 \end{tabular}  \\ \hline
 
\begin{tabular}[c]{@{}c@{}} Mayo\\ Armband \end{tabular} 
&2016 & HGR& sEMG & 13  & 21 & -&8 & 88.92  \\ \hline
UC2018&2024 & HGR & sEMG& 20  & 8 & -&20 & 85.34 \\ \hline
Arm Band&2019 & HGR& sEMG & 36  & 6 & -&8 & 90.00\\ \hline
\begin{tabular}[c]{@{}c@{}} EMG-EPN\\-612\cite{BaronaLopez2024} \end{tabular}  &2022 & HGR & sEMG &5&  612  & 183600 & -&94.60 \cite{chen2023real}   \\  \hline
\begin{tabular}[c]{@{}c@{}} EMG\\-5\cite{wang2024hand} \end{tabular}  &2023 & HGR & sEMG &5&  10  & 7500 & -&89.72   \\  \hline

\begin{tabular}[c]{@{}c@{}} EMG \\High Density \end{tabular}  &2021 & HGR & sEMG &5& 41  & - & 256&81.74 \cite{wang2024optimization}   \\  \hline
\begin{tabular}[c]{@{}c@{}} DualMyo\end{tabular}  &2022 & HGR & sEMG &8& 1  &  880&--&99.00 \cite{abdelaziz2024hand}  \\  \hline
\begin{tabular}[c]{@{}c@{}} EMG36\end{tabular}  &2023 & HGR & sEMG &8& 36  &  4237907&--&97.00 \cite{abdelaziz2024hand}  \\  \hline
\begin{tabular}[c]{@{}c@{}} ISRMyo-i\end{tabular}  &2023 & HGR & sEMG &4& 6  &  -&--&85.75 \cite{leelakittisin2023enhanced}  \\  \hline
\begin{tabular}[c]{@{}c@{}} CapgMyo \\DB-b DB-c\end{tabular}  &2017 & HGR & sEMG &4& 10  &  -&--&\begin{tabular}[c]{@{}c@{}} 82.43\\ 84.75\end{tabular}\cite{leelakittisin2023enhanced} \\  \hline
\begin{tabular}[c]{@{}c@{}} SEU\end{tabular}  &2022 & HGR & sEMG &4& 20  &  -&--&\begin{tabular}[c]{@{}c@{}} 83.57\\ 84.75\end{tabular}\cite{leelakittisin2023enhanced} \\  \hline 
\end{tabular}
\end{table*}

\begin{table*}[]
\centering
\caption{Methodological review with the EMG data modality}
\label{tab:EMG_data_performance}
\begin{tabular}{|c|c|c|c|c|c|c|c|c|}
\hline
\textbf{Author}    &Year       & \textbf{\begin{tabular}[c]{@{}c@{}}Dataset\\  Name and \\ Types \end{tabular}} & \textbf{\begin{tabular}[c]{@{}c@{}}No. of \\ EMG \\ Channel\end{tabular}} & \textbf{Class} & \textbf{Sample} & \textbf{\begin{tabular}[c]{@{}c@{}}Feature \\ Extraction\end{tabular}} & \textbf{Classifier}                                  & \textbf{Performance} \\ \hline
Côté-Allard et al. \cite{catallard2019deep} &2019 & American SL                                                            & 8                                                                         & 7              & 52              & CNN                                                                    & CNN                                                  & 97.81\%              \\ \hline
Wei et al. \cite{wei2019fusion} &2019         & American SL                                                            & 16                                                                        & 41             & N/A             & CNN                                                                    & DL                                        & 90.00\%              \\ \hline
Zhang et a. \cite{zhang2019two} &2019       & American SL                                                            & 8                                                                         & 20             &                 & RNN                                                                    & RNN                                                  & 89.60\%              \\ \hline
Yang et al. \cite{yang2023real} &2023        & American SL                                                            & 16                                                                        & 52             & 10000           & MResLSTM                                                               & MResLSTM                                             & 93.52\%              \\ \hline
Lee et al. \cite{lee2021electromyogram} &2021         & American SL                                                            & 3                                                                         & 7              & N/A             & \begin{tabular}[c]{@{}c@{}}RMS, VAR\\ MAV, SSC \\ ZC, WL\end{tabular}  & ANN                                                  & 94.00\%              \\ \hline
Colli Alfaro et al.\cite{colli2022user} &2022& American SL                                                            & 5                                                                         & 7              & N/A             & \begin{tabular}[c]{@{}c@{}}MAV, MAVS\\ WL, AR, ZC\end{tabular}         & \begin{tabular}[c]{@{}c@{}}LS-SVM\\ MLP\end{tabular} & 92.90\%              \\ \hline

Cruz et al. \cite{cruz2023deep} & 2023 & EMG-IMU-EPN-100+ & 12 & 6 & \begin{tabular}[c]{@{}c@{}}300\end{tabular} &  DQN & Softmax & \begin{tabular}[c]{@{}c@{}}97.45 \\ 88.05\end{tabular} \\ \hline

Kim et al. \cite{kim2023emg} & 2023 & Myo armband  & 8 & 10 & \begin{tabular}[c]{@{}c@{}}37000\end{tabular} &  CRNN & Softmax& \begin{tabular}[c]{@{}c@{}}96.57\% (training) \\ 95.10\% (testing )\end{tabular} \\ \hline
Kerdjidj et al. \cite{kerdjidj2023implementing} & 2023 & Public EMG dataset &8 & 36 & 40,000  &  Temporal feature & k-NN & 98\% \\ \hline

Xion et al. \cite{xiong2023global} & 2023 & \begin{tabular}[c]{@{}c@{}}NinaPro DB4, \\ NinaPro DB5, \\ DB1-DB3, \\ Mendeley \end{tabular} & Various & - & - & - & GLF-CNN & \begin{tabular}[c]{@{}c@{}}88.34\% (average) \\ 91.4  \\ 91.0 \\ 88.6\end{tabular} \\ \hline

Leelakittisin et al.  \cite{leelakittisin2023enhanced} & 2023& \begin{tabular}[c]{@{}c@{}}CapgMyo\\ ISRMyo-I\\ SEU \end{tabular} &- & 4 & - &\begin{tabular}[c]{@{}c@{}}Enhanced \\Lightweight\\ CNN with\\ JCAP\end{tabular}   &Softmax&\begin{tabular}[c]{@{}c@{}}CapgMyo DB-b: 82.43, \\ CapgMyo DB-c: 84.75, \\ ISRMyo-I: 85.75, \\ SEU: 83.57\end{tabular} \\ \hline

Chen et al. \cite{chen2023real}  & 2023 & EMG-EPN-612 & 6 & - & - &  \begin{tabular}[c]{@{}c@{}}Neural Feature\\ Extraction\end{tabular} & \begin{tabular}[c]{@{}c@{}}Supervised \\learning\end{tabular} &\begin{tabular}[c]{@{}c@{}}94.60\end{tabular} \\ \hline

Zhang et al. \cite{zhang2023lightweight} & 2023 & \begin{tabular}[c]{@{}c@{}}HGR1, \\ OUHANDS\end{tabular} & \begin{tabular}[c]{@{}c@{}}25, \\ 10\end{tabular} & \begin{tabular}[c]{@{}c@{}}12, \\ 23\end{tabular} & \begin{tabular}[c]{@{}c@{}}899, \\ 3,000\end{tabular} &  \begin{tabular}[c]{@{}c@{}}BaseNet, \\ MSS, \\ LAS\end{tabular} & LHGR-Net & \begin{tabular}[c]{@{}c@{}}HGR1: 93.36\%, \\ OUHANDS: 98.57\%\end{tabular} \\ \hline
Xu et al. \cite{xu2023novel} & 2023 & \begin{tabular}[c]{@{}c@{}} DB4 \\  DB5\end{tabular} & 53 & \begin{tabular}[c]{@{}c@{}}12 \\ 16\end{tabular} & \begin{tabular}[c]{@{}c@{}}30 tr\end{tabular} &  SE-CNN&Softmax & \begin{tabular}[c]{@{}c@{}} 89.54 \\ 77.61\end{tabular} \\ \hline
Zabihi et al. \cite{zabihi2023light} & 2023 &  DB2 & 12&17 & 30 tr & \begin{tabular}[c]{@{}c@{}}HDConv \\ MHSAtten\end{tabular} & HDCAM & \begin{tabular}[c]{@{}c@{}}XXSmall: 81.73\% \\ XSmall: 82.61\% \\ Small: 82.91\%\end{tabular} \\ \hline
Zabihi et al. \cite{zabihi2023trahgr} & 2023 & DB2 & 12 & 9 & \begin{tabular}[c]{@{}c@{}}30\end{tabular} &Tnet,Fnet  &TraHGR & \begin{tabular}[c]{@{}c@{}} 93.84\%\end{tabular} \\ \hline

Kim et al. \cite{kim2023emg}  &2023        & American SL                                                            & 8                                                                         & 10             & 50              & CNN                                                                    & CRNN                                                 & 96.04\%              \\ \hline

Baroni et al. \cite{BaronaLopez2024}  &2024        & HandGesture                                                            & 5-612                                                                         &              &  183,600               & CNN-LSTM                                                                    & CRNN                                                 & 90.45\%              \\ \hline

Wang et al. \cite{wang2024hand} & 2024 & EMG Dataset & 5 & 10 & 7500 &  BP Neural Network&Softmax & 89.72\% (4 channels) \\ \hline

Li et al. \cite{zhang2024electromyographic} & 2024 & \begin{tabular}[c]{@{}c@{}} NinaPro DB5\\ Myo \end{tabular} & 10 &  - & 7500 &   \begin{tabular}[c]{@{}c@{}} multi-attention\\ (CNN,channel \\spatial-temporal)\end{tabular}&Softmax & 91.64\%  \\ \hline

Wang et al. \cite{wang2024optimization} & 2024 & \begin{tabular}[c]{@{}c@{}}High-density sEMG, \\ Ninapro DB4, \\ Ninapro DB5\end{tabular} & 10 &  - & 7500 &   \begin{tabular}[c]{@{}c@{}}RMS\\WL\\ ZC\\ SSC \\ CDEM\end{tabular}&\begin{tabular}[c]{@{}c@{}}Simplified\\ Cross-Domain\\ Error \\Minimization\\ (CDEM)\end{tabular}  &  \begin{tabular}[c]{@{}c@{}}81.74  \\ 93.50 , \\ 84.00 \end{tabular}  \\ \hline

Abdelaziz et al. \cite{abdelaziz2024hand} & 2024 & \begin{tabular}[c]{@{}c@{}}DualMyo\\ EMG36\end{tabular} & \begin{tabular}[c]{@{}c@{}}8 \\ 8\end{tabular} &   \begin{tabular}[c]{@{}c@{}}8 \\ 8\end{tabular} &  \begin{tabular}[c]{@{}c@{}}880\\ 4237907\end{tabular} &   \begin{tabular}[c]{@{}c@{}}CNN+LSTM\end{tabular}&\begin{tabular}[c]{@{}c@{}}Softmax\end{tabular}  &  \begin{tabular}[c]{@{}c@{}}99\% (DualMyo), \\ 97\% (EMG36)\end{tabular}   \\ \hline
Vasconez et al. \cite{vasconez2023comparison} & 2024 & EMG-EPN-612 & 6 & - & - & - & \begin{tabular}[c]{@{}c@{}}Supervised \\learning, \\ Reinforcement \\learning\end{tabular} & \begin{tabular}[c]{@{}c@{}}93.57\% (supervised)\end{tabular} \\ \hline

\end{tabular}
\end{table*}

\subsection{EMG Based Methodology}
Utilizing Electromyography (EMG) signals for HGR involves capturing muscle activity to interpret gestures \cite{miah2019eeg,joy2020multiclass,miah2020motor,zobaed2020real,miah2021event}. This methodology typically includes signal acquisition, feature extraction, and classification algorithms to decode gestures. Analyzing EMG patterns enables real-time and intuitive interaction in applications like prosthetics and human-computer interfaces. Leveraging ML and DL enhances many researchers who have been working to develop EMG-based HGR by enabling complex pattern recognition. Table \ref{tab:EMG_data_performance} demonstrated the existing HGR methodology and its performance. 

\subsubsection{Preproecssing Filtering and Signal Segmentation}
Filtering, Motion detection, frequency controlling, and noise reduction are the most crucial preprocessing in this data modality. Adhering to the preprocessing methodology outlined in earlier investigations~\cite{wei2019multistream,atzori2016deep}, a 1st-order low-pass Butterworth filter was employed to dampen electrical activity data of muscles. Additionally,  $\mu-law$ transformation amplifies the output of sensors with smaller magnitudes logarithmic while maintaining the scale of sensors with larger values over time. The $\mu-law$ transformation method is traditionally used for the purpose of quantisation in the speech or telecommunication fields. Many researchers used this method as a pre-processing tool to scale sEMG signals.

\subsubsection{ML Based Approach with Mathematical and Statistical Features.}
Many researchers used statistical and mathematical feature extraction techniques. Among them, Lee et al. \cite{lee2021electromyogram} utilize six time-domain (TD) features, including root mean square (RMS), variance (VAR), mean absolute value (MAV), slope sign change (SSC), zero crossing (ZC), and waveform length (WL). In the research presented in \cite{colli2022user}, the mean absolute value (MAV), mean absolute value slope (MAVS), waveform length (WL), 4th-order auto-regressive coefficients (AR), and zero crossings (ZC) are extracted from each EMG channel. Then these mathematical features feed into the Traditional ML techniques like Linear Discriminant Analysis (LDA) \cite{esposito2020piezoresistive, tavakoli2018robust} and SVM \cite{esposito2020piezoresistive} have been utilized for HGR using sEMG signals. While classical pattern recognition methods have been extensively studied, advanced techniques have not been widely implemented commercially due to challenges in addressing real-world complexities and variations in sEMG signals.
Lee et al. \cite{lee2021electromyogram} experiment with four ML methods, including artificial neural network (ANN), SVM, random forest (RF), and logistic regression (LR), to build personalized classifiers for gestures.

\subsubsection{CNN Based Approaches}
The size of the EMG signal-based dataset has increased because the number of channels and time of the EMG signal is larger than the previous dataset, and it is difficult to get good performance accuracy with the traditional ML algorithm. To overcome the issues, many researchers have been working to apply end-end DL technology. Among them,  Wei et al. \cite{wei2019surface} utilize DL techniques for gesture classification. Kim et al. \cite{kim2023emg} employ a CNN for time-domain feature extraction. To capture intricate features, the number of filters increases across neural layers, specifically to 16, 32, and 64. Recently, researchers used CNN \cite{cote2017transfer, wei2019surface,zhang2020novel}, recurrent neural network (RNN),multi-stream residual network (MResLSTM) \cite{zhang2020novel}.
Côté-Allard et al. \cite{cote2017transfer} employ a CNN architecture for classification, which is further improved using transfer learning techniques to enhance accuracy. Kim et al. \cite{kim2023emg} adopt a convolutional recurrent neural network (CRNN) structure for training and testing hand gesture classification.

\subsubsection{RNN-LSTM Based Methods}
Deep Neural Networks (DNNs) trained on subsets of Ninapro data demonstrate promising performance on unseen repetitions \cite{catallard2019deep}. However, accuracy diminishes for less explored repetitions, necessitating extensive training. Domain adaptation techniques like Transfer Learning (TL) and domain alignment aim to enhance model robustness across users and gestures \cite{catallard2019deep,cote2017transfer}. Zhang et al. \cite{zhang2020novel} make use of a recurrent neural network (RNN) model specifically designed to learn from raw surface electromyographic (sEMG) data and predict hand gestures in real time. 
Yang et al. \cite{zhang2020novel} introduce a multi-stream residual network (MResLSTM) for dynamic gesture recognition, leveraging surface EMG (sEMG) signals. The MResLSTM model integrates both the residual model and convolutional short-term memory components to classify various types of gestures. 

\subsubsection{Attention, TCN and GNN Based Methods}
\begin{figure*}[htp]
    \centering
    \includegraphics[width=10cm]{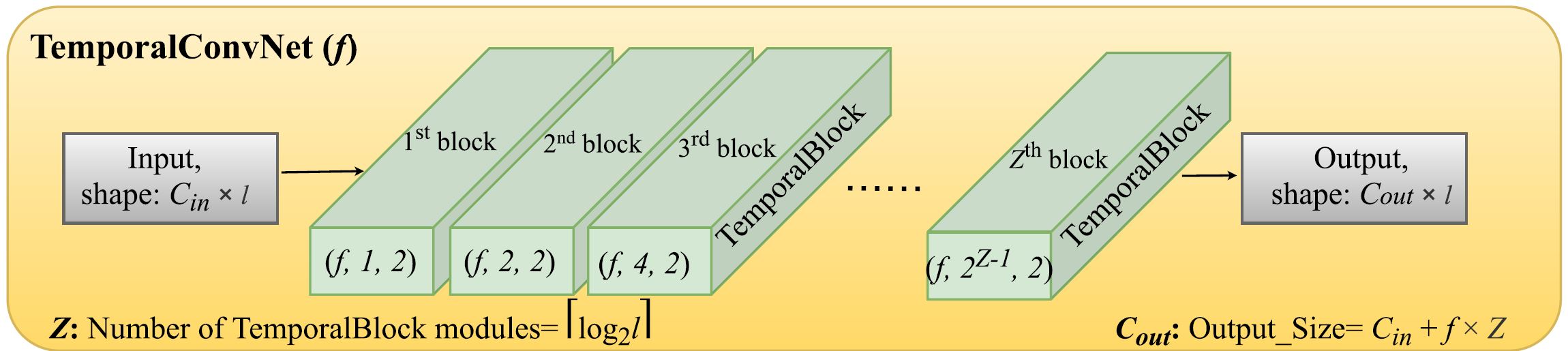}
    \caption{Temporal convolutional network (TCN) module\cite{tsinganos2019improved_TCN}}
    \label{fig:emg_tcn}
\end{figure*}

\begin{figure*}[htp]
    \centering
    \includegraphics[width=12cm]{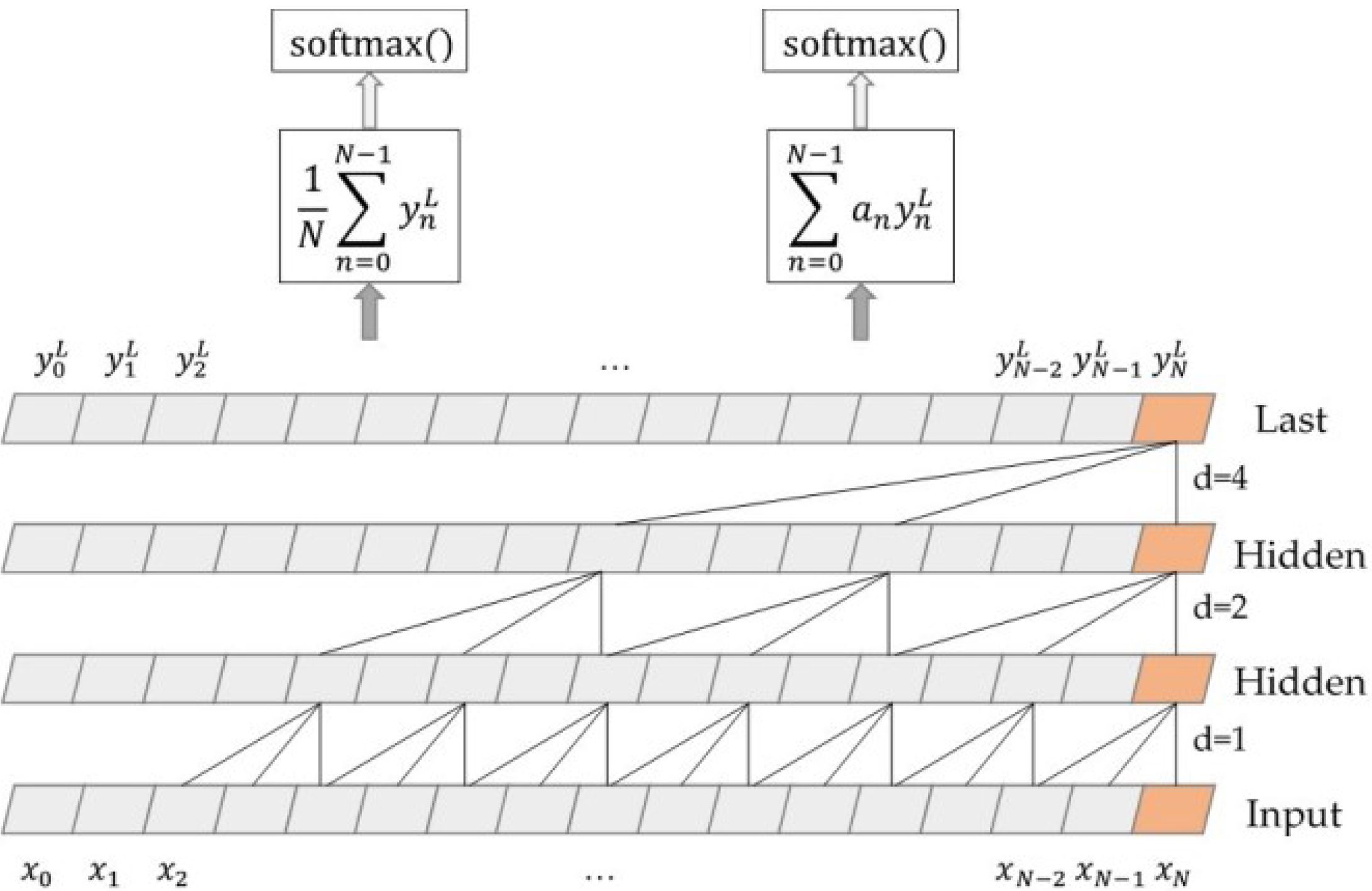}
    \caption{TCN internal process sructure\cite{tsinganos2019improved_TCN}}
    \label{fig:emg_tcn_internal_description}
\end{figure*}

\begin{figure*}[htp]
    \centering
    \includegraphics[width=9cm]{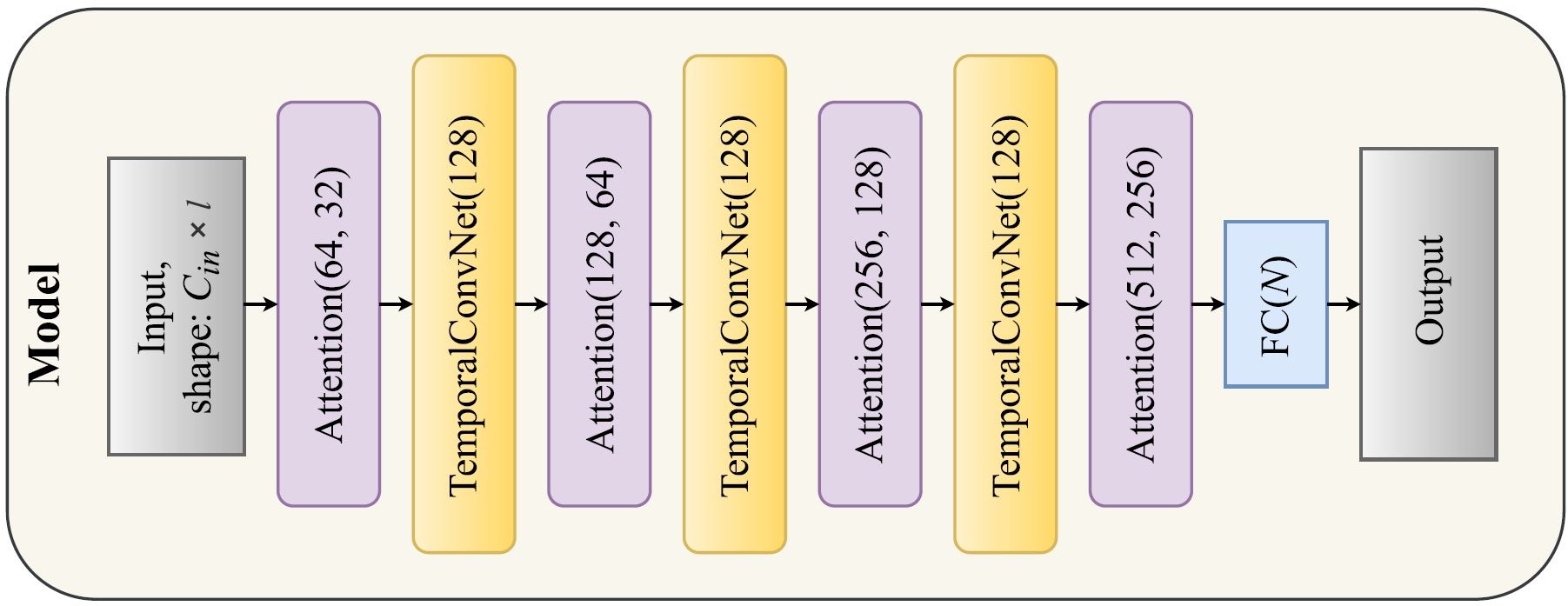}
    \caption{TCN integrating with attention module\cite{rahimian2021fs_TCN_Attention}}
    \label{fig:emg_tcn_attention}
\end{figure*}

\begin{figure*}[htp]
    \centering
    \includegraphics[width=12cm]{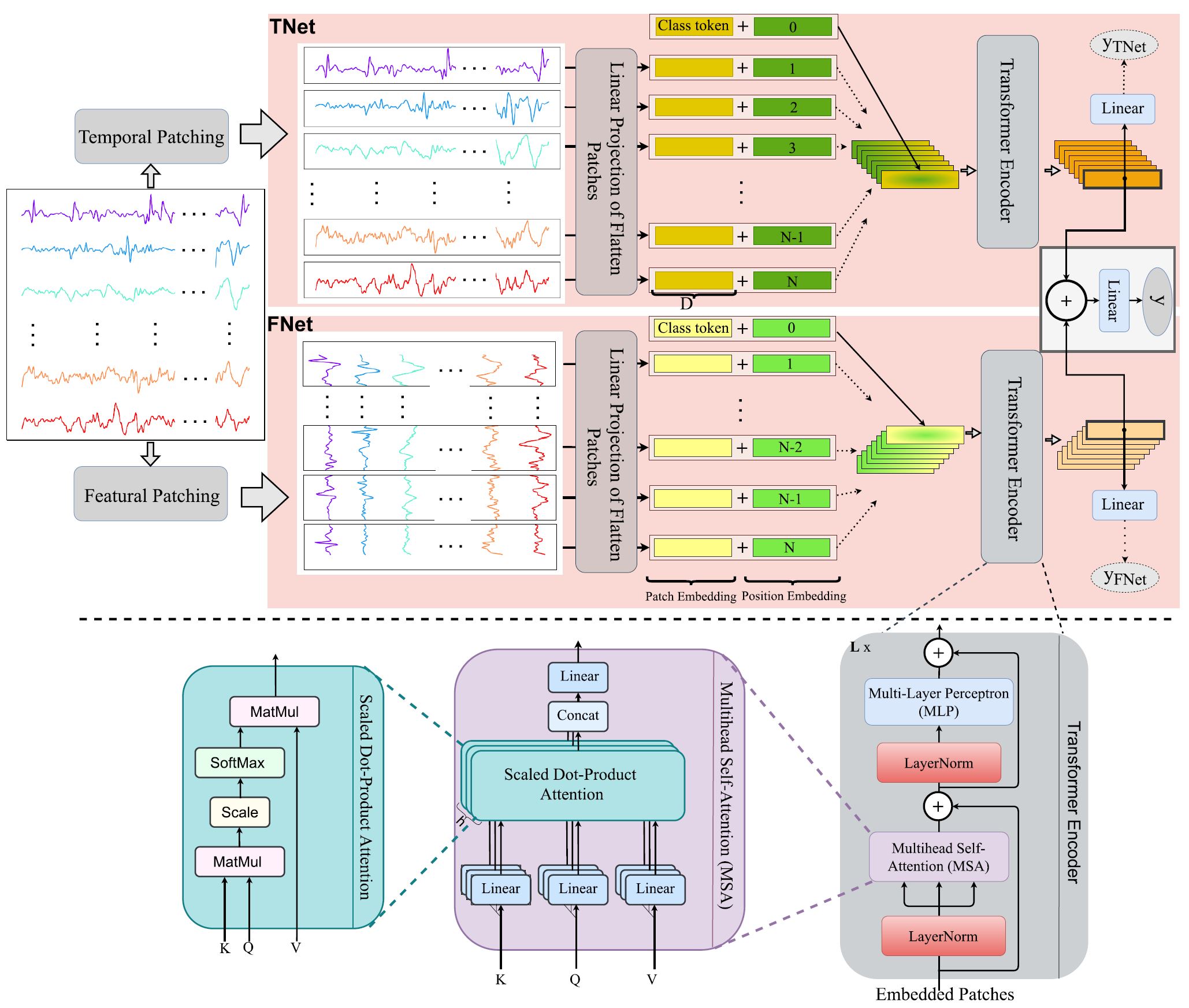}
    \caption{TrHGR model details\cite{zabihi2023trahgr}}
    \label{fig:emg_TrHGR}
\end{figure*}

As the spatial enhancement feature extracted with CNN is not so effective researchers found that Temporal modelling is a very crucial feature for EMG signal-based HGR. Recently, Tsinganos et al. applied a TCN module aiming to sequence the classification problem of HGR \cite{tsinganos2019improved_TCN}. After evaluating their model with the Ninapro DB1 dataset, they reported 89.76\% top-1 accuracy. Figure \ref{fig:emg_tcn} demonstrated the TCN module architecture. They also reported a good performance accuracy, but only the temporal feature is not so effective in generalized cases, and it may face difficulties in achieving good performance accuracy due to lacking feature effectiveness. Figure \ref{fig:emg_tcn_internal_description} demonstrated the temporal model of this work. To overcome the problems, Rahimian et al. proposed a few-short learning method by integrating TCN with attention module \cite{miah2023dynamic}.   They emphasize the challenges faced by Deep Neural Networks (DNNs) when dealing with limited data and introduce a novel approach known as Few-Shot learning- HGR (FS-HGR) \cite{rahimian2021fs_TCN_Attention}. This method aims to enhance gesture detection accuracy by minimizing the reliance on extensive training data. The FS-HGR model integrates temporal convolutions and attention mechanisms, allowing it to generalize effectively with minimal training instances. By leveraging Few-Shot learning, FS-HGR efficiently infers outputs based on a small number of training observations, making it a practical and promising solution for real-life applications where large datasets are not readily available. Figure \ref{fig:emg_tcn_attention} demonstrated the diagram of this module. Zahihi et al. proposed a Hierarchical Depth-wise Convolution and Attention Mechanism (HDCAM) model to recognize EMG-based HGR where they reported 82.91\% accuracy with DB2 dataset \cite{zabihi2023light}. Zahibe et al. again proposed a TraHGR: Transformer for HGR module \cite{zabihi2023trahgr} where they reported 93.84\% accuracy for the Ninapro DB2 dataset and Figure \ref{fig:emg_TrHGR} demonstrated the TrHGR details. TrAGR is mainly composed of Tnet and Fnet, aiming to extract temporal and frequency domain features and make a hybrid framework based on the transformer architecture.

Gestures recognition with sEMG data benefits from domain adaptation techniques \cite{rehman2018multiday,du2017surface}. However, challenges persist in adapting to varied signal spaces and postures, and performance with limited training data remains uncertain \cite{rehman2018multiday}. 
Studies incorporating domain adaptation techniques and DL algorithms address the challenge of inter-session classification in sEMG-based HGR \cite{rehman2018multiday,du2017surface}. However, their adaptability to signal variability and performance with limited data and varied subjects remain uncertain \cite{rehman2018multiday}.
Côté-Allard et al. \cite{cote2017transfer} employ transfer learning techniques to enhance the accuracy of their DL-based approach to classification.

Colli Alfaro et al. \cite{colli2022user} employ a sensor fusion technique that combines electromyography (EMG) data and inertial measurement unit (IMU) data for user-independent gesture classification. They utilize three different classification methods: Adaptive Least Squares Support Vector Machines (LS-SVM), Bilinear Model-based classification, and Multilayer Perceptron (MLP) Network. 

\subsection{Future Direction}
Recognizing hand gestures using EMG signals faces significant challenges due to variability between individuals and within the same person over time. Factors like muscle fatigue, electrode placement, and skin condition impact signal consistency, complicating universal model effectiveness. Noise in EMG data from movement or environmental interference further hampers accurate gesture recognition, necessitating effective noise reduction techniques. Speed is crucial for applications like prosthetic control or computer interfaces, requiring rapid, delay-free recognition, which is challenging for devices with limited computing power.

Future research should focus on adaptive, personalized models that continuously learn and adjust to individual EMG signal variations, employing techniques like transfer learning. This can lead to more accurate and user-friendly systems for assistive technologies, virtual reality, and human-robot collaboration.

\section{Audio Signal Based Modality} \label{sect5}
Audio signals, representing sound waves captured via microphones or digital recordings, offer valuable cues for HGR. They provide contextual information and supplement visual data, enhancing system robustness in challenging environments. Many researchers have been working to develop audio signal-based HGR using various ML and DL tools. Below, we discuss the dataset, methodology, current challenges and future direction of these modalities. 

\subsection{Dataset}
Few researchers have been working to develop an HGR system using audio signals. Figure \ref{tab:audio_performance} demonstrated the audio modality dataset and its performance.  Saad et al. \cite{saad2018ultrasonic} comprise reflected ultrasonic signals obtained from one transmitter and four receivers, enabling gesture detection. Siddiqui et al. \cite{siddiqui2017wearable} collected hand gesture data from three right-handed subjects performing ASL alphabets, numbers, and relaxation gestures. Sang et al. \cite{sang2018micro} utilized a dataset with hand gesture samples from nine subjects, each performing six gestures. Luo et al. \cite{luo2020hci} experimented with acoustic signals for device-free gesture recognition, testing various surface materials' impact on recognition. Ling et al. \cite{ling2020ultragesture} employed Channel Impulse Response (CIR) measurements for gesture recognition, providing a resolution of 7 mm. Wang et al. \cite{wang2020push} converted CIR measurements into CIR images for gesture representation. Lastly, Wang et al. \cite{wang2023ultrasonicgs} gathered data from 12 volunteers performing single motions and sign language movements under different impacting factors.

\begin{table*}[]
\centering
\caption{Methdological review of the Audio-based HGR}
\label{tab:audio_performance}
\begin{tabular}{|c|c|c|c|c|c|c|c|}
\hline
\textbf{Author}   &Year      & \textbf{Supported Language} & \textbf{Class} & \textbf{Sample} & \textbf{\begin{tabular}[c]{@{}c@{}}Feature \\ Extraction\end{tabular}}                                 & \textbf{Classifier}                                       & \textbf{Performance [\%]} \\ \hline
Saad et al.  \cite{saad2018ultrasonic} &2018    & American SL                 & 7              & 128             & Amplitude                                                                                              & SVM                                                       & 96.00              \\ \hline
Wang et al. \cite{wang2018depth} &2018     & American SL                 & 15             & 312             & CNN                                                                                                    & LSTM                                                      & 98.40              \\ \hline
Wang et al. \cite{wang2018depth}  &2018    & American SL                 & 15             & N/A             & CNN                                                                                                    & Bi-LSTM                                                   & 98.80              \\ \hline

Sang et al. \cite{sang2018micro}  &2018    & American SL                 & 6              & 2700            & \begin{tabular}[c]{@{}c@{}}Moving   Trajectories, \\    Precise   Ranges,\\    Velocities\end{tabular} & SVM                                                       & 96.32              \\ \hline

Siddiqui et al. \cite{siddiqui2020multimodal}&2020  & American SL                 & 36             & N/A             & \begin{tabular}[c]{@{}c@{}}RMS,   MAV\\    Skew,  Kurtosis\end{tabular}                               & \begin{tabular}[c]{@{}c@{}}SVM, DT,\\    LDA\end{tabular} & 80.00              \\ \hline
Luo et al. \cite{luo2020hci} &2020      & American SL                 & 7              & 256             & STE,   ZCR                                                                                             & SVM                                                       & 93.20\%              \\ \hline
Ling et al. \cite{ling2020ultragesture} &2020     & American SL                 & 12             & 480             & CNN                                                                                                    & CNN                                                       & 99.60              \\ \hline

\end{tabular}
\end{table*}

\subsection{Audio Signal Based Methodology}
Table \ref{tab:audio_performance} demonstrated a summary of the existing technologies used to implement an audio signal-based HGR system, including year, dataset information,  feature extraction and classification method besides performance. In this section, we explain details about the ML and DL-based HGR system with Audio data modality. 
\subsubsection{ML-Based Methods}
Table \ref{tab:audio_performance} demonstrated various existing models for audio-based HGR. Saad et al. \cite{saad2018ultrasonic} developed an ultrasonic gesture recognition system using SVM classification, achieving a gesture detection sensitivity and specificity of 99\% each, with a classification accuracy of 96\%. Siddiqui et al. \cite{siddiqui2017wearable} explored HGR through acoustic measurements, achieving an average classification accuracy exceeding 80\% using Linear Discriminant Analysis (LDA). Sang et al. \cite{sang2018micro} compared the HMM and end-to-end neural network methods, achieving accuracies of 89.38\% and 96.34\%, respectively.
Luo et al. \cite{luo2020hci} developed an HCI mechanism achieving a gesture recognition accuracy of 93.2\% for seven common gestures on smart devices.  Wang et al.'s RobuCIR system \cite{wang2020push} demonstrated high accuracy and robustness in recognizing 15 gestures, outperforming existing approaches. Wang et al. \cite{wang2023ultrasonicgs} achieved a combined recognition rate of 98.8\% for single gestures and maintained high accuracy for continuous and sign language gestures.

\subsubsection{CNN, RNN-LSTM Based Methods}
Wang et al. \cite{wang2020push} integrated CNN and LSTM networks, demonstrating high accuracy and robustness in recognizing 15 gestures. Wang et al. \cite{wang2023ultrasonicgs} used the CTC algorithm and Bi-LSTM network, achieving a combined recognition rate of 98.8\% for single gestures and maintaining high accuracy for continuous and sign language gestures.
 Ling et al. \cite{ling2020ultragesture} introduced UltraGesture with an average accuracy exceeding 99\% for 12 gestures, leveraging Channel Impulse Response (CIR) measurements and a CNN model.

\subsection{Challenges and Future Direction}
Audio-based HGR systems face challenges such as ambient noise interference, varying acoustic conditions, limited gesture vocabulary, user dependency, privacy concerns, and integration difficulties. These systems must handle environmental noise, adapt to different acoustic settings, recognize diverse gestures, accommodate user variations, address privacy issues, and integrate smoothly into existing devices. Improvements in signal processing, noise reduction, ML, and user interface design are essential to enhance robustness, accuracy, and usability for broader adoption.
 Future work should integrate advanced algorithms like TCN-LSTM networks, develop attention-based temporal modelling, expand recognition to complex gestures, and explore novel data augmentation techniques for improved robustness.
 
\section{EEG Modality Based HGR} \label{sec6}
EEG-based HGR measures brain activity using scalp electrodes, and they try to use different hand movements to create unique neural patterns. The process includes acquiring EEG data, noise removal, feature extraction, and training algorithms like support vector machines or neural networks to link EEG patterns to gestures. Real-time recognition, crucial for brain-computer interfaces, involves continuous data acquisition and processing. Challenges include low spatial resolution, non-stationary signals, inter-subject variability, and signal artefacts. Research focuses on advanced signal processing, better algorithms, and integrating modalities like EMG or motion tracking. These advancements could enhance human-computer interaction and expand applications in assistive technologies, virtual reality, and human-robot collaboration.

\subsection{Dataset}
Table \ref{tab:EEG_peformance_table} demonstrated the dataset information and performance of the EEG modality. D. AlQattan et al. \cite{alqattan2017towards} used an Enobio wireless system to acquire EEG data, utilizing sixteen of the twenty available channels due to connectivity issues with channels Pz, O1, and O2. The data was sampled at a rate of 500 samples per second. 

\begin{table*}[ht]
\centering
\caption{Databases and performance accuracy with EEG modality for existing model}
\label{tab:EEG_peformance_table}
 \begin{tabular}{|c|c|c|c|c|c|c|c|c|c|c|}
\hline
\textbf{Author} & Year & \textbf{\begin{tabular}[c]{@{}c@{}}Dataset\\Name\\ and \\Type \end{tabular}} & \textbf{Classes} & Subject & \textbf{Sample} & \textbf{\begin{tabular}[c]{@{}c@{}}No. of \\ EEG \\ Channel\end{tabular}} & \textbf{\begin{tabular}[c]{@{}c@{}}Feature \\ Extraction\end{tabular}} & \textbf{Classifier} & \textbf{Performance} \\ \hline
AlQattan et al. \cite{alqattan2017towards} & 2017 & American SL & 6 & - & 500 & 16 & \begin{tabular}[c]{@{}c@{}}Energy, Entropy,\\ SD\end{tabular} & \begin{tabular}[c]{@{}c@{}}SVM and\\ LDA\end{tabular} & 75.00 \\ \hline
Chaves et al. \cite{bozal2017personalized} & 2017 & American SL & 40 & - & 50 & 128 & LSTM & \begin{tabular}[c]{@{}c@{}}Deep \\ Learning\end{tabular} & 90.34 \\ \hline
Spampinato et al.\cite{spampinato2017deep} & 2017 & American SL & 40 & - & 40 & 128 & CNN & RNN & 89.70 \\ \hline
Al-Anbary et al. \cite{al2021proposed} & 2021 & American SL & 10 & - & 6487 & 14 & PCA & \begin{tabular}[c]{@{}c@{}}Deep \\ Learning\end{tabular} & 95.75 \\ \hline
Wang et al. \cite{9906948_wang2022} & 2022 & Hand Movement & 4 & - & - & 61 &  \begin{tabular}[c]{@{}c@{}}Decoder-\\ensemble \\framework\end{tabular}  & Ensemble & 70 \\ \hline
Hossain et al.\cite{hosseini2022continuous} & 2022 & Hand Movement & - & - & - & 63 & \begin{tabular}[c]{@{}c@{}}Phase-locking \\value (PLV) \\ Multiple linear\\ regression (MLR)\end{tabular} & Anova & - \\ \hline
Altameem et al. \cite{altameem2022performance} & 2022 & NEUROML2020 & - & - & - & 19 & FFT & KNN, XGBoost & 88 \\ \hline
Tao et al. \cite{9899476_tao_2022} & 2022 & NEUROML2020 & - & - & - & 19 & FFT & KNN, XGBoost & 88\% \\ \hline
Y Ai et al.\cite{ai2023convolutional} & 2023 & American SL & 7 & - & N/A & 3 to 9 & \begin{tabular}[c]{@{}c@{}}Temporal \\ Contrast \\ Coding\end{tabular} & \begin{tabular}[c]{@{}c@{}}Spike \\ Encoding \\ and Spatial \\ Clustering\end{tabular} & 97.07 \\ \hline
Ganesan et al. \cite{iasc.2023.033759_ganesan_eeg2023} & 2023 & Hand Movement & 4 & - & - & 6 & Wavelet and FFT & DL & 97.20 \\ \hline
Kim, et al. \cite{kim2023highly} & 2023 & \begin{tabular}[c]{@{}c@{}}Ultrasonic \\patterns\\finger\\motion \end{tabular} & - & - & - & - & \begin{tabular}[c]{@{}c@{}}3D GB-RRAM \\ neuromorphic\\ sensory system\end{tabular} &Softmax & \begin{tabular}[c]{@{}c@{}}97.9 ( motion) \\ 97.4 \end{tabular} \\ \hline
López et al. \cite{BaronaLopez2024} & 2024 & \begin{tabular}[c]{@{}c@{}}EMG-EPN-612 \\ (EMG)\end{tabular} & 5 & - & 183,600 & 8 & \begin{tabular}[c]{@{}c@{}}Spectrograms, \\ Rectification, \\ Filtering\end{tabular} & \begin{tabular}[c]{@{}c@{}}CNN, \\ CNN-LSTM\end{tabular} & \begin{tabular}[c]{@{}c@{}}90.55 \end{tabular} \\ \hline
\end{tabular}
\end{table*}

\subsection{EEG Based Methodology}
The process of recognizing hand gestures using EEG signals involves three main steps: signal preprocessing, feature extraction, and classification, which is demonstrated in Figure \ref{fig:basic_hgr_recognition_model}.

\subsubsection{Preproecssing Filtering and Signal Segmentation}
EEG signals contain high noise due to their narrow strategy, necessitating preprocessing for noise removal. Researchers commonly use a bandpass filter to eliminate artefacts from the raw EEG signal, such as eye blinking, sudden sounds, and muscle movements. For Motor Imagery (MI) tasks, the EEG bandwidth is often subdivided into narrower frequency bands like Mu-band (8-13 Hz), low-beta (13-22 Hz), and high-beta (22-35 Hz) to enhance classification accuracy. Studies show that brain activity during MI tasks primarily falls between 7 Hz and 36 Hz, supporting the use of narrowband signals for feature extraction \cite{miah2019motor, miah2017motor}

\subsubsection{Feature Extraction and ML algorithms}
In the research presented by AlQattan et al. \cite{alqattan2017towards}, EEG signal analysis involved utilizing nine different feature types: cD1, cD2, cD3, cD4, cD5, cA5, Energy, Entropy, and standard deviation. These features were employed to analyze EEG signals and identify hand movements associated with sign language from brain activity. 
In the research by Al-Quattan et al. \cite{al2021proposed}, EEG signals were classified into five types (Theta, Delta, Beta, Alpha, and Gamma waves) to capture various brain activities. Preprocessing was performed on these signals, followed by the application of PCA for unsupervised feature extraction. PCA is effective in reducing data dimensionality while preserving essential information, enhancing the characterization of EEG signals by emphasizing relevant information and reducing noise and non-relevant data. 
In the work by Spampinato et al. \cite{spampinato2017deep}, a low-dimensional manifold within the multidimensional and temporally varying EEG signals was extracted, resulting in a 1D representation referred to as EEG features. These features primarily encoded visual data, facilitating the extraction of corresponding image descriptors for automated classification. Additionally, in \cite{ai2023convolutional}, spike-related features were utilized by applying a temporal contrast coding scheme, translating measured analog signals into spike streams. These spike streams, consisting of positive and negative spikes, were used as features in the classification process. Alqattan et al. \cite{alqattan2017towards} used SVM and Linear Discriminant Analysis (LDA) algorithms for classifying EEG signals, achieving an accuracy rate of approximately 75\% with the Entropy feature type. In EEG-based BCI for SLR, the SVM and LDA algorithms achieved 75\% accuracy. Bozal et al. \cite{bozal2017personalized} developed DL models, achieving an 89.03\% accuracy in semantic image classification.

\subsubsection{CNN-Based Methods}
Bozal et al. \cite{bozal2017personalized} introduced a universal end-to-end DL model designed to predict the semantic content of images from the ImageNet dataset, achieving an accuracy of 89.03\%. In \cite{al2021proposed}, the researchers employed a neural network architecture with three hidden layers and a DL classifier for classifying ten classes of EEG signals, encompassing facial expressions and motor execution processes.
In another study by Bozal et al. \cite{bozal2017personalized}, LSTM, a type of recurrent neural network (RNN), was used to extract features from raw EEG signals. LSTM effectively captures sequential information from time-series data like EEG signals, enhancing the accuracy of gesture recognition.
Ai et al. \cite{ai2023convolutional} utilized Recurrent Neural Networks (RNNs) to capture discriminative brain activity related to visual categories using EEG data. They also trained a CNN-based regressor to project images onto the learned manifold, enabling machines to leverage human brain-based features for automated visual classification. The methodology integrated spike encoding and spatial clustering techniques to achieve accurate gesture signal classification, capturing rich spatiotemporal patterns in brain signals.
The study in \cite{alqattan2017towards} demonstrated the potential of EEG-based motor imagery brain-computer interfaces (BCIs) for linguistic communication with paralyzed patients, achieving approximately 75\% accuracy in identifying hand movements associated with sign language. In \cite{bozal2017personalized}, DL models predicted the semantic content of images based on EEG signals, with the universal model achieving 89.03\% accuracy and the personalized model reaching 90.34\%. The sign language software model in \cite{al2021proposed} achieved a high classification accuracy of 95.75\% for EEG signal samples, offering a promising approach to assist speechless individuals in communicating their thoughts non-invasively. The brain-driven approach for automated object categorization using EEG signals in \cite{spampinato2017deep} achieved an average accuracy of approximately 83\%, demonstrating the potential of brain signals for visual classification tasks.
In \cite{ai2023convolutional}, the algorithmic framework achieved superior accuracy in identifying hand gestures and motor imagery tasks, with accuracy ranging from 92.74\% to 96.51\%, highlighting the effectiveness of the proposed convolutional spiking neural network for BCI classification tasks. Al et al. \cite{al2021proposed} propose an EEG-based sign language software model, achieving 95.75\% accuracy. Spampinato et al. \cite{spampinato2017deep} achieve an average accuracy of 83\% in automated object categorization via EEG signals. In addition, some researchers have employed EEG-based hand movement recognition. Wang et al. demonstrated the decoding of hand movement parameters from low-frequency EEG signals, highlighting the correlation between physical hand movements and neural activity \cite{9906948_wang2022}. Ganesan et al. explored spectral analysis and validation of parietal signals for different arm movements, emphasizing the differentiation of continuous EEG signals based on finger flexion movement \cite{iasc.2023.033759_ganesan_eeg2023}. Additionally, Hosseini Shalchyan et al. investigated the continuous decoding of hand movements from EEG signals using phase-based connectivity features, demonstrating the feasibility of this approach \cite{hosseini2022continuous}.
Furthermore, Altameem et al. analyzed the performance of ML algorithms for classifying hand motion-based EEG brain signals, focusing on controlling prosthetic hands for amputees \cite{altameem2022performance}. Tao et al. proposed a novel algorithm using Multivariate Empirical Mode Decomposition and CNNs to decode multi-class EEG signals of hand movements, demonstrating advancements in decoding methodologies \cite{9899476_tao_2022}. Zakrzewski et al. utilized EEG recordings from 52 healthy subjects engaged in motor imagery hand movements to classify tasks related to these movements, achieving notable accuracy \cite{zakrzewski2022vr}. Crell demonstrated the classification of hand movements in four directions using EEG signals, with accuracies ranging from 55.9\% to 80.2\%. This study focused on continuous kinematic decoding of hand movements, showcasing the potential of EEG signals for accurately classifying various hand movement tasks \cite{crell2024towards}. Fujiwara and Ushiba applied deep residual CNNs to differentiate between rest, left-hand movement, and right-hand movement tasks with high accuracy \cite{fujiwara2022deep}. Utilizing a larger dataset and advanced neural network architectures, they achieved precise decoding of hand movements, highlighting the importance of dataset size and model complexity in achieving high classification accuracy.

\subsection{Future Direction}
EEG-based HGR faces several challenges hindering its widespread adoption and accuracy. One significant hurdle is the inherent variability and noise in EEG signals caused by muscle activity, eye movements, and environmental interference. Filtering out these unwanted signals while retaining relevant information for gesture recognition is complex. Additionally, the limited spatial resolution of EEG electrodes makes capturing fine-grained hand movements accurately difficult, which restricts the ability to decode complex gestures with high precision. Furthermore, the temporal dynamics of hand gestures present a challenge in extracting relevant features from EEG signals, as gestures involve intricate patterns that evolve over time.

Future research in EEG-based HGR aims to overcome current challenges and advance the field towards more robust and accurate systems. One direction involves exploring advanced signal processing techniques to enhance the signal-to-noise ratio and extract discriminative features from EEG data effectively.  Finally, efforts should focus on developing subject-adaptive models and exploring practical applications of EEG-based HGR in domains such as human-computer interaction and assistive technology, emphasizing usability and real-world deployment considerations.

\section{Multimodality Dataset Based HGR} \label{sect7}
Multimodal data-based HGR integrates various data sources, including RGB, skeleton, EMG, and EEG, to accurately interpret hand movements. This approach offers enhanced accuracy, robustness, flexibility, reduced ambiguity, and improved user experience compared to single-modality methods. Combining information from multiple perspectives provides a comprehensive understanding of gestures, making it valuable for applications in human-computer interaction, virtual reality, and SLR. Multimodal HGR (HGR)is crucial for effective human-robot interaction (HRI)  \cite{cardenas2018multimodal,kazakos2019epic,khaire2018human}. 
\subsection{Dataset}
According to our study, multimodal datasets combining sEMG signals, RGB images, and depth images enhance HGR precision and reliability \cite{gao2021hand} demonstrated in Table \ref{tab:multimodal_fusion_dataset}. Wang et al. \cite{wang2020gesture} created a dataset of 3,000 hand gesture samples across ten classes, each sample including an image with intricate backgrounds and strain data from sensors on finger knuckles. Siddiqui et al. \cite{siddiqui2020multimodal} used a dataset of 140 trials per participant, featuring parallel acoustic recordings from ten microphones and motion data from IMUs, each trial lasting three seconds. Mahmud et al. evaluated their method using the DHG-14/28 and SHREC'17 track datasets, which contain depth sequences of 14 gestures performed by 28 and 50 individuals, respectively. Sun et al. \cite{sun2023gesture} compiled a database of 20,000 RGB and depth images representing various gestures, with each category containing 2,000 samples for training and testing. Qi et al. \cite{qi2023adaptive} utilized multimodal data sources, including depth vision data and sEMG signals from devices like the Leap Motion Controller and the Myo Armband. Table \ref{tab:multimodal_fusion_performance} demonstrated the multimodal datasets. 

\begin{table*}[]
\centering
\caption{Databases for Fusion Modality}
\label{tab:multimodal_fusion_dataset}
\begin{adjustwidth}{0cm}{0cm}
\begin{tabular}{|l|l|l|l|l|l|l|l|}
\hline
\begin{tabular}[c]{@{}l@{}}Dataset \\ Names\end{tabular} &Year& \begin{tabular}[c]{@{}c@{}}Modality \\ Name\end{tabular} &\begin{tabular}[c]{@{}l@{}}Language\end{tabular} & \begin{tabular}[c]{@{}l@{}}Class\\ sign\end{tabular} & Sub. & \begin{tabular}[c]{@{}l@{}}Total\\ videos\end{tabular} &  \begin{tabular}[c]{@{}l@{}}Latest \\ Performance\end{tabular} \\ \hline
WLASL \cite{li2020word_d_skel_WLASL}    &2020& \begin{tabular}[c]{@{}c@{}}RGB  \\ Skeleton\end{tabular}     & ASL&2000   &-&8227&54.69\cite{hu2021signbert}  \\\hline
AUTSL \cite{sincan2020autsl_autsl}& 2020 &\begin{tabular}[c]{@{}c@{}}RGB  \\ Skeleton\\Depth\end{tabular}& Turkey & 226 & \begin{tabular}[c]{@{}l@{}}43 \end{tabular} & \begin{tabular}[c]{@{}l@{}}38336\\\end{tabular} &  98.00\cite{Jiang2021} \\
Wang et al. \cite{wang2020gesture}   &2020  & RGB, Depth &Roman numerals                                     & 10   &-&-&-   \\ \hline
Siddiqui et al. \cite{siddiqui2020multimodal}  &2020& RGB, Depth  & N/A& 13 &-&-&-    \\ \hline
Mahmud et al. \cite{mahmud2021deep}  &2021 & RGB, Depth & N/A& 14    &-&-&-  \\ \hline
Gao et al. \cite{gao2021hand} &2021   & RGB, Depth  & N/A& 10    &-&-&-  \\ \hline

HKSL \cite{Zhou2021b}    &2021& \begin{tabular}[c]{@{}c@{}}RGB  \\ Skeleton\end{tabular}      & N/A& N/A   &-&-&94.6\cite{Zhou2021b}  \\\hline

 SLR500 \cite{Zhou2021b}    &2021& \begin{tabular}[c]{@{}c@{}}RGB  \\ Skeleton\end{tabular}     & American&200   &-&125000&-\cite{Zhou2021b}  \\\hline

NMFCSL \cite{Zhou2021b}    &2021& \begin{tabular}[c]{@{}c@{}}RGB  \\ Skeleton\end{tabular}     & Chinese&500   &-&32,010&78.40\cite{hu2021signbert}  \\\hline

MSASL \cite{Zhou2021b}    &2021& \begin{tabular}[c]{@{}c@{}}RGB  \\ Skeleton\end{tabular}     & ASL&200   &-&32,010&71.4\cite{hu2021signbert}  \\\hline

KArSL\cite{Sidig2021_KArSL}& 2021&\begin{tabular}[c]{@{}c@{}}RGB  \\ Skeleton\\Depth\end{tabular}& Arabic & 502 & \begin{tabular}[c]{@{}l@{}} \end{tabular} & \begin{tabular}[c]{@{}l@{}}75,300 \\\end{tabular} &  95.84 \cite{shin2024japanese_jsl1} \\

How2Sign\cite{duarte2021how2sign}& 2021&\begin{tabular}[c]{@{}c@{}}RGB  \\ Skeleton\\Depth\end{tabular}& American & - & \begin{tabular}[c]{@{}l@{}} 11 \end{tabular} & \begin{tabular}[c]{@{}l@{}}2,456 \\\end{tabular} &  91.60 \cite{duarte2021how2sign} \\
GSL\cite{Adaloglou2021_GSL} & 2021&\begin{tabular}[c]{@{}c@{}}RGB
\\ Depth\end{tabular} & \begin{tabular}[c]{@{}l@{}} Greece
\end{tabular}  & 441&100&51,080 &  \begin{tabular}[c]{@{}l@{}}95.68 \cite{Adaloglou2021_GSL}  \\ 
\end{tabular} \\\hline

Sun et al.\cite{sun2023gesture}&2023 & RGB, Depth  & na& \begin{tabular}[c]{@{}c@{}}1000\\81 \end{tabular} &-&-&-\\  \hline
Qi et al. \cite{qi2023adaptive}    &2023& RGB, Depth      & N/A& N/A   &-&-&-  \\\hline
PHOENIX-2014 \cite{qi2023adaptive}    &2023& RGB, Depth      & N/A& N/A   &-&-&-  \\\hline
CSL \cite{qi2023adaptive}    &2023& RGB, Depth      & N/A& N/A   &-&-&-  \\
 PHOENIX-2014-T \cite{qi2023adaptive}    &2023& RGB, Depth      & N/A& N/A   &-&-&-  \\\hline

\hline
\end{tabular} 
\end{adjustwidth}
\end{table*}

\begin{table*}[]
\centering
\caption{Fusion data modality-based HGR. }
\label{tab:multimodal_fusion_performance}
\begin{tabular}{|c|c|c|c|c|c|c|c|c|}
\hline
\textbf{Author}   &Year  & \begin{tabular}[c]{@{}c@{}}Modality \\ Name\end{tabular}     & \textbf{Lang.} & \textbf{Class} & \textbf{Sample} & \textbf{\begin{tabular}[c]{@{}c@{}}Feature \\ Extraction\end{tabular}}                                 & \textbf{Classifier}                                       & \textbf{Performance} \\ \hline
Cheng et al. \cite{Cheng2016}&2016&\begin{tabular}[c]{@{}c@{}}RGB  \\, Depth\end{tabular}  & ChaLearn L.I. &  & SoftMax & C3D+LSTM &  & 68.14 \\ \hline
Sun et al. \cite{Sun2018}&2018&\begin{tabular}[c]{@{}c@{}}RGB  \\, Depth\end{tabular}   & SKIG &  10 & 6  & C3D+LSTM &  SoftMax& 98.60 \\ \hline
Wang et al. \cite{wang2018depth}  &2018&\begin{tabular}[c]{@{}c@{}}RGB  \\, Depth\end{tabular}    & American SL                 & 15   & N/A  & CNN  & Bi-LSTM   & 98.80\% \\ \hline

Sang et al. \cite{sang2018micro} &2018&\begin{tabular}[c]{@{}c@{}}RGB  \\, Depth\end{tabular}    & American SL                 & 6              & 2700            & \begin{tabular}[c]{@{}c@{}}Moving   Trajectories, \\    Precise   Ranges,\\    Velocities\end{tabular} & SVM                                                       & 96.32\%              \\ \hline
Siddiqui et al. \cite{siddiqui2020multimodal}&2020 &\begin{tabular}[c]{@{}c@{}}RGB  \\, Depth\end{tabular} & American SL                 & 36             & N/A             & \begin{tabular}[c]{@{}c@{}}RMS,   MAV\\    Skew,  Kurtosis\end{tabular}                                & \begin{tabular}[c]{@{}c@{}}SVM, DT,\\    LDA\end{tabular} & 80.00\%              \\ \hline
Neverova, et, al.  \cite{ling2020ultragesture} &2020&\begin{tabular}[c]{@{}c@{}}RGB  \\, Depth\end{tabular}      & Chalearn 2014                 &       20      & 13,858              & ModDrop                                                                                                   & CNN                                                       & 96.81\%              \\ \hline

Zhou et al. \cite{Zhou2020} &2020&\begin{tabular}[c]{@{}c@{}}PHOENIX\\-2014\\ CSL \\ PHOENIX-\\2014-T.\end{tabular}       & \begin{tabular}[c]{@{}c@{}}RGB  \\Skeleton\end{tabular}             & 41              & -             & CNN + TCN                                                                                              &  \begin{tabular}[c]{@{}c@{}} SBiLSTM \\+ CTC \end{tabular}                                                         & -             \\ \hline
Gao et al. \cite{gao2020two} &2020 &\begin{tabular}[c]{@{}c@{}}RGB  \\, Depth\end{tabular} & ASL &  &  & 2S-CNN &  SoftMax& 92.00 \\ \hline
Gao et al. \cite{gao2021hand} &2020 &\begin{tabular}[c]{@{}c@{}}RGB  \\, Depth\end{tabular} & ASL & 10 &  & CNN &  SoftMax& - \\ \hline

Duarte et al. \cite{duarte2021how2sign} &2021 &\begin{tabular}[c]{@{}c@{}}RGB  \\, Depth\end{tabular}    &   Multi-modal ASL                & 8           & 16000           &            How2Sign                                                                                     &DL                                                  & 91.60\%              \\ \hline
Mahmood  et al. \cite{mahmud2021deep} &2021 &\begin{tabular}[c]{@{}c@{}}RGB  \\, Depth\end{tabular}     & Depth and Shrec                 & 28            & 2800            & CRNN                                                                                                   & Softmax                                                  & 90.04\%              \\ \hline

Zhou et al.  \cite{Zhou2021b} &  2021 &\begin{tabular}[c]{@{}c@{}}RGB  \\, Skeleton \\Depth \end{tabular} & \begin{tabular}[c]{@{}c@{}}PHOENIX14-T\\ CSL\\HKSL \\GSl\end{tabular}            & \begin{tabular}[c]{@{}c@{}}8227\\100\\ 50 \\ \end{tabular}           &   \begin{tabular}[c]{@{}c@{}}8227\\25000\\ 2400\\ \end{tabular}        & \begin{tabular}[c]{@{}c@{}}(3+2+1)D ResNet\\ +
BiLSTM +\\ BERT\end{tabular}  & CTC  & \begin{tabular}[c]{@{}c@{}}18.60\\ 19.80\\7.19 \\ 19.80 WEN\end{tabular}            
\\ \hline
Hu et al.  \cite{hu2021signbert} &  2021 &\begin{tabular}[c]{@{}c@{}}RGB  \\ Skeleton\end{tabular}& \begin{tabular}[c]{@{}c@{}}WLASL\\ MSASL\\SLR500\\NMFCSL\\STB\\HANDS17 \end{tabular}            & \begin{tabular}[c]{@{}c@{}}2000\\ 200\\100\\ 500\\1067 \\-\\ -\\ \end{tabular}           &   \begin{tabular}[c]{@{}c@{}}8227\\25000\\ 125,000\\32,010\\18000\\292820 \end{tabular}        & \begin{tabular}[c]{@{}c@{}} BERT\end{tabular}  & SoftMax  & \begin{tabular}[c]{@{}c@{}}97.60\\ 59.54\\-\\46.39\\95.48(AUC)\\- \end{tabular}            
\\ \hline

Ding et al. \cite{ding2022cnn}  &2022&\begin{tabular}[c]{@{}c@{}}RGB  \\, Depth\end{tabular}      & depth-grayscale                  & 10            & N/A             & VGG-16 CNN                                                       & Softmax                & 83.88\%              \\ \hline

Yang et et al. \cite{yang2023real} &2023  &\begin{tabular}[c]{@{}c@{}}RGB  \\, Depth\end{tabular}    & FMCW                 & 8           & -           & Multi-modal                                                                                                  &ML                                                  & 93.30\%              \\ \hline
Qi et al. \cite{qi2023adaptive}&2023 &\begin{tabular}[c]{@{}c@{}}RGB  \\, Depth\end{tabular}    & Hand gesture                  & 7              & 128             & Amplitude                                                                                              & KNN                                                     & 96.00\%              \\ \hline

Liu et al. \cite{liu2023multimodal} & 2023 & \begin{tabular}[c]{@{}c@{}}RGB  \\ Signal\end{tabular} & HGR&10 &  \begin{tabular}[c]{@{}c@{}}2400tr\\ 7200 tst\end{tabular} &  \begin{tabular}[c]{@{}c@{}}Two-branch \\fusion \\ network \\ LSTM\end{tabular} & Softmax& \begin{tabular}[c]{@{}c@{}}94.58\% (cr-valid), \\ 94.63\% (out nor), \\ 93\% (in nor), \\ 93.68\% (out dark), \\ 89\% (in dark), \\ 83\% (bright light)\end{tabular} \\ \hline
Duan et al. \cite{duan2023alignment_aifusion} & 2023 & \begin{tabular}[c]{@{}c@{}}sEMG\\ ACC\end{tabular}&  \begin{tabular}[c]{@{}c@{}}DB2, DB3\\ DB5, DB6,\\ DB7\end{tabular}  &\begin{tabular}[c]{@{}c@{}}50, 50, 41, \\ 8, 41\end{tabular} & \begin{tabular}[c]{@{}c@{}}- \end{tabular} &  \begin{tabular}[c]{@{}c@{}}CNN\\ Transformer\end{tabular} & AiFusion & \begin{tabular}[c]{@{}c@{}} 95.28 \\ 91.11 \\ 87.04 \\ 80.62 \\ 96.76\end{tabular} \\ \hline
Wang et al. \cite{wang2023hand} & 2023 & DB5(E+A) & HGR & - & - & HybridCNN & \begin{tabular}[c]{@{}c@{}} CNN-LSTM\\-CBAM\end{tabular}  & 92.16 \\ \hline
Duan et al. \cite{duan2023hybrid} & 2023 & sEMG,ACC& \begin{tabular}[c]{@{}c@{}}DB2 \\DB3 \\ DB7\end{tabular}  & - &  \begin{tabular}[c]{@{}c@{}}30 tr\end{tabular} &  HyFusion &Softmax & \begin{tabular}[c]{@{}c@{}} 94.73 \\ 89.60\\ 96.44\end{tabular} \\ \hline

Shin et al. \cite{shin2024japanese_jsl1} &2024&\begin{tabular}[c]{@{}c@{}}RGB  \\, Skeleton\end{tabular}       & JSL                & 41              & -             & \begin{tabular}[c]{@{}c@{}} Handcrafted \\and CNN\end{tabular}                                                                                              & SoftMax                                                       & 90.00\%              \\ \hline
Balaji  et al. \cite{Balaji2024} &2024 &\begin{tabular}[c]{@{}c@{}}RGB  \\ skeleton \\ Depth\end{tabular}    & Shrec   & \begin{tabular}[c]{@{}c@{}}14\\28\end{tabular}  & 2800    &  \begin{tabular}[c]{@{}c@{}} MF-HAN.\end{tabular}  & Softmax     & \begin{tabular}[c]{@{}c@{}}93.93\\92.26    \end{tabular}           \\ \hline
Balaji  et al. \cite{Balaji2024} &2024 &\begin{tabular}[c]{@{}c@{}}RGB  \\ skeleton \\ Depth\end{tabular}     & Shrec  & \begin{tabular}[c]{@{}c@{}}14\\28\end{tabular}  & 2800   &  \begin{tabular}[c]{@{}c@{}} MF-HAN\\ 2 stream.\end{tabular}   & Softmax  & \begin{tabular}[c]{@{}c@{}}94.17\\93.21   \end{tabular} \\ \hline
Wang  et al. \cite{wang2024fusion} &2024 &\begin{tabular}[c]{@{}c@{}} 
- \end{tabular} &- & \begin{tabular}[c]{@{}c@{}}14\\28\end{tabular} & 2800 &  \begin{tabular}[c]{@{}c@{}}  Haar\\ Wavelet \\Transform, \\LBP.\end{tabular} &Softmax&-\\ \hline
\end{tabular}
\end{table*}
\subsection{Fusion Based Methodology}
Many researchers have been working to develop fusion-based HGR systems. These systems commonly use ML and DL techniques for feature extraction and classification. Below, we describe the technologies involved in these systems. Figure \ref{fig:basic_hgr_recognition_model} also demonstrated the multimodal data fusion-based approach outline diagram. Table \ref{tab:multimodal_fusion_performance} demonstrated the multimodal fusion-based HGR methodology and its performance. 

\subsubsection{ML-based Fusion System} 
\begin{figure}[htp]
    \centering
    \includegraphics[width=9cm]{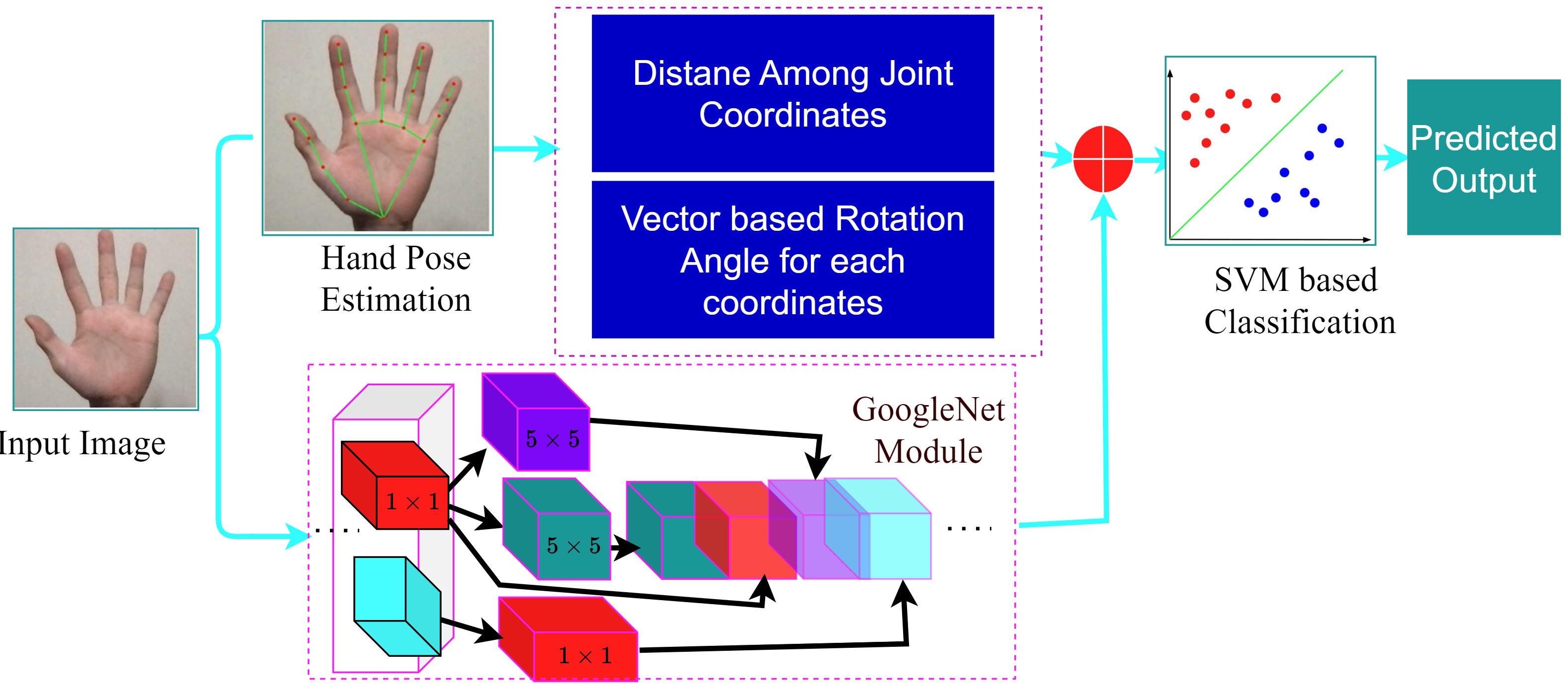}
    \caption{Fusion skeleton and RGB based fusion \cite{shin2024japanese_jsl1}}
    \label{fig:skelton_musa_jsl1}
\end{figure}

Siddiqui et al. \cite{siddiqui2020multimodal} achieved 75\% accuracy with a general model using the top 25 features selected via the mRMR algorithm. When these features were taken from a single microphone unit at the mid-anterior wrist, and an IMU, intra-subject average accuracy exceeded 80\%. Using SVM and LDA methods with these selected features, average accuracy levels also surpassed 80\%. Qi et al. \cite{qi2023adaptive} achieved high accuracy and computational efficiency, with the depth vision-based k-NN classifier attaining a remarkable 100\% identification accuracy for gestures, highlighting its effectiveness.
Shin et al. proposed a fusion-based Japanese SLR system using a skeleton and RGB information as inputs \cite{shin2024japanese_jsl1}. They combined hand-crafted features from skeleton data with DL features from RGB data for classification, achieving good performance with Japanese and Arabic sign language datasets. Figure \ref{fig:skelton_musa_jsl1} illustrates their module's architecture. Ding et al. integrated CCD RGB-IR and depth-grayscale sensor data using CNN DL for hand gesture intention recognition, emphasizing the importance of multi-modal approaches to improve recognition performance \cite{ding2022cnn}.

\subsubsection{CNN-based Systems}
Gao et al. \cite{gao2021hand} implemented a multiscale parallel CNN for HGR by fusing sEMG signals, RGB images, and depth images. These inputs were downscaled and processed through parallel CNN subnetworks, effectively extracting relevant features for final gesture recognition. Wang et al. \cite{wang2020gesture} used a CNN to extract hierarchical deep spatial and shift-invariant features from hand gesture images. They applied a sparse neural network for feature-level fusion and recognition of sensor data, achieving high accuracy. They reported 100\% accuracy in human gesture recognition, even with noisy or over/under-exposed images, and an error rate of 1.7\% under standard illumination and 3.3\% in low-light conditions for robot navigation through hand gestures.

\begin{figure}[htp]
    \centering
    \includegraphics[width=8cm]{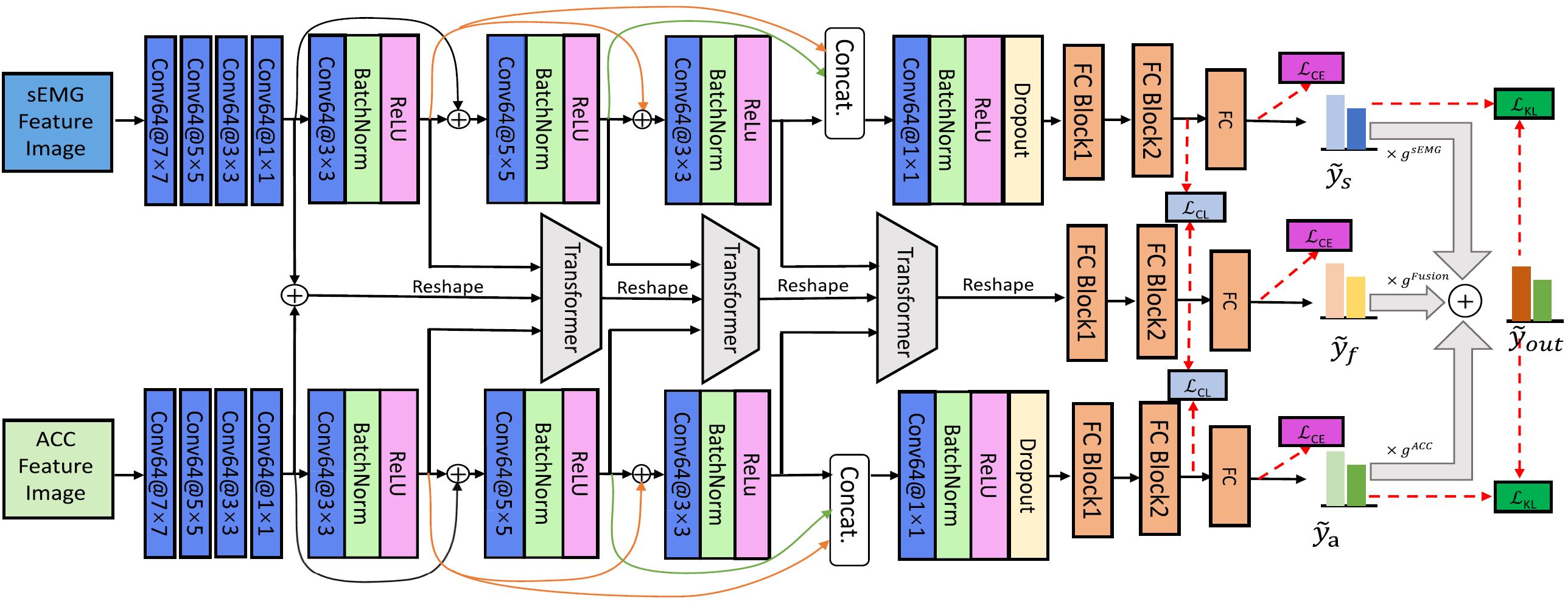}
    \caption{ AiFusion: Fusion model for complete and incomplete multimodal HGR\cite{duan2023alignment_aifusion}.}
    \label{fig:multi_CNN_Transformer_AIFusion}
\end{figure}

\begin{figure}[htp]
    \centering
    \includegraphics[width=8cm]{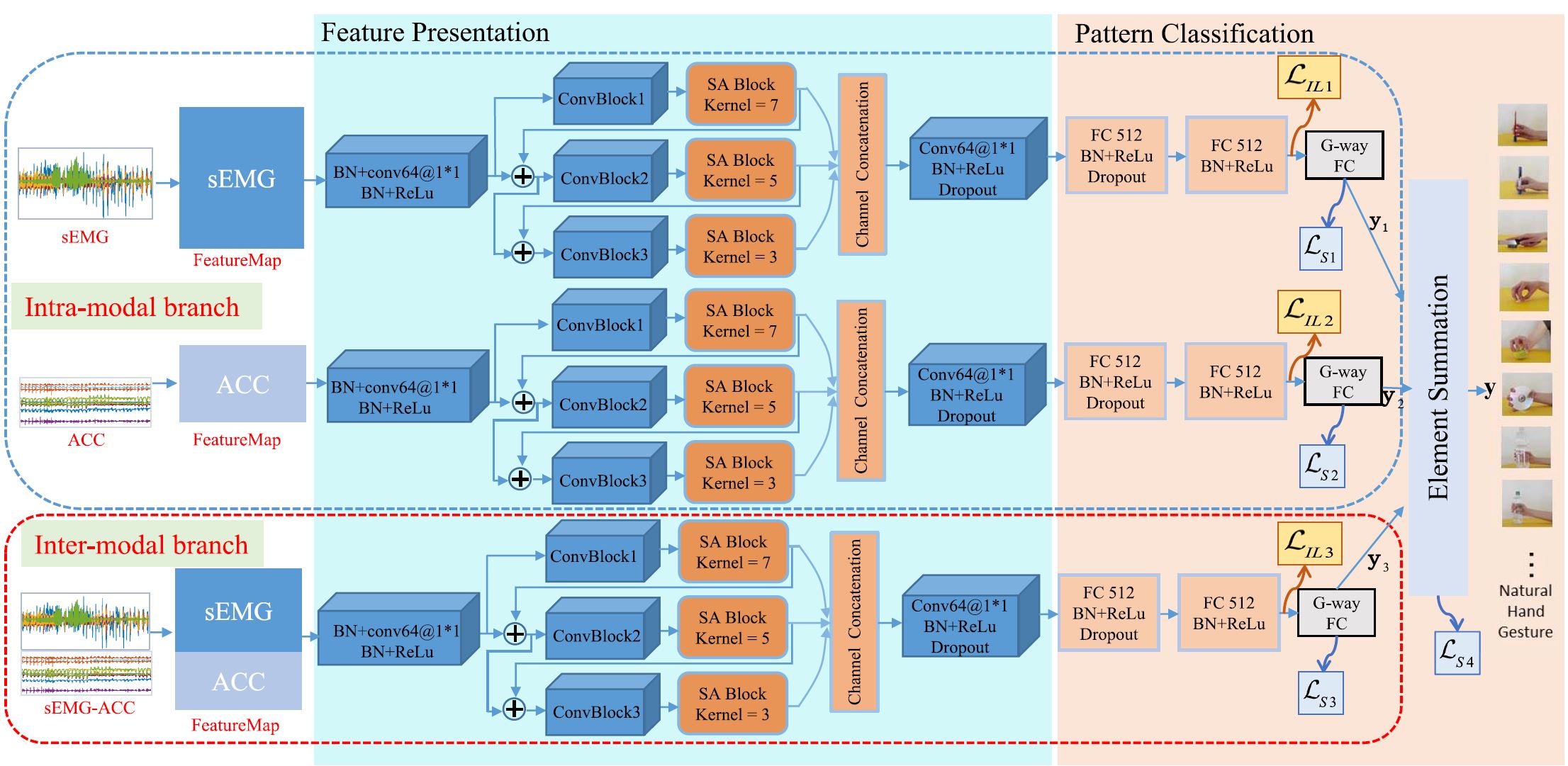}
    \caption{ HyFusion: Hybrid Multimodal Fusion Model\cite{duan2023hybrid}.}
    \label{fig:multi_HyFusion}
\end{figure}

Neverova et al. proposed the ModDrop model, an adaptive multi-modal gesture recognition approach utilizing multi-scale and multi-modal DL. This method employs a training strategy known as ModDrop, which involves the careful initialization of individual modalities and their gradual fusion to learn cross-modality correlations while preserving each modality's unique representation \cite{ling2020ultragesture}. Experiments on the ChaLearn 2014 Looking at People Challenge gesture recognition track demonstrated that fusing multiple modalities at various spatial and temporal scales significantly increases recognition rates, compensating for errors and noise in individual channels, achieving 96.81\% accuracy.
How2Sign\cite{duarte2021how2sign}is a multimodal and multiview dataset for continuous ASL, featuring RGB, depth, and 2D key point data from both frontal and side views. The dataset includes up to 80 hours of video content.
Jiang et al. \cite{jiang2021sign} employed a multi-modal approach for feature extraction, incorporating RGB-based 3DCNN, hand RGB, key points, RGB frames, and RGB flow. They used the Sign Language Graph Convolution Network (SL-GCN) to capture the embedded dynamics of skeleton key points. Their model achieved the highest performance in both the RGB (98.42\%) and RGB-D (98.53\%) tracks, demonstrating its effectiveness in SLR. The results indicated that their approach outperforms both ResNet3D and ResNet21D, showcasing its superiority. Recently, Duan et al. proposed an AiFusion: Alignment-Enhanced Interactive Fusion Mode by integrating the CNN with various levels of the transformer \cite{duan2023alignment_aifusion} where they reported     DB2: 95.28\%,    DB3: 91.11\%,    DB5: 87.04\%,    DB6: 80.62\%,    DB7: 96.76\% accuracy. The details of the AiFusion working mechanism are shown in Figure \ref{fig:multi_CNN_Transformer_AIFusion}.  Dean et al. also proposed a hybrid multimodal fusion model including DL and transformer \cite{duan2023hybrid}. They reported DB2: 94.73\%,  DB3: 89.60\%,  DB7: 96.44\%\ accuracy with their model. Figure \ref{fig:multi_HyFusion} demonstrated the working mechanism of this model.

\subsubsection{BiLSTM and Hybrid Systems}
Mahmud et al. \cite{mahmud2021deep} utilized a variational autoencoder, CNNs, and Long-Short Term Memory (LSTM) networks for fusion-based HGR. They effectively captured spatial and temporal information by extracting finger motion and global motion features, reporting a 14\% improvement in performance accuracy. However, they encountered limitations related to dataset size and training time.
\begin{figure}[htp]
    \centering
    \includegraphics[width=6cm]{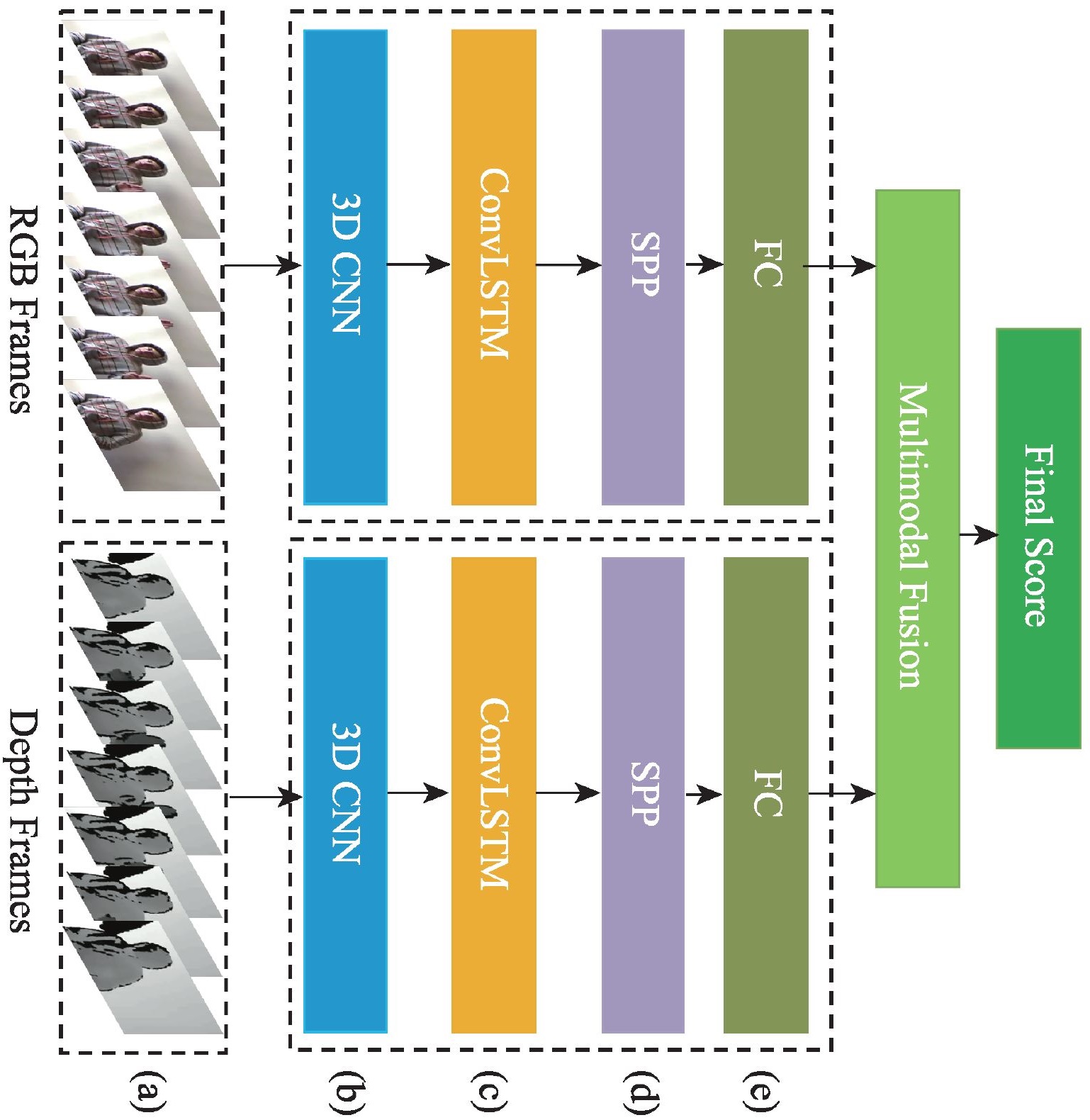}
    \caption{ C3D+LSTM method approach\cite{mahmud2021deep}.}
    \label{fig:multi_c3D_LSTM}
\end{figure}

\subsubsection{Transformer-based Systems}
Sun et al. \cite{sun2023gesture} integrated multi-level feature fusion using a two-stream CNN, reporting a 1.08\% improvement in average detection accuracy and a 3.56\% increase in mean average precision (mAP) compared to a single-channel model. They achieved an average gesture recognition rate of 93.98\% under occlusion and varying light conditions, enhancing robustness against challenging scenarios but facing potential limitations with dataset variability. Qi et al. \cite{qi2023adaptive} used an ensemble classifier combining the k-nearest neighbour method for finger angle analysis and a deep CNN for gesture identification based on sEMG signals. Despite its adaptive learning mechanism, this approach involves complexities in model training and optimization.

\subsection{Challenges and Future Direction}
Challenges in multimodal dataset-based HGR include managing data variability across different sensors, synchronizing multimodal data streams, addressing computational complexity, ensuring robustness in diverse environments, and overcoming limited training datasets. Advanced integration techniques and efficient algorithms are required to enhance recognition accuracy and system performance. Integrating DL architectures tailored to handle data variability across different sensors and synchronizing data streams, reducing computational complexity, can be fruitful work in this domain in the future.
\section{Discussion}
In this study, we review the current advancements in HGR (HGR) across diverse data modalities. We present a comprehensive framework that addresses key elements of feature extraction using ML, as well as end-to-end DL modules. Each data modality is explored in detail, including primary datasets, preprocessing techniques, proposed architectures, fusion methodologies, state-of-the-art performance, and the challenges and future trends specific to each modality. For dynamic HGR, we emphasize the importance of managing temporal contextual features. This section aims to guide future research by highlighting the challenges and potential directions involving 3D DL models. We evaluate the associated problems, complexities, and the need for practical solutions within the field, providing a clear pathway for future advancements.

We identify and discuss new research gaps and offer guidance to overcome challenges in next-generation HGR technologies. Multimodal datasets, which integrate hand-crafted features with new CNN features combined with RGB, skeleton, and depth-based information, are examined for their efficacy in improving gesture recognition accuracy. In the skeleton modality section, we focus on GCNs and their application to large-scale, real-time HGR. This has become a focal point of the research community due to its potential to solve significant challenges. We highlight the need for gesture localization within realistic, uncut, and extended videos, predicting that emerging challenges such as early recognition, multi-task learning, gesture captioning, recognition from low-resolution sequences, and life log devices will gain increased attention in the coming years. Our discussion underscores the necessity of addressing these challenges to advance HGR systems, with a particular emphasis on multimodal approaches and the integration of new technologies. By providing a detailed overview of the current state and future directions, we hope this study serves as a guideline for researchers and practitioners in the field of HGR.
        
\section{Challenges and Future Directions in Diverse Data Modality-Based HGR}
Currently, research in diverse data modality-based HGR (HGR) faces several significant challenges. While isolated gesture recognition achieves high accuracy in small to medium-sized datasets, it struggles with large-scale datasets. Continuous gesture recognition performs well in simple scenarios but has room for improvement in complex environments. The following are specific challenges and future trends in this field:
\subsection{Short Duration of Videos}
One major challenge in HGR (HGR) is the short length of videos in most datasets. Real-world gestures often form long sequences without clear breaks between gestures. Current methods struggle with these long sequences due to limitations like RNNs' long-term dependency issues and the high computational demand of Transformer models. Future research should focus on improving models' ability to process long sequences and accurately segment gestures within continuous streams while also handling shorter sequences effectively.

\subsection{Unseen Gestures and User Independence}
Recognizing unseen gestures from different users is crucial for practical HGR applications. User independence means the model performs well across various individuals despite differences in gesture speed and body dimensions. Many current methods capture specific traits of individuals, leading to performance drops with new users. Recognizing unseen gestures, where test gestures weren't seen during training, is also challenging. Future research should develop methods that allow models to learn generalized features from large datasets and improve algorithms for gesture segmentation to handle unseen gestures effectively.
\subsection{Generalization Beyond Training Data}
The wide range of hand gestures makes it impractical to include all possible gestures in a dataset. Few-shot and zero-shot learning approaches help models recognize new gestures with minimal training data. While progress has been made in isolated gesture recognition using these techniques, continuous gesture recognition remains largely unexplored. Advancing these techniques for isolated and continuous gestures is crucial for creating robust and versatile HGR systems.

\subsection{Multi-Person Recognition}
Most current datasets focus on single-person scenarios, but real-world applications require robust recognition in multi-person environments. Models need to distinguish between gestures from the intended user and interference from others. Future research should focus on developing robust models that accurately focus on the target user while ignoring background interference.

\subsection{Model Complexity and Deployment}
Many HGR systems achieve high accuracy on servers but are too complex to use on portable devices. The goal is to create HGR systems that aid everyday communication, which requires lightweight models suitable for mobile use. Future research should prioritize developing efficient models that maintain high accuracy and are feasible for real-time use on portable devices.

\subsection{Online Recognition}
Current methods often use recorded datasets for validation, allowing models to access future context within sequences. Real-time gesture recognition requires models to predict based solely on past information, as future data isn't available. Developing methods that can perform online recognition and accurately predict gestures from one-directional sequences is crucial. Improving real-time performance will enhance the practicality and usability of HGR systems in dynamic, real-world scenarios.

\section{Conclusion} \label{sect8}
This comprehensive review has provided an in-depth analysis of advancements in vision-based HGR (HGR) for sign language from 2014 to 2024. By examining over 200 articles from reputable online databases, we have highlighted significant achievements and identified critical areas needing further exploration. The findings reveal a dynamic research landscape with a steady stream of publications across various journals and conferences, focusing predominantly on three critical aspects: data collection, data contextualization, and hand gesture representation. The review underscores the efficacy of HGR systems, particularly in terms of recognition accuracy, which remains a crucial benchmark in this domain. Notably, there is a significant gap in research on continuous gesture recognition, emphasizing the need for further efforts to enhance the precision and applicability of vision-based gesture recognition systems. This gap offers valuable perspectives for future research directions and highlights the evolving nature of this important field.
The advantages of current HGR systems include improved accuracy, robustness in varied conditions, and the potential for real-time application. Future work should focus on addressing the challenges in continuous gesture recognition, integrating multimodal approaches, and developing more sophisticated models that can handle diverse and complex gesture patterns. These efforts will be crucial in advancing the capabilities and practical applications of HGR systems, ultimately contributing to more natural and efficient human-computer interactions.


\section*{ABBREVIATIONS}
\begin{table}[H]
\label{Appendix_A}
\setlength{\tabcolsep}{3pt}
\begin{tabular}{ll}

HCI & Human-Computer Interfacing \\
BCI & Brain-Computer Interface \\
EEG & Electroencephalography \\
MEG & Magnetoencephalogram \\
RQ & Research Question \\ 
HGR & Hand Gesture Recognition \\
ASL & American Sign Language \\
ArSL & Arabic Sign Language \\
ISL & Indian Sign Language  \\
KSL & Kurdish Sign Language \\
JSL & Japanese sign language\\
PSL & Persian Sign Language \\
GSL & German Sign Language\\
DSL & Danish Sign Language\\
SLRS & Sign Language Recognition System\\
CNN & Convolutional Neural Network\\
ADDSL & Annotated Dataset for Danish Sign Language\\
ML & Machine Learning\\
DL & Deep Learning\\
HMM & Hidden Markov Model \\
HOG & Histogram of Oriented Gradient\\
PCA & Principal Component Analysis\\
CRNN & Convolutional Recurrent Neural Network\\
LSTM & Long Short-Term Memory \\
Bi-LSTM & Bidirectional Long Short-Term Memory \\
SVM & Support Vector Machine\\
ArSLRS & Arabic Sign Language Recognition System\\
CTC & Connectionist Temporal Classification\\
RTDHGRS & Real-Time Dynamic HGR System \\
SMKD &Self-Mutual Knowledge Distillation \\
CTC&connectionist temporal classification \\
MUD& Massey University Dataset \\
ASLAD & American Sign Language Alphabet Dataset \\
SSC-DNN & Spotted Hyena-based Sine Cosine \\
        &  Chimp Optimization Algorithm with Deep Neural Network\\
DMD & Dynamic Mode  Decomposition \\
MsMHA-VTN & Multiscaled Multi-Head  Attention Video Transformer  Network \\
HSL&Hongkong Sign Language \\
FPHA & First Person Hand Action\\
STr-GCN& Spatial Graph Convolutional Network and \\
& Transformer Graph Encoder for 3D HGR\\
MF-HAN& Multimodal Fusion Hierarchical Self-Attention Network\\
SMLT& Simultaneous Multi-Loss Training \\
ResGCNeXt&Efficient Graph Convolution Network\\
HDCAM & Hierarchical Depth-wise Convolution and Attention Mechanism\\
\end{tabular}
\end{table}
\newpage
\nocite{*} 
\bibliographystyle{unsrt}
\bibliography{reference}

\begin{IEEEbiography}[{\includegraphics[width=1in,height=1.25in,clip,keepaspectratio]{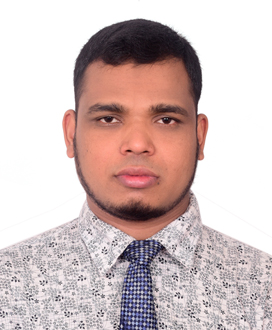}}]{Abu Saleh Musa Miah} received the B.Sc.Engg., and M.Sc.Engg. degrees in computer science and engineering from the Department of Computer Science and Engineering, University of Rajshahi, Rajshahi-6205, Bangladesh, in 2014 and 2015, respectively. He became a Lecturer and an Assistant Professor at the Department of Computer Science and Engineering, Bangladesh Army University of Science and Technology (BAUST), Saidpur, Bangladesh, in 2018 and 2021, respectively. He started his PhD degree in the School of Computer Science and Engineering at the University of Aizu, Japan, in 2021, under a scholarship from the Japanese government (MEXT). His research interests include CS, ML, DL, HCI, BCI and neurological disorder detection. He has authored and co-authored more than 20 publications published in widely cited journals and conferences.
\end{IEEEbiography}
\begin{IEEEbiography}[{\includegraphics[width=1in,height=1.25in, clip,keepaspectratio]{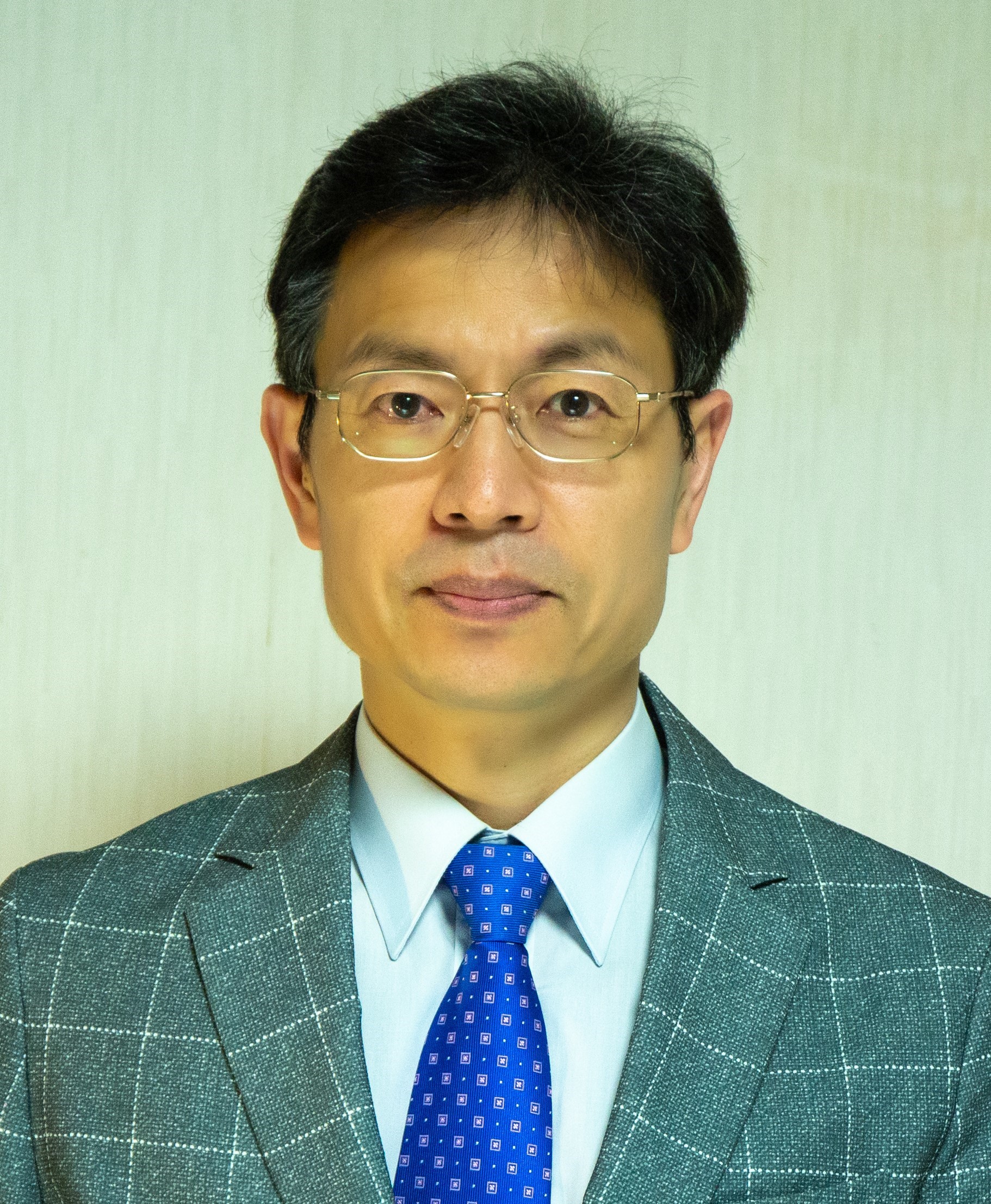}}]
{Jungpil Shin} (Senior Member, IEEE) received a B.Sc. in Computer Science and Statistics and an M.Sc. in Computer Science from Pusan National University, Korea, in 1990 and 1994, respectively. He received his Ph.D. in computer science and communication engineering from Kyushu University, Japan, in 1999, under a scholarship from the Japanese government (MEXT). He was an Associate Professor, a Senior Associate Professor, and a Full Professor at the School of Computer Science and Engineering, The University of Aizu, Japan in 1999, 2004, and 2019, respectively. His research interests include pattern recognition, image processing, computer vision, machine learning, human-computer interaction, nontouch interfaces, human gesture recognition, automatic control, Parkinson’s disease diagnosis, ADHD diagnosis, user authentication, machine intelligence, bioinformatics, as well as handwriting analysis, recognition, and synthesis. He is a member of ACM, IEICE, IPSJ, KISS, and KIPS. He served as general chair, program chair, and committee for numerous international conferences. He serves as an Editor of IEEE journals, Springer, Sage, Taylor and Francis, MDPI Sensors and Electronics, and Tech Science. He serves as an Editorial Board Member of Scientific Reports. He serves as a reviewer for several major IEEE and SCI journals. He has co-authored more than 400 published papers for widely cited journals and conferences.
\end{IEEEbiography}

\begin{IEEEbiography}[{\includegraphics[width=1in,height=1.25in,clip,keepaspectratio]{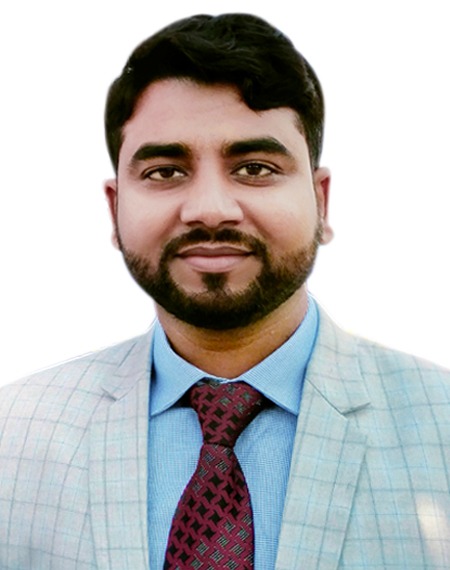}}]{MD. HUMAUN KABIR} is currently working as an Assistant Professor in the Department of Computer Science and Engineering at Bangamata Sheikh Fojilatunnesa Mujib Science \& Technology University, Jamalpur, Bangladesh. He has completed B.Sc. Engineering and M.Sc. Engineering degrees in Applied Physics \& Electronic Engineering from the University of Rajshahi, Bangladesh,  in 2014 and
2016, respectively. He is an active researcher in the fields of Digital Signal Processing, Brain-Computer Interface (BCI), Communication \& Networks, Machine Learning and Deep Learning. He has authored and co-authored more than 20 research articles published in widely cited national and international journals and conferences.
\end{IEEEbiography}

\begin{IEEEbiography}[{\includegraphics[width=1in,height=1.25in,clip,keepaspectratio]{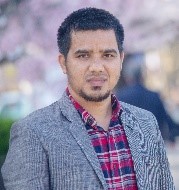}}]{Dr. Md Abdur Rahim} received his Ph.D. degree in 2020 from the Graduate School of Computer Science and Engineering, The University of Aizu, Fukushima, Japan. He received the Bachelor of Science (Honours) and Master of Science (M.Sc.) degrees in Computer Science and Engineering from the University of Rajshahi, Bangladesh, in 2008 and 2009. He is currently working as an associate professor and head of the Department of Computer Science and Engineering at Pabna University of Science and Technology, Pabna 6600, Bangladesh. His research interests include human-computer interaction, pattern recognition, computer vision and image processing, human recognition, and machine intelligence. He has several publications in major journals (SCI and SCIE indexed) and conferences and also serves as a reviewer for several SCI/SCIE indexed journals and international conferences.
\end{IEEEbiography}

\begin{IEEEbiography}[{\includegraphics[width=1in,height=1.25in,clip,keepaspectratio]{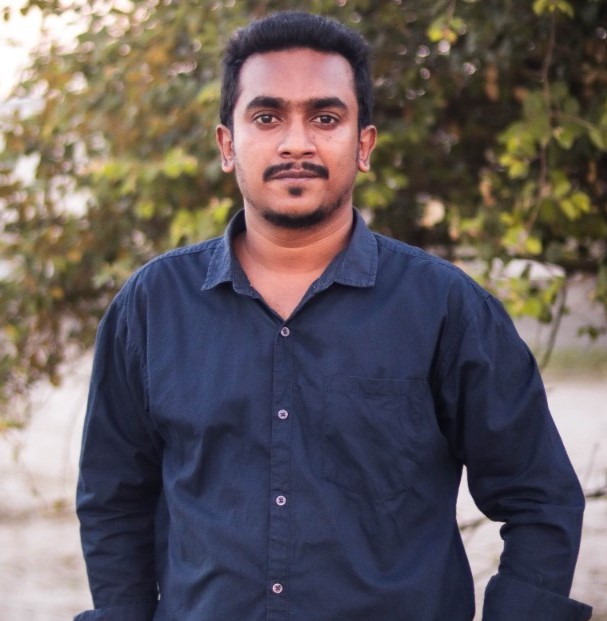}}]{ABDULLAH AL SHIAM} received the B.Sc. and M.Sc. degrees in computer science and engineering from the University of Rajshahi, Bangladesh, in 2017 and 2019, respectively. He was a Research Fellow with the Information and Communication Technology Division (ICT Division), Ministry of Posts, Telecommunications and Information Technology, Government of the People’s Republic of Bangladesh, from 2018 to 2019. From January, 2019 to May, 2020, he worked as a Lecturer in the dept. of Computer Science and Engineering, Varendra University, Rajshahi, Bangladesh. He is currently working as a Lecturer in the Department of Computer Science and Engineering, Sheikh Hasina University, Netrokona, Bangladesh. His research interests include EEG Signal Processing, Brain-Computer Interface (BCI), and Biomedical Engineering.
\end{IEEEbiography}

\EOD

\end{document}